\PassOptionsToPackage{unicode}{hyperref}
\PassOptionsToPackage{hyphens}{url}
\documentclass[
]{article}
\usepackage{xcolor}
\usepackage[margin=1in]{geometry}
\usepackage{amsmath,amssymb}
\usepackage{iftex}
\ifPDFTeX
  \usepackage[T1]{fontenc}
  \usepackage[utf8]{inputenc}
  \usepackage{textcomp} % provide euro and other symbols
\else % if luatex or xetex
  \usepackage{unicode-math} % this also loads fontspec
  \defaultfontfeatures{Scale=MatchLowercase}
  \defaultfontfeatures[\rmfamily]{Ligatures=TeX,Scale=1}
\fi
\usepackage{lmodern}
\ifPDFTeX\else
\fi
\IfFileExists{upquote.sty}{\usepackage{upquote}}{}
\IfFileExists{microtype.sty}{% use microtype if available
  \usepackage[]{microtype}
  \UseMicrotypeSet[protrusion]{basicmath} % disable protrusion for tt fonts
}{}
\makeatletter
\@ifundefined{KOMAClassName}{% if non-KOMA class
  \IfFileExists{parskip.sty}{%
    \usepackage{parskip}
  }{% else
    \setlength{\parindent}{0pt}
    \setlength{\parskip}{6pt plus 2pt minus 1pt}}
}{% if KOMA class
  \KOMAoptions{parskip=half}}
\makeatother
\usepackage{longtable,booktabs,array}
\usepackage{calc} % for calculating minipage widths
\usepackage{etoolbox}
\makeatletter
\patchcmd\longtable{\par}{\if@noskipsec\mbox{}\fi\par}{}{}
\makeatother
\IfFileExists{footnotehyper.sty}{\usepackage{footnotehyper}}{\usepackage{footnote}}
\makesavenoteenv{longtable}
\usepackage{graphicx}
\makeatletter
\newsavebox\pandoc@box
\newcommand*\pandocbounded[1]{% scales image to fit in text height/width
  \sbox\pandoc@box{#1}%
  \Gscale@div\@tempa{\textheight}{\dimexpr\ht\pandoc@box+\dp\pandoc@box\relax}%
  \Gscale@div\@tempb{\linewidth}{\wd\pandoc@box}%
  \ifdim\@tempb\p@<\@tempa\p@\let\@tempa\@tempb\fi% select the smaller of both
  \ifdim\@tempa\p@<\p@\scalebox{\@tempa}{\usebox\pandoc@box}%
  \else\usebox{\pandoc@box}%
  \fi%
}
\def\fps@figure{htbp}
\makeatother
\providecommand{\tightlist}{%
  \setlength{\itemsep}{0pt}\setlength{\parskip}{0pt}}
\usepackage{bookmark}
\IfFileExists{xurl.sty}{\usepackage{xurl}}{} % add URL line breaks if available
\makeatletter
\@ifundefined{xmpquote}{}{}
\makeatother
\hypersetup{
  pdftitle={The Giant Hippocampus: From Structural Monoculture to a System of Systems},
  pdfauthor={Jaeho Seol (Independent Researcher, jaehoseol@gmail.com)},
  hidelinks,
  pdfcreator={LaTeX via pandoc}}

\title{The Giant Hippocampus: From Structural Monoculture to a System of
Systems}
\author{Jaeho Seol (Independent Researcher, jaehoseol@gmail.com)}
\date{}

\begin{document}
\maketitle

{
\setcounter{tocdepth}{3}
\tableofcontents
}
\subsection{Abstract}\label{abstract}

Ask a neuroscientist what the cortex looks like and they will describe a
mosaic: primary visual cortex crammed with a dense Layer 4 for spatial
encoding, motion-selective cortex built around thick Layers 5 and 6 for
temporal integration and long-range projection --- different problems
solved by different structures. Ask an AI researcher what a
state-of-the-art model looks like and they will describe one thing,
repeated at scale: the Transformer, wired identically whether it is
parsing text, classifying pixels, or transcribing speech. This paper
argues that gap is not a stylistic difference between two fields. It is
a structural error, and it is measurable.

The argument proceeds in three steps. First, a century of quantitative
cytoarchitecture, from Brodmann's original cortical parcellation through
the probabilistic Julich-Brain atlas to single-cell Patch-seq
recordings, establishes that distinct cognitive functions are
implemented by qualitatively different structural organizations, not by
rescaling one generic template. The convolutional neural network is the
field's own proof of this: it reached strong image-recognition
performance on a fraction of the data later architectures required,
precisely because its local receptive fields and hierarchical depth
encoded a structural prior instead of learning one from scratch. Second,
the paper traces how that lesson was discarded. The ``Hardware Lottery''
--- the economic dominance of GPUs optimized for dense matrix
multiplication --- made the Transformer the path of least resistance
rather than the principled choice, and the field mistook convenience for
a discovery about intelligence. Mixture-of-Experts routing, often cited
as evidence of architectural diversity, in fact partitions parameters
among structurally identical experts; it is quantitative specialization,
not qualitative. Third, a functionalist analysis grounded in multiple
realizability and modularity shows what the Transformer actually is: not
a general-purpose cortex, but a close functional analog of the
hippocampal formation, a system built for relational binding and
episodic, sequential prediction. Deploying it everywhere is the same
mistake AI made when it treated the cortex as one giant Broca's area ---
except now the whole field has standardized on a giant hippocampus, and
applied it to every task a hippocampus was never built to do:
spectro-temporal audition, reward-gated executive control, working
memory, multisensory binding.

Naming the error is not the contribution; fixing it is. The paper closes
with a concrete, biologically grounded alternative --- a Heterogeneous
Topological Network, a System of Systems in which structurally distinct
modules (a convolutional visual pathway, a spectro-temporal auditory
module, a Transformer-based episodic memory core, a
basal-ganglia-inspired executive gate, and a working-memory buffer) each
keep the domain-appropriate inductive bias their computation demands,
and communicate through standardized interfaces rather than a shared
substrate. This is a design discipline for AI architects, not a research
program for cognitive scientists: it asks that modularity constraints
and functional decompositions be specified before training begins, using
structural evidence about what different computations require as a
design input, rather than treating architecture as an afterthought to be
reverse-engineered from a monolith's behavior once it is already
trained.

\begin{center}\rule{0.5\linewidth}{0.5pt}\end{center}

\section{Part I: The Epistemological Blind
Spot}\label{part-i-the-epistemological-blind-spot}

\section{Chapter 1. Why Current Methods Cannot Find What They Are
Looking
For}\label{chapter-1.-why-current-methods-cannot-find-what-they-are-looking-for}

\subsubsection{1.1. The MOS 6502 Lesson: What Connectomics Cannot
Recover}\label{the-mos-6502-lesson-what-connectomics-cannot-recover}

A complete wiring diagram reveals which parts of a system can influence
each other. It does not reveal what the system computes. That gap
between mapping a structure and explaining its function grounds Part I,
and the clearest demonstration of it comes not from the brain but from a
chip.

\pandocbounded{\includegraphics[keepaspectratio,alt={Marr's tri-level hierarchy applied to microprocessor and AI system analysis. The diagram illustrates the failure to bridge implementation-level structure and computational-level function, even with perfect connectivity data. It maps the gap between (top) the computational task of object recognition, (middle) algorithmic processes and representations, and (bottom) the implementational substrate of silicon circuitry. Left: For the MOS 6502, complete transistor-level connectivity and standard neuroscience analyses (lesion maps, tuning curves, Granger causality) fail to recover the chip's native instruction-set architecture and computation. Right: The same analytical challenge scales to large language models, where the distance between implementational data and computational-level explanation is exponentially greater.}]{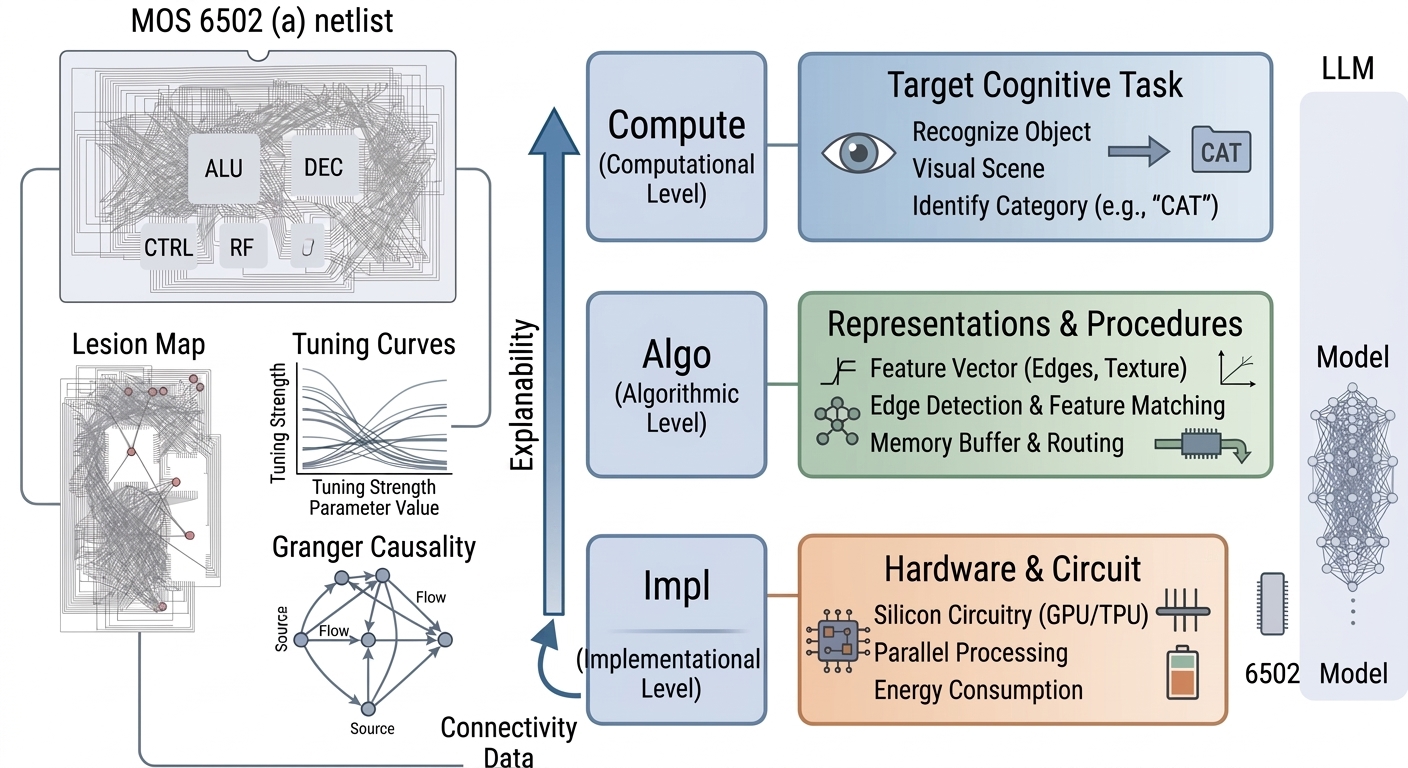}}
\emph{Figure 1. Marr's tri-level hierarchy applied to microprocessor and
AI system analysis. The diagram illustrates the failure to bridge
implementation-level structure and computational-level function, even
with perfect connectivity data. It maps the gap between (top) the
computational task of object recognition, (middle) algorithmic processes
and representations, and (bottom) the implementational substrate of
silicon circuitry. Left: For the MOS 6502, complete transistor-level
connectivity and standard neuroscience analyses (lesion maps, tuning
curves, Granger causality) fail to recover the chip's native
instruction-set architecture and computation. Right: The same analytical
challenge scales to large language models, where the distance between
implementational data and computational-level explanation is
exponentially greater.}

In 2017, Eric Jonas and Konrad Kording asked whether the standard
analytical toolkit of neuroscience could recover the workings of a
system that engineers already understand completely {[}Jonas and
Kording, 2017{]}. Their subject was the MOS 6502, the 8-bit
microprocessor behind the Apple II, the Commodore 64, and the original
Nintendo Entertainment System. Unlike cortex, the 6502 hides nothing.
The Visual6502 project had already reverse-engineered its full
transistor-level netlist, roughly 3,510 transistors, with every
connection and timing relation verified in simulation {[}James et al.,
2010{]}. This is the ideal that connectivity-based neuroscience can only
approximate: zero measurement noise, total coverage, and exact ground
truth.

The methods failed anyway. Jonas and Kording ran three programs,
\emph{Donkey Kong}, \emph{Space Invaders}, and \emph{Pitfall}, and
applied tools used across thousands of neuroscience papers {[}Jonas and
Kording, 2017{]}. Lesion analysis flagged transistors whose removal
halted one game while sparing another, a result that looks exactly like
finding a ``Donkey Kong neuron.'' The inference is false: those
transistors run generic register or decode operations that one program
happens to exercise and another does not. Tuning-curve analysis produced
apparent selectivity indistinguishable from single-unit recordings in
sensory cortex. Granger causality mapped information flow between
transistors yet never recovered the chip's actual functional blocks, the
arithmetic logic unit, the instruction decoder, and the register file.
Every pattern was statistically real. None was computationally
informative.

Marr framed why this happens. He argued that an information-processing
system demands explanation at three levels: the computational task it
solves, the algorithm and representations that solve it, and the
physical implementation {[}Marr, 1982{]}. Exhaustive implementational
data does not, on its own, climb to the higher two levels. Mechanistic
understanding requires identifying organized components and the
operations they perform, not merely cataloging dependencies {[}Craver,
2007{]}. Description is not explanation.

This limit is not peculiar to silicon. \emph{C. elegans} has had a
complete cellular connectome since 1986, yet that map alone never
delivered a full account of how the worm converts sensation into
behavior {[}White et al., 1986{]}. The graph constrains hypotheses; it
does not fix computation. The deficit then compounds with scale: the
6502 carries about 3,510 transistors, whereas a model such as GPT-3
carries roughly 175 billion parameters {[}Brown et al., 2020{]}, eight
orders of magnitude more. If complete knowledge of the smaller system
resists interpretation, adding that much complexity cannot rescue it.

The lesson is constructive, not merely skeptical. What the 6502 study
lacks is not data but a prior theory of which structures perform which
computational roles. Modern AI interpretability, including attention
maps, probing classifiers, and activation patching, repeats the same
move by draping correlation maps over systems whose architectural priors
were never specified. The missing ingredient is a structural theory of
function. For biological systems, that theory has a name:
cytoarchitecture. The next section sharpens the diagnosis by examining
how correlation-based network analysis and Granger causality describe
temporal dependency while still failing to recover the computation that
produces it.

\begin{center}\rule{0.5\linewidth}{0.5pt}\end{center}

\subsubsection{1.2. Network Analysis and Granger Causality: Correlation
Without
Computation}\label{network-analysis-and-granger-causality-correlation-without-computation}

Network neuroscience gives brain science a powerful descriptive
language. Graph measures such as degree, clustering, path length,
modularity, hubs, and small-world organization allow researchers to
compare structural and functional organization across regions, subjects,
and conditions {[}Bullmore \& Sporns, 2009{]}. These tools matter
because cognition rarely belongs to one isolated location. Large-scale
organization constrains what a circuit can do, and network measures can
identify candidate systems that deserve closer mechanistic study. The
limitation is conceptual rather than technical: a graph describes
relations among nodes, but it does not specify the algorithm those nodes
execute. A hub may support routing, gain control, memory access, task
switching, or some combination of these operations. Topology narrows the
search space for explanation; it does not select the computational
interpretation by itself.

This distinction becomes sharper in fMRI-based functional connectivity.
BOLD imaging measures a vascular signal related to neural activity, not
spikes, synaptic currents, or local circuit transformations directly
{[}Logothetis, 2008{]}. Hemodynamic delay, regional differences in
vascular response, physiological noise, and the mismatch between neural
timing and imaging acquisition all constrain what can be inferred from
correlated time series {[}Logothetis, 2008{]}. A reliable correlation
between two regions can therefore mark a real relation that needs
explanation, but the correlation does not reveal the variables
represented, the transformation applied to those variables, or the
control structure that organizes the computation.

Granger causality improves on simple correlation by adding temporal
prediction. In Granger's original formulation, one signal has causal
relevance for another when its past values improve prediction beyond
what the target signal's own past already provides {[}Granger, 1969{]}.
Neuroscience applications often implement this idea with time-series
models, but the core principle is predictive precedence, not direct
physical causation {[}Seth et al., 2015{]}. This distinction matters. A
Granger-significant relation can arise from a direct interaction, but it
can also arise from an unobserved common driver, an indirect pathway,
feedback dynamics, preprocessing choices, measurement noise, or
sampling-rate differences {[}Seth et al., 2015{]}. Granger analysis can
therefore generate directed hypotheses about influence, but it cannot by
itself identify the mechanism that performs a cognitive operation.

Marr's three-level framework clarifies why these methods stop short of
computation. Statistical dependence describes observed association.
Predictive precedence adds temporal structure. Mechanistic explanation
requires organized components and operations. Computational explanation
asks what problem the system solves and why that problem has that form
{[}Marr, 1982{]}. The MOS 6502 case made this boundary concrete. The
processor contains 3,510 transistors, which is small enough for
exhaustive transistor-level simulation and complete access to the
system's physical states {[}James et al., 2010{]}. Yet when standard
neuroscience-style analyses were applied to that fully known machine,
they recovered correlations, lesion effects, and tuning-like patterns
without reconstructing the program-level logic of fetch, decode, and
execute cycles {[}Jonas \& Kording, 2017{]}. Complete observability did
not guarantee computational understanding.

The same category error can reappear in AI. GPT-3's 175 billion
parameters and benchmark performance curves demonstrate the engineering
power of large-scale training, but parameter count and behavioral
performance do not explain which internal mechanisms produce which
capability {[}Brown et al., 2020{]}. The lesson is not that network
analysis, Granger causality, or scaling studies are useless. The lesson
is that dependency structure becomes misleading when treated as
computational explanation. The next section therefore turns from
analytic method to architectural assumption: the ``Giant Broca's Area''
fallacy, in which intelligence is flattened into one undifferentiated
model instead of being understood as a system of structurally
specialized computations.

\pandocbounded{\includegraphics[keepaspectratio,alt={A hierarchy of explanatory levels applied to network neuroscience, distinguishing statistical description from mechanistic and computational explanation. (a) Graph theoretical analysis of complex brain networks identifies topological features like hubs and modules, but does not reveal the underlying algorithms. (b) fMRI functional connectivity measures correlated BOLD signals which are hemodynamically delayed and filtered, capturing only a proxy for neural activity. (c) Granger causality attempts to infer directed influence through predictive precedence, but these estimates are vulnerable to confounds from common drivers, indirect paths, and noise. (d) Marr's tripartite framework clarifies the distinction between observed dependence (bottom), mechanism (middle), and the overarching computation (top), showing how network and causal analyses provide data that remains to be explained computationally.}]{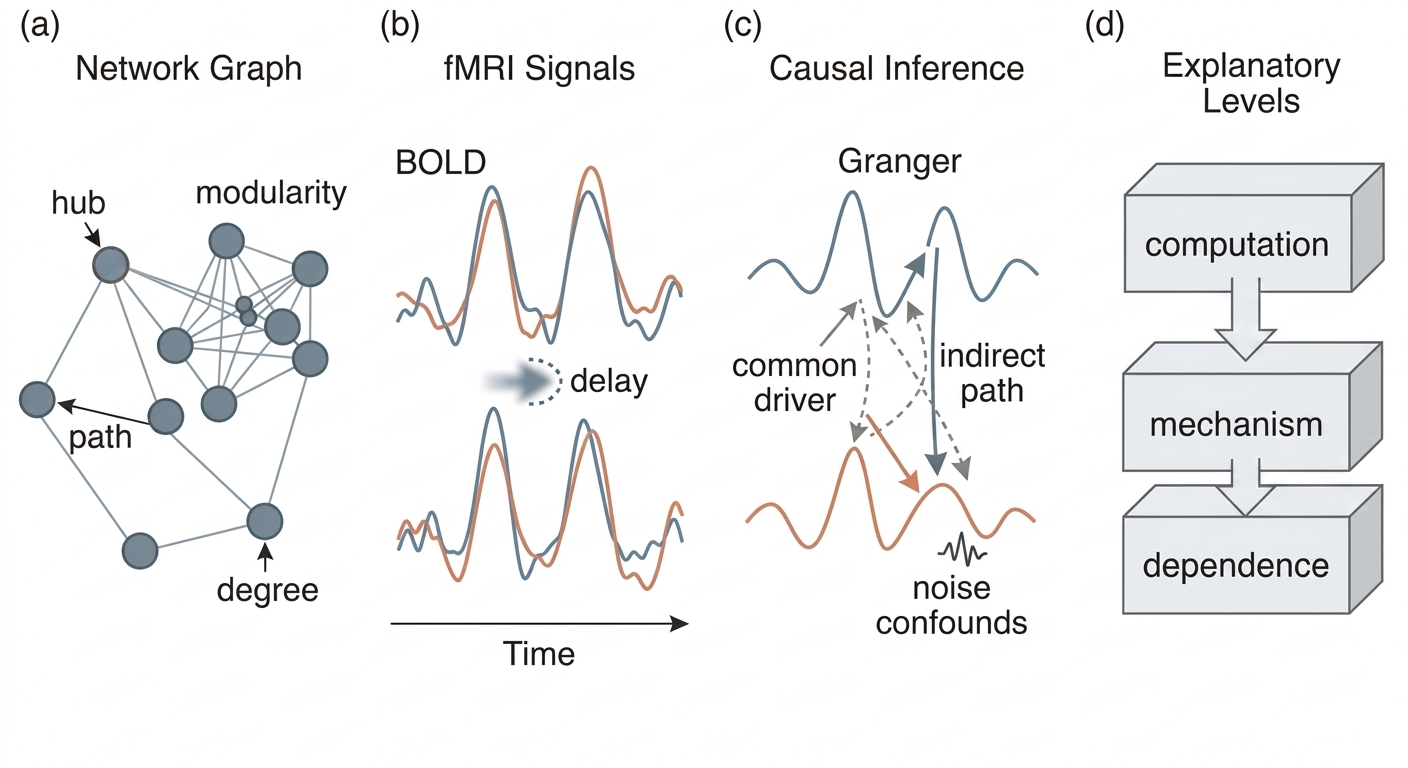}}
\emph{Figure 2. A hierarchy of explanatory levels applied to network
neuroscience, distinguishing statistical description from mechanistic
and computational explanation. (a) Graph theoretical analysis of complex
brain networks identifies topological features like hubs and modules,
but does not reveal the underlying algorithms. (b) fMRI functional
connectivity measures correlated BOLD signals which are hemodynamically
delayed and filtered, capturing only a proxy for neural activity. (c)
Granger causality attempts to infer directed influence through
predictive precedence, but these estimates are vulnerable to confounds
from common drivers, indirect paths, and noise. (d) Marr's tripartite
framework clarifies the distinction between observed dependence
(bottom), mechanism (middle), and the overarching computation (top),
showing how network and causal analyses provide data that remains to be
explained computationally.}

\begin{center}\rule{0.5\linewidth}{0.5pt}\end{center}

\subsubsection{1.3. The ``Giant Broca's Area'' Fallacy: AI's Homogenized
Model of the
Cortex}\label{the-giant-brocas-area-fallacy-ais-homogenized-model-of-the-cortex}

The failure of current analytic methods is not only methodological. It
also reflects a deeper architectural assumption: intelligence can be
treated as if it were produced by one large, undifferentiated processor.
The ``Giant Broca's Area'' fallacy names this error. It is the
temptation to take a successful localization story---a region, module,
or activation pattern associated with some cognitive function---and
inflate it into a general theory of computation. Localization may
identify where a function is supported, but it does not by itself
explain what structural organization makes that function possible.

Cytoarchitectonic evidence blocks the homogenized interpretation at the
outset. Cortical areas are not interchangeable copies of a single
generic cortical unit. They differ in cellular architecture, and those
differences matter for functional interpretation {[}Zilles, 2010{]}.
This point does not require the stronger claim that a cortical map alone
explains computation. It supports a narrower but decisive claim:
biological cognition is organized through structurally differentiated
tissue, so any account of intelligence that ignores structural
differentiation begins with the wrong abstraction. A cortex cannot be
understood as a giant version of any one localized area, because the
biological premise of cortical organization is regional differentiation
rather than uniform replication.

Contemporary AI has moved in the opposite direction. The Transformer
began as a sequence model for language, but its basic design has been
reused across domains with strikingly little architectural
differentiation. GPT-3 scaled the language-model version of this
paradigm to 175 billion parameters and 96 layers, demonstrating the
power of large Transformer systems for text prediction and few-shot task
performance {[}Brown, 2020{]}. Vision Transformer then applied the same
broad template to images by dividing a 224 × 224 image into 196 patches
of size 16 × 16 pixels and processing those patches as a sequence
{[}Dosovitskiy, 2021{]}. Audio Spectrogram Transformer extended the same
family of methods to audio classification by applying a
Vision-Transformer-style encoder to fixed-size spectrogram patches
{[}Gong, 2021{]}. These systems differ in input formatting and training
regime, but their shared premise is clear: pixels, words, and
spectrogram regions can all be made tractable by converting them into
token-like units for a common attention-based architecture.

That engineering move should not be mistaken for a discovery that
cognition itself has a universal substrate. Cross-modal reuse of the
Transformer shows architectural convergence inside AI, not proof that
language, vision, and audition require the same internal structure. The
distinction matters because the input conversion step can hide the loss
of domain-specific inductive bias. An image patch is not a word; a
spectrogram patch is not a cortical auditory computation; a text token
is not a general unit of cognition. Treating all of them as sequence
elements may produce useful models, but usefulness under a shared
computational interface does not establish that the interface captures
the structure of the underlying domain.

The ``Giant Broca's Area'' fallacy therefore has two sides. In
neuroscience, it appears when localization is confused with
computational explanation. In AI, it appears when one successful
architecture is generalized across domains until architectural sameness
begins to look like a theory of intelligence. Chapter 1 has shown why
this is epistemologically dangerous: methods that recover correlations,
graphs, or input-output success can still miss the structural logic that
makes cognition possible. The next chapter turns from this negative
diagnosis to the positive biological evidence: a century of
cytoarchitecture showing that form is not incidental to function, but
one of its governing constraints.

\begin{center}\rule{0.5\linewidth}{0.5pt}\end{center}

\section{Chapter 2. Form Dictates Function: The Cytoarchitectural
Evidence}\label{chapter-2.-form-dictates-function-the-cytoarchitectural-evidence}

\subsubsection{2.1. From Brodmann to Jülich: A Century of Structural
Proof}\label{from-brodmann-to-juxfclich-a-century-of-structural-proof}

Modern cytoarchitecture begins with a simple but disruptive observation:
the cortex does not present the same cellular organization everywhere.
Brodmann's early twentieth-century cortical map divided the cerebral
cortex into areas according to differences in cellular composition,
density, and laminar arrangement, rather than by gross anatomical
landmarks alone {[}Brodmann, 1909{]}. The enduring importance of that
map does not lie in the exact permanence of every boundary. Later work
revised many details. Its importance lies in the principle it
established: cortical areas can be identified by reproducible structural
differences, and those differences matter for understanding brain
function.

The Brodmann tradition therefore challenges the idea of a uniform
cortex. If the cortex were merely one repeated processing sheet whose
regions differed only by input and output, then cellular architecture
would play a secondary role. Cytoarchitectonic mapping shows the
opposite. Regional differences in lamination, cell-body density, and
areal boundaries appear systematically enough to support anatomical
classification. Zilles and Amunts situate Brodmann's map within a longer
historical arc, showing how the original parcellation became the
starting point for increasingly rigorous methods of cortical mapping
rather than a closed historical artifact {[}Zilles and Amunts, 2010{]}.
The key continuity is methodological: structure provides evidence for
regional differentiation, while each generation improves how that
evidence is measured, registered, and compared.

The Jülich approach marks a major step in that progression. Contemporary
Jülich-Brain mapping uses observer-independent cytoarchitectonic
analysis and probabilistic three-dimensional representations to describe
cortical and subcortical areas in stereotaxic space {[}Amunts et al.,
2020{]}. This shift matters because it turns cytoarchitecture from a
descriptive atlas tradition into a reproducible quantitative framework.
A cortical area is no longer treated simply as a named patch on a
diagram. It becomes a statistically represented anatomical entity whose
boundaries can vary across individuals yet still show enough regularity
to support comparison. That probabilistic framing is crucial: biological
structure is not rigid like a circuit board, but it is also not
arbitrary. It displays patterned variation.

\pandocbounded{\includegraphics[keepaspectratio,alt={Regional differentiation and structural variation across the cerebral cortex. (a) Evolution of cortical atlasing from Brodmann's historical qualitative parcellation to modern Jülich-Brain probabilistic maps, illustrating that borders reflect reproducible cytoarchitectonic differences; (b) laminar profiles (L1--L6) contrasting a sensory region with a prominent granular Layer 4 against a motor/association area dominated by deep output layers; (c) diagrams showing that while sharing a canonical columnar motif, sensory and motor regions have divergent thalamocortical and corticocortical projection patterns.}]{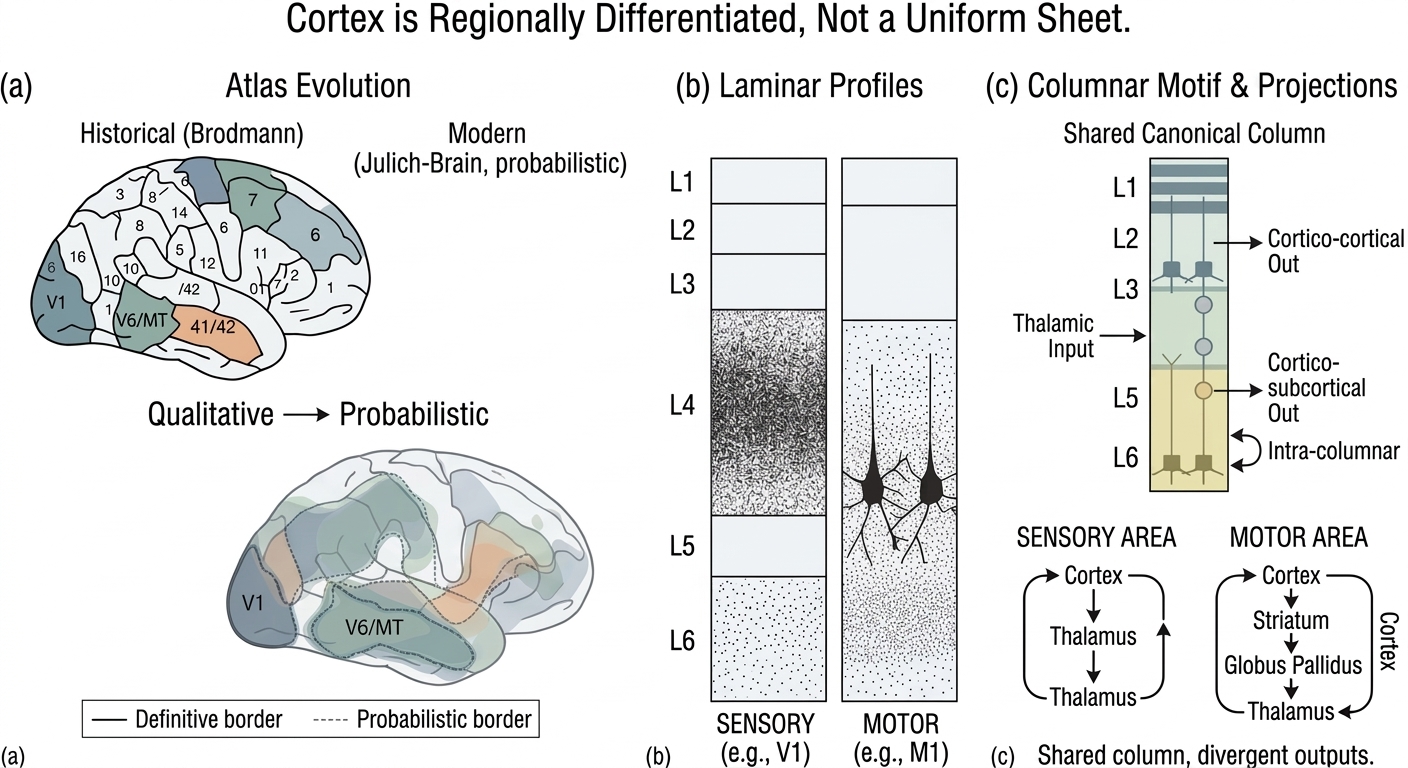}}
\emph{Figure 3. Regional differentiation and structural variation across
the cerebral cortex. (a) Evolution of cortical atlasing from Brodmann's
historical qualitative parcellation to modern Jülich-Brain probabilistic
maps, illustrating that borders reflect reproducible cytoarchitectonic
differences; (b) laminar profiles (L1--L6) contrasting a sensory region
with a prominent granular Layer 4 against a motor/association area
dominated by deep output layers; (c) diagrams showing that while sharing
a canonical columnar motif, sensory and motor regions have divergent
thalamocortical and corticocortical projection patterns.}

This evidence does not imply that structure alone fully specifies
computation. A cytoarchitectonic boundary does not directly reveal an
algorithm, just as a wiring diagram does not automatically explain a
machine's operation. The stronger and more defensible point is that
computation in biological cortex is constrained by material
organization. Laminar proportions, cell types, and regional architecture
shape what kinds of signals a region can receive, transform, maintain,
and project. Cytoarchitecture therefore supplies a structural prior for
functional interpretation: it narrows the space of plausible
computations without pretending to solve the entire problem by anatomy
alone.

The contrast with contemporary AI is instructive. Many leading systems
rely on repeated instances of the same Transformer-style attention
block, a mechanism originally defined as scaled dot-product attention
over token sequences {[}Vaswani et al., 2017{]}. GPT-3 scaled this
repeated decoder topology to 175 billion parameters {[}Brown et al.,
2020{]}. Vision Transformer and Audio Spectrogram Transformer extended
the same general attention mechanism to image patches and spectrogram
patches, respectively {[}Dosovitskiy et al., 2021; Gong et al., 2021{]}.
These systems differ in input formatting and scale, but their core
architectural pattern remains strikingly homogeneous. The
cytoarchitectural record suggests that such homogeneity should not be
mistaken for a natural endpoint of intelligent design. Biology's most
successful cognitive substrate did not solve diverse tasks by endlessly
repeating one undifferentiated block; it developed regionally distinct
structures whose forms constrain their functions. The next section moves
from areal mapping to finer evidence: laminar thickness, transcriptomic
identity, and functional specialization.

\begin{center}\rule{0.5\linewidth}{0.5pt}\end{center}

\subsubsection{2.2. Laminar Thickness, Transcriptomes, and Functional
Specialization}\label{laminar-thickness-transcriptomes-and-functional-specialization}

Cytoarchitecture does not treat cortical layers as decorative anatomy.
Laminar thickness, cell density, cellular morphology, and molecular
identity define the physical medium through which a cortical area
computes. Brodmann's early maps already made this point at a coarse
anatomical scale: different cortical territories show reproducible
differences in cellular layering rather than variations around one
uniform template {[}Brodmann, 1909{]}. Modern probabilistic
cytoarchitectonic mapping extends that insight with higher-resolution
tools, showing that cortical areas can be distinguished by measurable
differences in their cellular composition and laminar organization
{[}Zilles and Amunts, 2010; Amunts et al., 2020{]}.

This matters because laminar structure constrains information flow. A
cortical column does not merely contain neurons; it arranges different
neuronal populations into input, recurrent processing, and output
pathways. Canonical-circuit models identify recurring motifs in
neocortex, but they do not imply that every region implements the same
computation with interchangeable parts {[}Douglas and Martin, 2004{]}.
Harris and Shepherd describe a cortical microcircuit whose excitatory
and inhibitory components vary in ways that support different local
computations across cortical systems {[}Harris and Shepherd, 2015{]}. In
practical terms, a region with a dense granular input layer is not
structurally equivalent to a region dominated by deep projection layers.
That difference changes what signals the area receives, how long it can
integrate them, and where its outputs can be sent.

Transcriptomic evidence strengthens this structural argument without
inviting overclaim. Patch-seq does not, by itself, establish a complete
cross-area map of laminar specialization. Its contribution is more
precise: it allows single neurons to be jointly characterized by
morphology, electrophysiology, and gene expression, linking a cell's
shape, firing properties, and molecular profile in the same experimental
frame {[}Cadwell et al., 2016{]}. Integrated morphoelectric and
transcriptomic classification further shows that cortical cell types,
especially inhibitory interneuron classes, cannot be adequately
described by position alone {[}Gouwens et al., 2020{]}. At a broader
scale, transcriptomic atlases of the adult human brain show that
molecular organization varies systematically across brain regions,
confirming that anatomical differences are accompanied by differences in
gene-expression architecture {[}Hawrylycz et al., 2012{]}. The
defensible conclusion is not that every laminar transcriptomic rule has
already been fully mapped, but that cortical specialization appears
simultaneously at anatomical, physiological, and molecular levels.

This point also clarifies why output behavior and correlation alone
cannot settle the question of internal organization. Marr's levels of
analysis separate the computational goal of a system from the
algorithmic and implementational structures that realize it {[}Marr,
1982{]}. Craver's mechanistic account makes the same demand in
biological explanation: a satisfactory explanation identifies organized
components, their activities, and their causal arrangement {[}Craver,
2007{]}. Functional similarity therefore does not prove structural
equivalence. Correlational and causal-dependence tools can help identify
statistical relationships among signals, but they do not automatically
reveal the mechanism that produces those relationships; this limitation
is especially important in neuroscience applications of Granger-style
causal analysis {[}Seth et al., 2015{]}.

The contrast with contemporary AI is instructive. The Transformer
introduced scaled dot-product self-attention as a general mechanism for
relating tokens in a sequence {[}Vaswani et al., 2017{]}. GPT-3 scales
this design into a deep stack of repeated decoder blocks {[}Brown et
al., 2020{]}. Vision Transformer applies the same attention block to
image patches, and Audio Spectrogram Transformer applies it to
spectrogram patches {[}Dosovitskiy et al., 2021; Gong et al., 2021{]}.
These systems can be powerful, but their design strategy differs sharply
from cortical specialization: modality-specific structure is minimized
while a common computational block expands across domains.
Cytoarchitecture suggests the opposite lesson. Biological intelligence
preserves shared motifs, but it modifies tissue architecture when
functional demands change. The next section therefore turns directly
against the uniform-cortex assumption and treats regional divergence as
a biological necessity rather than anatomical noise.

\begin{center}\rule{0.5\linewidth}{0.5pt}\end{center}

\subsubsection{2.3. Against the Uniform Cortex: Regional Divergence as
Biological
Necessity}\label{against-the-uniform-cortex-regional-divergence-as-biological-necessity}

The strongest version of the uniform-cortex hypothesis does not claim
that every cortical area looks identical. It makes a subtler claim: that
the cortex is built from one broadly repeated computational template,
and that regional differences mostly reflect input, output, scale, or
training history. That view has an important historical basis.
Mountcastle's account of cortical columns framed the neocortex as a
structure with repeated local organization, and later canonical
microcircuit models showed why recurrent excitatory and inhibitory
motifs serve as useful abstractions for neocortical computation
{[}Mountcastle, 1997; Douglas and Martin, 2004{]}. The problem is not
that these abstractions are false. The problem is that they become
misleading when treated as a complete theory of cortical design.

The empirical record now supports a more constrained interpretation:
cortex has recurrent organizational themes, but those themes appear with
substantial regional variation in cell types, connectivity, laminar
proportions, and molecular identity. Harris and Shepherd describe this
relationship as ``cortical circuit themes and variations,'' a phrase
that captures the central point: the cortex reuses motifs, but it does
not implement cognition by copying one circuit everywhere and merely
changing its size {[}Harris and Shepherd, 2015{]}. The adult human brain
transcriptome reinforces this conclusion at the molecular level.
Hawrylycz and colleagues show anatomically patterned gene-expression
differences across human brain regions, indicating that regional
differentiation is written not only into gross anatomy but also into
molecular organization {[}Hawrylycz et al., 2012{]}. Patch-seq
strengthens the same methodological lesson at single-neuron resolution
by allowing morphology, electrophysiology, and transcriptomic profile to
be examined together rather than as isolated descriptors {[}Cadwell et
al., 2016{]}. Neuronal identity is therefore multidimensional: shape,
firing behavior, local circuit role, and gene-expression state jointly
define what a neuron can contribute to computation.

This evidence does not require the claim that every function is
genetically predetermined, or that experience plays no role. It supports
a narrower and more defensible claim: learning occurs inside
architectures that already impose structural constraints. A cortical
area does not begin as an abstract processor waiting to become visual,
auditory, motor, or executive through data alone. Its laminar
organization, cell-type composition, connectivity, and molecular profile
bias the kinds of transformations it can perform efficiently. Regional
divergence is biologically necessary in this sense: not as rigid
destiny, but as the material condition that makes specialized
computation possible.

A useful computational contrast appears in recent attempts to extend the
same attention-based architecture across sensory modalities. The Vision
Transformer converts an image into a sequence of fixed-size patches and
processes those patches with the same self-attention mechanism
originally designed for token sequences {[}Dosovitskiy et al., 2021{]}.
The Audio Spectrogram Transformer applies a similar strategy to
mel-spectrograms, turning time-frequency structure into spectrogram
patches that a Transformer stack can also consume {[}Gong et al.,
2021{]}. These models demonstrate the flexibility of attention over
tokenized inputs, but flexibility is not the same as a
domain-appropriate structural prior. Dosovitskiy and colleagues
explicitly report that Vision Transformers trained from scratch on
mid-sized image datasets underperform comparable convolutional networks,
and they attribute this disadvantage to the absence of convolutional
inductive biases such as locality and translation equivariance
{[}Dosovitskiy et al., 2021{]}. In plain terms, the model can learn
visual structure, but it must spend data to recover what a
better-matched architecture builds in from the start.

That comparison clarifies the biological argument without reducing
cortex to an engineering metaphor. The brain's answer to heterogeneous
computational demands is not a single processor stretched across
domains. It is a family of related but regionally differentiated
architectures whose structural biases make different forms of learning
and inference tractable. The next chapter therefore moves from the fact
of regional divergence to its mesoscale mechanisms: layers, dendrites,
and cellular identity as the physical blueprint through which form
constrains function.

\begin{center}\rule{0.5\linewidth}{0.5pt}\end{center}

\section{Part II: How AI Lost Its Structural
Priors}\label{part-ii-how-ai-lost-its-structural-priors}

\section{Chapter 3. The Mesoscale Blueprint: Layers, Dendrites, and
Cellular
Identity}\label{chapter-3.-the-mesoscale-blueprint-layers-dendrites-and-cellular-identity}

\subsubsection{3.1. V1 vs.~MT/V5: Why Laminar Proportions Are Functional
Specifications}\label{v1-vs.-mtv5-why-laminar-proportions-are-functional-specifications}

The contrast between primary visual cortex and MT/V5 shows why cortical
architecture cannot be treated as a uniform sheet enlarged or reduced
for different tasks. V1 occupies the earliest cortical stage of the
visual hierarchy, where the system must preserve fine spatial detail,
retinotopic order, and high-fidelity input structure. Lund's account of
macaque striate cortex describes a highly differentiated laminar
organization, including an elaborated granular Layer 4 divided into
specialized sublaminae that receive and redistribute geniculocortical
input {[}Lund, 1988{]}. That arrangement is not a generic increase in
neural tissue. It is a structural allocation in which cellular and
laminar machinery is organized around the computational demand of
precise early visual encoding.

\pandocbounded{\includegraphics[keepaspectratio,alt={Functional specialization of visual cortex architectures. (a) Schematic of primary visual cortex (V1) showing its highly differentiated laminar profile, including a greatly elaborated Layer 4 that receives geniculocortical input and preserves precise retinotopic order; (b) schematic of the motion-sensitive medial temporal area (MT/V5), illustrating its distinct organization specialized for integrating complex signals of direction, speed, and binocular disparity across space and time. The contrast between these two architectures highlights that laminar proportions and wiring are structural adaptations to specific computational demands.}]{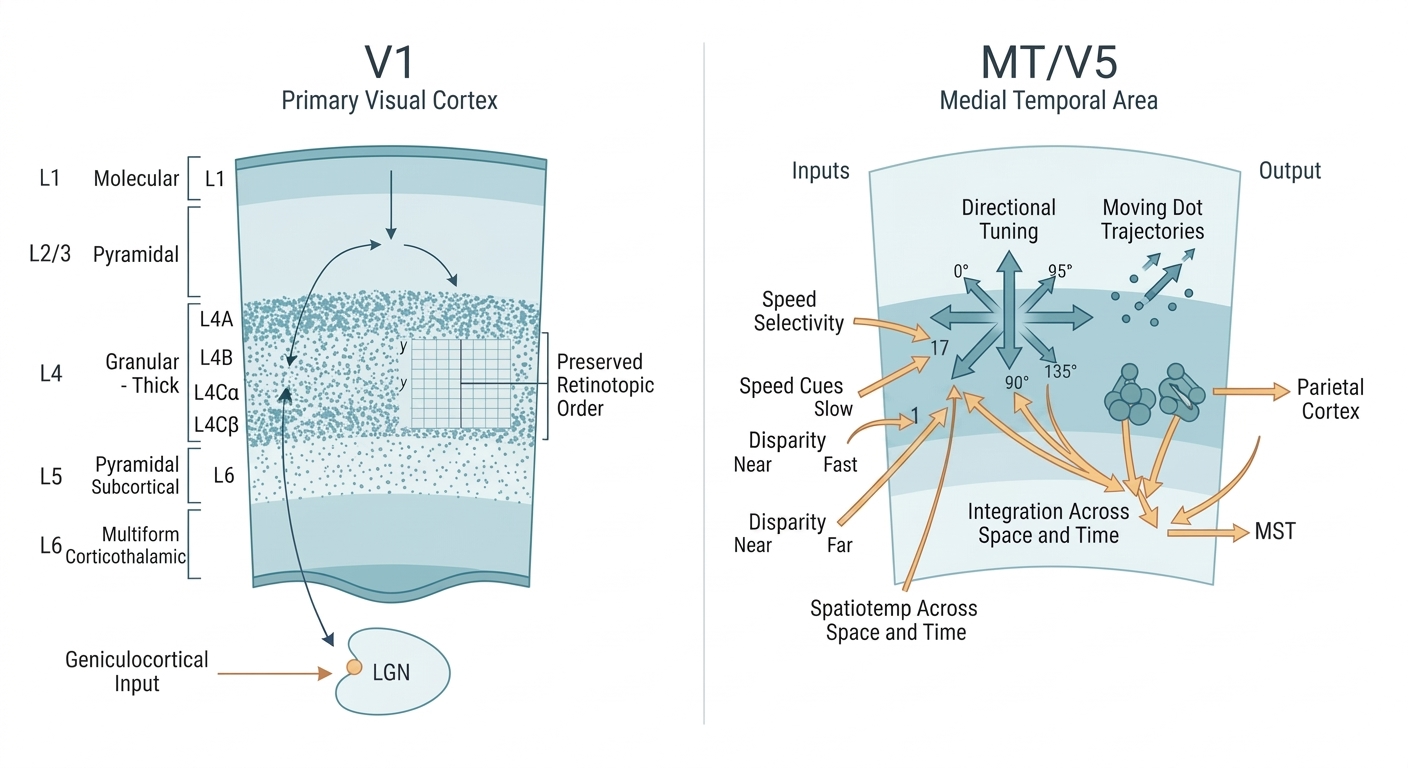}}
\emph{Figure 4. Functional specialization of visual cortex
architectures. (a) Schematic of primary visual cortex (V1) showing its
highly differentiated laminar profile, including a greatly elaborated
Layer 4 that receives geniculocortical input and preserves precise
retinotopic order; (b) schematic of the motion-sensitive medial temporal
area (MT/V5), illustrating its distinct organization specialized for
integrating complex signals of direction, speed, and binocular disparity
across space and time. The contrast between these two architectures
highlights that laminar proportions and wiring are structural
adaptations to specific computational demands.}

MT/V5 serves a different role in the dorsal visual stream. Born and
Bradley characterize MT as a motion-sensitive visual area with
direction-selective responses, retinotopic organization, and
connectivity suited to integrating signals relevant to motion, speed,
disparity, and visual dynamics {[}Born and Bradley, 2005{]}. The
defensible anatomical contrast is therefore specific: V1 exhibits an
elaborated granular input architecture, whereas MT/V5 organizes visual
information around motion-sensitive processing and dorsal-stream
coordination. This comparison does not depend on treating all
higher-order motion areas as if they shared a single laminar profile. It
is sufficient that the visual system assigns distinct mesoscale
organizations to distinct computational roles.

This difference matters because laminar proportions are functional
specifications, not decorative biology. A system that must preserve
local spatial precision needs dense input-sensitive organization and
stable topographic mapping. A system that must extract motion across
space and time needs circuitry that can integrate changing signals and
coordinate with other visual areas. V1 and MT/V5 therefore illustrate a
broader principle: biological cortex does not solve every perceptual
problem by applying one homogeneous template to different input streams.
It assigns different mesoscale organizations to different computational
demands.

Evidence from cellular identity reinforces the same point below the
level of cortical areas. Patch-seq demonstrates that a neuron's
electrophysiology, morphology, and transcriptomic profile can be
measured together, making neuronal identity a multidimensional
biological category rather than an interchangeable abstract unit
{[}Cadwell et al., 2016{]}. The adult human brain transcriptome atlas
likewise shows anatomically patterned gene-expression differences across
brain regions {[}Hawrylycz et al., 2012{]}. These findings do not merely
add molecular detail to a laminar story. They show that specialization
appears across linked scales: area-level architecture, laminar
organization, cell-type properties, and regional molecular profile.

The AI comparison sharpens the implication. Vision Transformers convert
images into patch sequences and process them with the same
self-attention machinery originally developed for token sequences,
explicitly reducing the convolutional inductive biases of locality and
translation equivariance that earlier vision systems built into their
structure {[}Dosovitskiy et al., 2021{]}. Audio Spectrogram Transformers
apply a similar patch-and-attention strategy to time-frequency
spectrograms, treating audio as another surface that can be tokenized
and passed through the same general scaffold {[}Gong et al., 2021{]}.
These models can perform strongly on benchmarks, so the point is not
failure. The point is architectural substitution. Where biology assigns
different structural priors to spatial vision, motion processing, and
other sensory demands, the homogeneous Transformer strategy often asks
data and scale to recover distinctions that architecture did not encode
in advance.

The V1-to-MT/V5 comparison therefore supplies the mesoscale version of
the paper's central argument: architecture is not a neutral container
for learning. It is a prior commitment about which computations a system
can perform efficiently, reliably, and with appropriate domain
structure. The next section shifts from laminar proportions to dendritic
geometry, showing how receptive-field architecture continues this same
principle below the level of cortical areas.

\begin{center}\rule{0.5\linewidth}{0.5pt}\end{center}

\subsubsection{3.2. Dendritic Complexity as Receptive Field
Architecture}\label{dendritic-complexity-as-receptive-field-architecture}

A cortical pyramidal neuron is not a biological version of the
artificial ``neuron'' used in standard deep learning. In the simplest
artificial abstraction, a unit receives a vector of inputs, computes a
weighted combination, and passes the result through a pointwise
nonlinearity. That abstraction has been enormously useful for
engineering, but it removes the internal spatial structure of the
biological cell. Real cortical neurons receive input across dendritic
trees whose branches differ in distance from the soma, local
conductances, synaptic clustering, and susceptibility to active
electrical events. Dendrites can generate local nonlinear responses,
including NMDA-dependent events, calcium spikes, and backpropagating
action-potential interactions, so the cell does not merely sum all
incoming signals at one central point {[}London and Häusser, 2005;
Stuart and Spruston, 2015{]}.

\pandocbounded{\includegraphics[keepaspectratio,alt={Dendritic computation versus artificial neuron and Transformer block abstractions. (a) The standard artificial abstraction reduces input to a single weighted sum and pointwise nonlinearity. (b) In contrast, a biological pyramidal neuron features complex dendritic morphology where inputs are clustered across apical and basal branches, triggering local nonlinear events and basal-branch spikes before integrating at the soma. (c) This demonstrates that biological complexity is located within a single unit, whereas Transformer complexity arises from layered interactions between units across token representations.}]{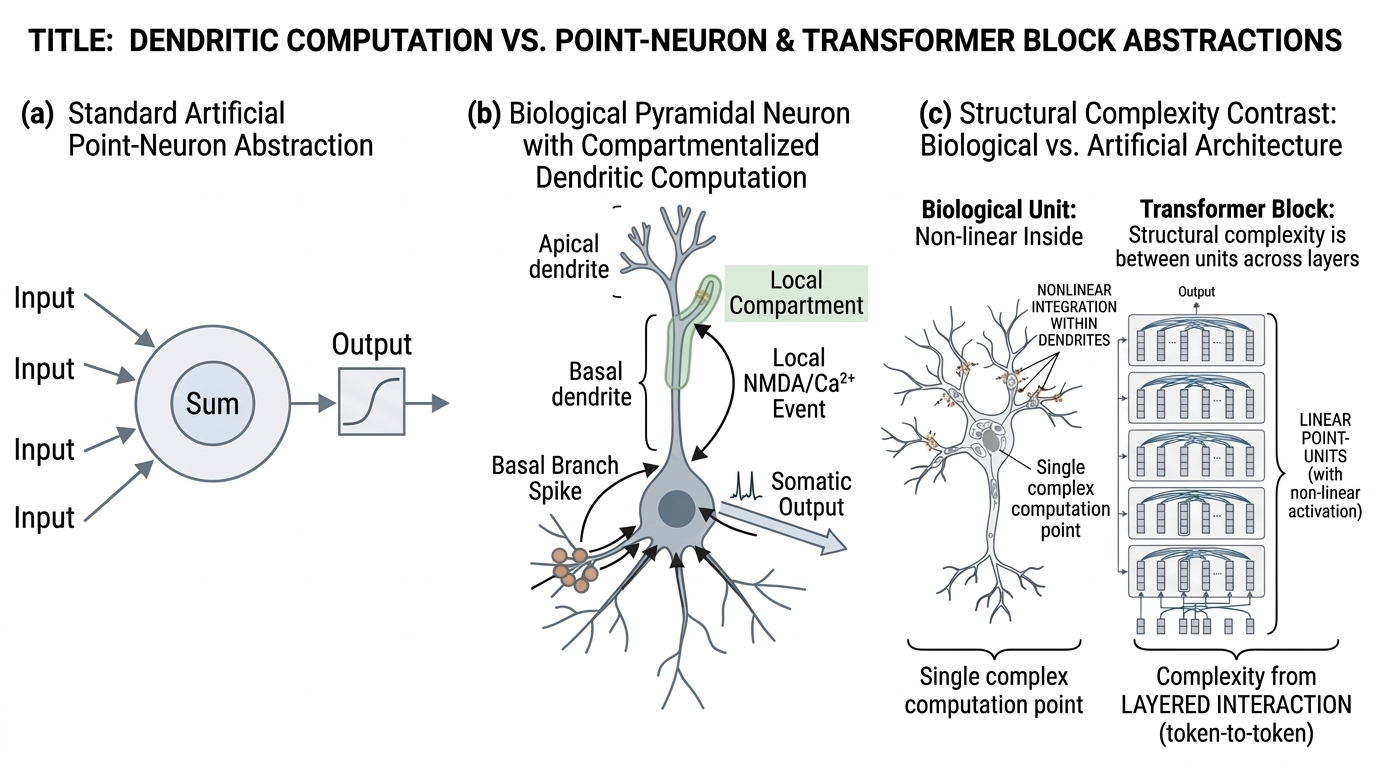}}
\emph{Figure 5. Dendritic computation versus artificial neuron and
Transformer block abstractions. (a) The standard artificial abstraction
reduces input to a single weighted sum and pointwise nonlinearity. (b)
In contrast, a biological pyramidal neuron features complex dendritic
morphology where inputs are clustered across apical and basal branches,
triggering local nonlinear events and basal-branch spikes before
integrating at the soma. (c) This demonstrates that biological
complexity is located }within* a single unit, whereas Transformer
complexity arises from layered interactions \emph{between} units across
token representations.*

This matters because dendritic morphology changes the effective
receptive field of the neuron. A receptive field is often described as
the pattern of input that drives a cell, but in dendritic terms it is
also the pattern that the cell's own internal geometry can distinguish.
Synapses that arrive close together on one branch can interact
differently from synapses distributed across distant branches. Inputs
arriving on apical dendrites can remain partly segregated from inputs
arriving on basal dendrites before both influence somatic output. The
dendritic tree therefore functions as a structured integration space: it
partitions input, creates local thresholds, and allows different parts
of the same neuron to perform partially independent computations before
producing a cellular response {[}London and Häusser, 2005; Stuart and
Spruston, 2015{]}.

This is not a metaphorical claim that neurons are ``little networks.''
The stronger and more precise claim is that dendritic structure helps
define the computational boundary conditions under which synaptic input
becomes output. Evidence from macaque primary visual cortex supports
this point at the anatomical level. Callaway and Wiser showed that spiny
neurons in layers 2 through 5 differ in dendritic and axonal
organization in ways that shape their contribution to local cortical
circuitry {[}Callaway and Wiser, 1996{]}. Such differences are not
interchangeable decorative variations. They alter where signals can
arrive, how locally they can interact, and how a neuron participates in
the surrounding circuit. In visual cortex, receptive-field architecture
therefore begins below the scale of the cortical area and even below the
scale of the canonical circuit; it is partly written into the branching
structure of individual cells.

The contrast with contemporary large-scale AI should be drawn carefully.
Transformers are not unstructured systems. GPT-3, for example, uses a
deep stack of transformer blocks with attention and feedforward
components at very large scale, and the Vision Transformer converts
image patches into token sequences processed by repeated attention-based
layers {[}Brown et al., 2020; Dosovitskiy et al., 2021{]}. These
architectures implement rich network-level computation. Their
limitation, for the present argument, is more specific: they generally
place structural complexity between units rather than inside units. A
token representation in a transformer can interact with other tokens
through attention, but the unit-level abstraction does not contain a
dendrite-like spatial substrate in which inputs are locally clustered,
compartmentalized, and nonlinearly integrated before the unit
contributes to the next stage.

That asymmetry clarifies why cytoarchitecture matters for AI design.
Biological cortex does not rely only on adding more layers, more units,
or more global routing. It also distributes computation into the
physical organization of the cells themselves. Dendritic morphology
supplies an inductive bias: it makes some input relations easier to
detect, some interactions easier to segregate, and some forms of
integration available before circuit-level processing begins. The next
step is to ask how such structure can be tied to cell identity rather
than described only as morphology. Patch-seq addresses that evidentiary
layer by measuring morphology, electrophysiology, and transcriptomic
identity in individual neurons, linking cellular form to functional
phenotype without reducing either to a single feature {[}Cadwell et al.,
2016{]}.

\begin{center}\rule{0.5\linewidth}{0.5pt}\end{center}

\subsubsection{3.3 Patch-seq: Morphology, Electrophysiology, and
Transcriptomics in
One}\label{patch-seq-morphology-electrophysiology-and-transcriptomics-in-one}

Patch-seq strengthens the case against cortical uniformity because it
measures several dimensions of neuronal identity in the same cell.
Earlier approaches could compare neuronal shape, firing pattern, or
molecular profile, but they often treated those dimensions as separate
streams of evidence. Patch-seq combines whole-cell patch-clamp
recording, recovery of cellular RNA for transcriptomic profiling, and
morphological reconstruction, allowing a neuron to be described
simultaneously by the genes it expresses, the way it fires, and the
arrangement of its dendrites and axons {[}Cadwell et al., 2016{]}.
Figure 6 summarizes this workflow and its relevance to regional
specialization. This matters because cortical specialization does not
reside at one privileged level of description. A neuron is not merely a
point in a wiring diagram, a generic threshold unit, or an
interchangeable member of a large population. It is a structured
biological device whose molecular identity, membrane dynamics, and
physical geometry jointly constrain the computations it can perform.

\pandocbounded{\includegraphics[keepaspectratio,alt={The Patch-seq multimodal single-cell assay and its regional context. (a) An intracellular pipette patch-clamps a single cortical neuron, simultaneously extracting RNA and recording electrophysiological properties, followed by morphological reconstruction from the same cell; (b) transcriptomic classes (e.g., Pyr L2/3, Pyr L5, VIP, Sst, Pvalb) correspond to distinct electrophysiological and morphological phenotypes; (c) brain-wide transcriptomic organization varies systematically across regions, illustrating that molecular diversity accompanies functional and anatomical specialization at the atlas scale.}]{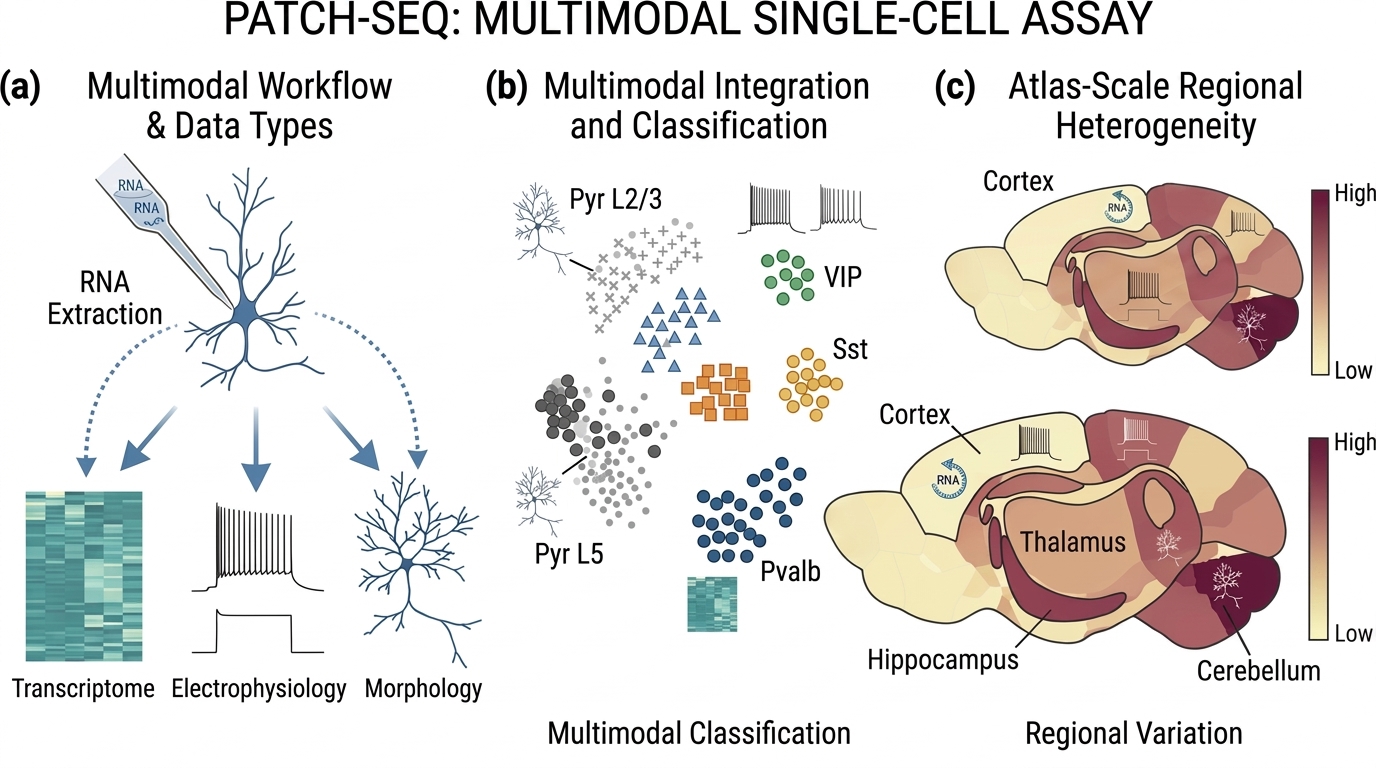}}
\emph{Figure 6. The Patch-seq multimodal single-cell assay and its
regional context. (a) An intracellular pipette patch-clamps a single
cortical neuron, simultaneously extracting RNA and recording
electrophysiological properties, followed by morphological
reconstruction from the same cell; (b) transcriptomic classes (e.g., Pyr
L2/3, Pyr L5, VIP, Sst, Pvalb) correspond to distinct
electrophysiological and morphological phenotypes; (c) brain-wide
transcriptomic organization varies systematically across regions,
illustrating that molecular diversity accompanies functional and
anatomical specialization at the atlas scale.}

Large-scale Patch-seq studies sharpen this point by showing that
transcriptomic classes align with distinctive electrophysiological and
morphological profiles rather than floating free of function {[}Gouwens
et al., 2020{]}. The result is not a simple one-to-one dictionary in
which every gene-expression cluster mechanically predicts a single
computational role. Biological categories remain noisy, overlapping, and
developmentally shaped. Yet the central architectural result remains
decisive: cellular identity is multidimensional, and those dimensions
covary in ways that affect computation. Brain-wide transcriptomic
atlases reinforce the same principle at a larger scale by showing that
human neural tissue expresses spatially organized molecular patterns
rather than a homogeneous genetic program repeated everywhere
{[}Hawrylycz et al., 2012{]}. Patch-seq brings that lesson down to the
level of the individual neuron, where molecular profile, electrical
behavior, and structure meet.

Dendrites make the architectural implication especially clear. A
dendritic tree does not simply collect inputs like wires feeding a
central processor. Dendrites contain active conductances, support local
nonlinear events, and shape how synaptic inputs combine before they
influence somatic output {[}London and Häusser, 2005; Stuart and
Spruston, 2015{]}. In practical terms, branching geometry helps define a
neuron's receptive-field architecture: which signals can interact
locally, which must compete across distance, and which patterns can be
amplified or suppressed before the cell emits spikes. Morphology
therefore cannot be treated as decorative biological detail. A neuron
with different dendritic depth, branching pattern, and channel
distribution implements a different input-integration regime, even if a
coarse network diagram assigns it the same node label.

The contrast with contemporary AI architecture is direct, but it
requires careful framing. Vision Transformers tokenize images into
patches and process them through a self-attention scaffold originally
developed for sequence modeling {[}Dosovitskiy et al., 2021{]}. Audio
Spectrogram Transformers apply a closely related strategy to
time-frequency representations of sound {[}Gong et al., 2021{]}. These
systems can be useful engineering artifacts, but their reuse of a common
attention topology across modalities expresses the opposite design
instinct from Patch-seq. Biology does not solve vision, audition,
executive control, and memory by copying one cell type or one circuit
motif at larger scale. It varies molecular identity, firing dynamics,
dendritic geometry, laminar position, and long-range projection pattern
together.

Patch-seq closes the mesoscale argument of this chapter. Layers matter
because they organize input and output flow; dendrites matter because
they define local integration; cellular identity matters because
morphology, electrophysiology, and transcriptomics form a coordinated
computational profile. If AI architecture aims to recover the efficiency
of biological intelligence rather than merely approximate its behavior
through scale, it must encode specialization into scaffold and topology.
The next chapter turns to the clearest historical case in which AI
briefly did exactly that: the convolutional neural network, whose
locality and hierarchy made visual structure part of the model rather
than something to be rediscovered from data alone.

\begin{center}\rule{0.5\linewidth}{0.5pt}\end{center}

\section{Chapter 4. CNN: The Last Architecture That Got Structure
Right}\label{chapter-4.-cnn-the-last-architecture-that-got-structure-right}

\subsubsection{4.1. Convolution as Visual Cortex: Spatial Locality and
Hierarchical
Depth}\label{convolution-as-visual-cortex-spatial-locality-and-hierarchical-depth}

The convolutional neural network became important not merely because it
added more layers to machine learning, but because it built assumptions
about the visual world directly into the model. Images are not arbitrary
lists of symbols. Neighboring pixels usually belong to neighboring parts
of the same object or surface, and useful visual features often recur
across different locations. A model designed for vision should therefore
treat local spatial neighborhoods as meaningful, reuse the same feature
detector across the image, and compose small features into larger ones.
Convolution does exactly this.

The biological relevance of this design choice begins with receptive
fields. Hubel and Wiesel showed that neurons in cat primary visual
cortex respond selectively to local visual patterns, including oriented
edges within restricted regions of visual space {[}Hubel and Wiesel,
1962{]}. Their work did not imply that cortex performs digital
convolution, and the argument here does not depend on that stronger
claim. The narrower and defensible point is that early visual processing
uses local detectors arranged across a spatial map. CNNs translate this
principle into engineering form: a convolutional filter examines a small
patch of an image, applies the same learned weights at many positions,
and produces a feature map that preserves spatial layout while detecting
the same pattern wherever it appears.

\pandocbounded{\includegraphics[keepaspectratio,alt={(a) Schematic illustrating the principles of biological retina-to-V1 visual processing, where local V1 receptive fields detect spatially restricted patterns; (b) and a corresponding computational framework in deep convolutional neural networks (CNNs), showing how shared weights enable the same filter to detect specific features across different spatial locations. This shared-weight architecture enforces the visual prior of translation invariance, while hierarchical depth allows for the progressive composition of features into increasingly complex representations, a concept culminating in modern large-scale models like ImageNet. The historical lineage traces this evolution from the Neocognitron and LeNet to modern deep CNNs.}]{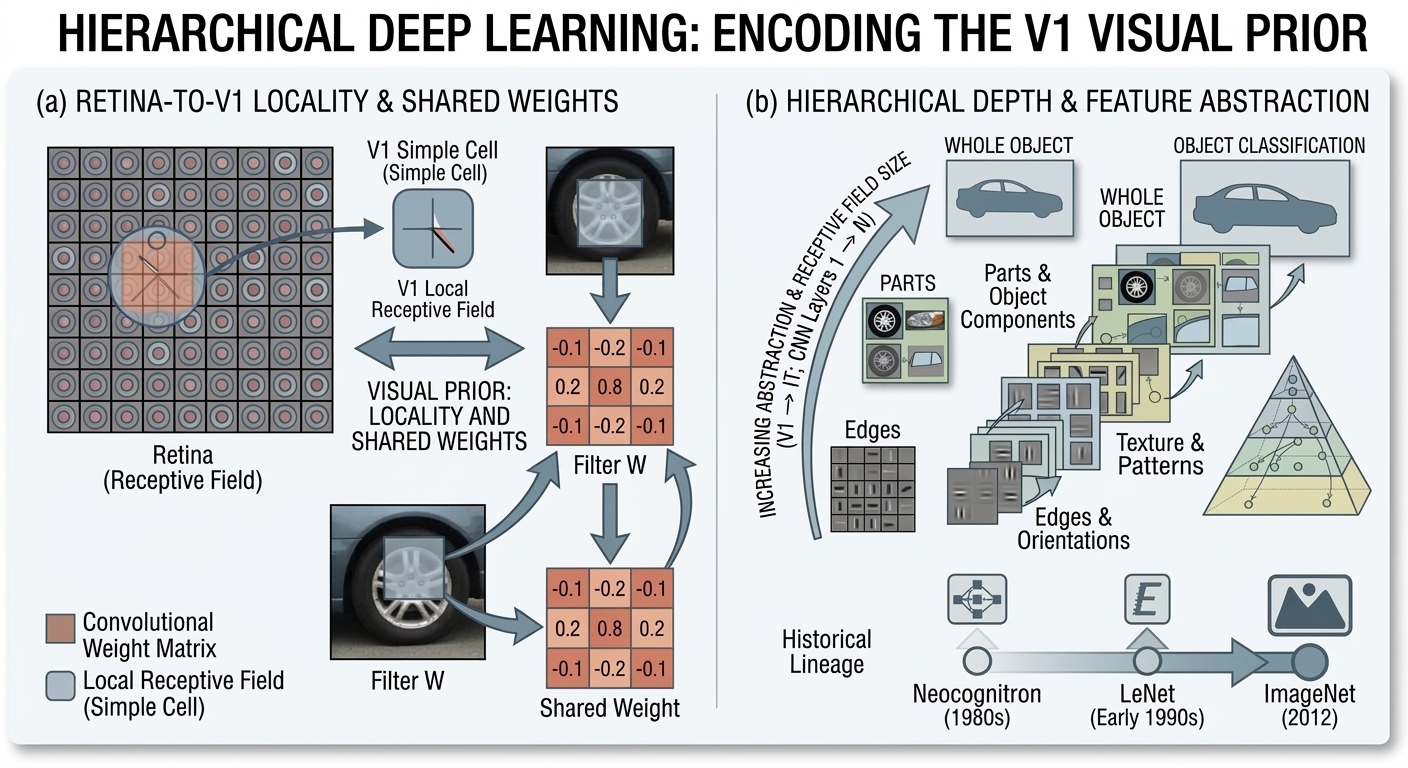}}
\emph{Figure 7. (a) Schematic illustrating the principles of biological
retina-to-V1 visual processing, where local V1 receptive fields detect
spatially restricted patterns; (b) and a corresponding computational
framework in deep convolutional neural networks (CNNs), showing how
shared weights enable the same filter to detect specific features across
different spatial locations. This shared-weight architecture enforces
the visual prior of translation invariance, while hierarchical depth
allows for the progressive composition of features into increasingly
complex representations, a concept culminating in modern large-scale
models like ImageNet. The historical lineage traces this evolution from
the Neocognitron and LeNet to modern deep CNNs.}

This locality becomes powerful only when stacked into hierarchy. Visual
cortex does not process a scene in one flat operation. Felleman and Van
Essen mapped primate visual cortex as a distributed hierarchy in which
information flows through many interconnected areas, supporting
progressively more complex visual representations {[}Felleman and Van
Essen, 1991{]}. CNNs again capture a structural principle rather than a
literal biological mechanism. Early layers tend to detect simple
patterns such as edges or textures; deeper layers combine those signals
into parts, shapes, and object-level configurations. The model's depth
therefore encodes a constraint: visual recognition should proceed
through staged composition rather than through a single undifferentiated
mapping from pixels to labels.

The engineering lineage makes this connection explicit. Fukushima's
neocognitron introduced a layered recognition architecture with local
feature extraction and increasing tolerance to shifts in position
{[}Fukushima, 1980{]}. LeCun and colleagues then showed how
convolutional architectures could be trained end-to-end for document
recognition, using local receptive fields, shared weights, and
subsampling to reduce the number of free parameters while preserving
visual structure {[}LeCun et al., 1998{]}. These choices were not
cosmetic. Weight sharing means that the model does not need to relearn
the same edge or stroke detector separately at every pixel location.
Local connectivity means that the model does not waste capacity by
treating every pixel as equally related to every other pixel from the
first layer. Hierarchical depth means that complex recognition emerges
through composition rather than through brute-force memorization of
whole-image patterns.

The 2012 ImageNet result demonstrated how far this structural prior
could scale. Krizhevsky, Sutskever, and Hinton's deep CNN achieved a
top-5 error rate of 15.3\%, far ahead of the next-best entry at 26.2\%,
and helped establish deep convolutional models as the dominant
visual-recognition architecture of that period {[}Krizhevsky et al.,
2012{]}. That success matters here because the model's advantage did not
come from architectural neutrality. It came from a strong,
domain-specific bias: images have local structure, repeated features,
and hierarchical organization.

CNNs therefore provide a concrete counterexample to the idea that
intelligence advances only by enlarging a general-purpose architecture.
They show that the right structural prior can make a system more aligned
with the problem it must solve. Section 4.2 turns from this
architectural correspondence to the efficiency consequence: how much
work structural prior knowledge can save when compared with data-hungry
generalization.

\begin{center}\rule{0.5\linewidth}{0.5pt}\end{center}

\subsubsection{4.2. Structural Prior vs.~Data Volume: The Efficiency
Argument}\label{structural-prior-vs.-data-volume-the-efficiency-argument}

The efficiency of convolutional neural networks did not come from a
vague resemblance to the brain. It came from a narrower and more
defensible principle: the architecture reduced the space of possible
solutions before training began. Biological vision is not a blank
statistical learner; neural systems differ in morphology,
electrophysiology, and molecular identity, and large-scale human brain
transcriptomic mapping shows that anatomical regions also differ in
molecular organization {[}Cadwell, 2016; Hawrylycz, 2012{]}. CNNs are
not miniature cortical circuits, but they do capture one functional
lesson from visual neuroscience: visual structure is not arbitrary, so
an image-recognition system should not have to learn every spatial
relation from scratch.

Convolution encodes this lesson through three linked constraints: local
receptive fields, weight sharing, and hierarchical composition. A local
receptive field means that a unit initially sees only a small
neighborhood of the image rather than every pixel at once. Weight
sharing means that the same detector is reused across spatial positions,
so a feature useful in one part of an image remains useful elsewhere.
Hierarchical composition means that early layers can represent simple
local patterns, while later layers combine them into larger and more
abstract configurations. LeCun et al.~built this logic directly into
document-recognition systems: convolutional layers and shared weights
allowed the model to exploit the spatial regularities of visual input
instead of treating each pixel location as an unrelated statistical
variable {[}LeCun, 1998{]}. Krizhevsky et al.~later showed that the same
architectural family could scale to large natural-image recognition,
preserving locality and hierarchy while increasing depth and
representational capacity {[}Krizhevsky, 2012{]}.

\pandocbounded{\includegraphics[keepaspectratio,alt={Comparison of architectural inductive biases across three image-processing models, illustrating the efficiency argument for structural priors. (a) A dense model connects all input pixels to all output units, treating each spatial position as independent. (b) A Convolutional Neural Network (CNN) pipeline incorporates strong priors through local receptive fields, weight sharing, and hierarchical feature composition, enabling efficient learning of visual structure by reducing parameter space. (c) A Vision Transformer (ViT) reduces these domain-specific constraints by tokenizing the image into patches and applying global self-attention, allowing for higher representational capacity but necessitating larger-scale pretraining.}]{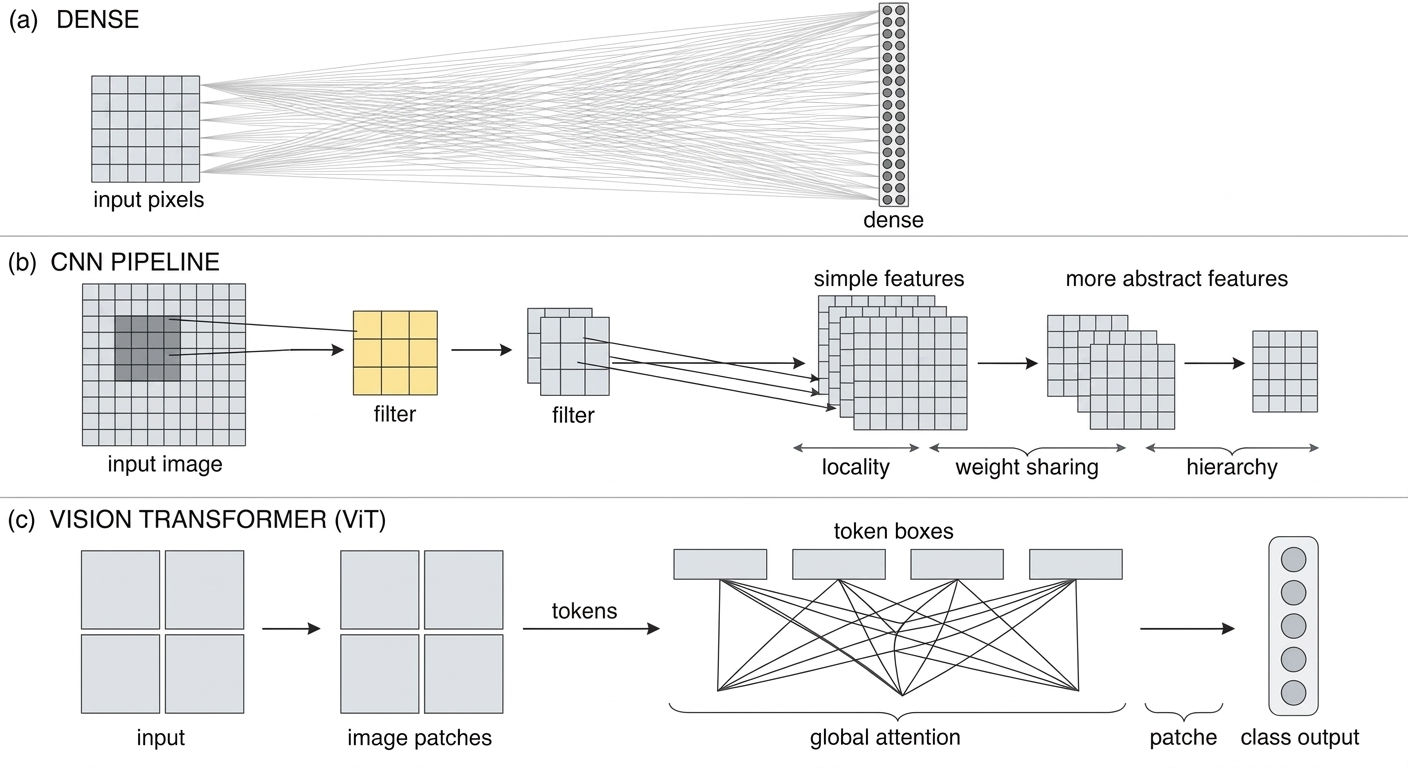}}
\emph{Figure 8. Comparison of architectural inductive biases across
three image-processing models, illustrating the efficiency argument for
structural priors. (a) A dense model connects all input pixels to all
output units, treating each spatial position as independent. (b) A
Convolutional Neural Network (CNN) pipeline incorporates strong priors
through local receptive fields, weight sharing, and hierarchical feature
composition, enabling efficient learning of visual structure by reducing
parameter space. (c) A Vision Transformer (ViT) reduces these
domain-specific constraints by tokenizing the image into patches and
applying global self-attention, allowing for higher representational
capacity but necessitating larger-scale pretraining.}

The parameter arithmetic makes the efficiency argument concrete. A
single 3 × 3 convolutional kernel applied to a 64-channel input requires
3 × 3 × 64 = 576 weights for one output channel. If the layer produces
64 output channels, it requires 3 × 3 × 64 × 64 = 36,864 weights. That
count does not grow merely because the input image becomes wider or
taller; the same learned filter slides across the spatial field. The
architecture therefore separates the number of learned parameters from
the full spatial resolution of the input. Without locality and sharing,
a model would need to learn far more independent relationships among
positions, many of which are redundant for ordinary visual recognition.
The CNN's advantage is not simply that it is smaller. Its advantage is
that its smaller parameterization follows from a correct structural
assumption about images.

The comparison with Vision Transformers clarifies what is lost when that
assumption is weakened. Dosovitskiy et al.~showed that a Transformer
applied to image patches can perform very well, but also that this
reduced image-specific inductive bias makes the model depend more
strongly on large-scale pretraining before it can compete with or exceed
strong convolutional baselines {[}Dosovitskiy, 2021{]}. This result does
not show that attention cannot process images. It shows something more
important for the present argument: when architecture supplies fewer
domain constraints, data and compute must carry more of the burden.

CNNs therefore demonstrate the central efficiency principle of
structural priors. A system can reach strong performance by building a
domain-appropriate assumption into its topology, or it can remove that
assumption and attempt to recover it statistically from much larger
exposure. The next chapter turns from architectural efficiency to the
hardware economics that made general-purpose dense computation the
field's default route.

\begin{center}\rule{0.5\linewidth}{0.5pt}\end{center}

\section{Chapter 5. The Hardware Lottery and the Monoculture It
Built}\label{chapter-5.-the-hardware-lottery-and-the-monoculture-it-built}

\subsubsection{5.1. The Transformer as Default: How One Architecture
Swallowed All
Modalities}\label{the-transformer-as-default-how-one-architecture-swallowed-all-modalities}

The Transformer entered machine learning as a solution to a specific
engineering problem: sequence-to-sequence modeling in language. Its
self-attention mechanism allowed each token in a sequence to compare
itself directly with other tokens, avoiding the bottlenecks of recurrent
encoder-decoder systems and achieving strong machine-translation results
with highly parallel computation {[}Vaswani et al., 2017{]}. That
success did not merely introduce a better language model. It created a
reusable computational template: convert the input into a sequence of
tokens, apply attention over those tokens, and scale the model.

Once this template became available, other modalities were increasingly
reformatted to fit it. Vision provides the clearest example. The Vision
Transformer did not begin from the structural assumptions that had made
convolutional networks powerful: local receptive fields, weight sharing
across space, and hierarchical feature extraction. Instead, it divided
an image into fixed-size patches, treated those patches as sequence
elements, and processed them with the same general attention machinery
originally designed for tokens in language {[}Dosovitskiy et al.,
2021{]}. This move did not prove that images and sentences share the
same computational structure. It proved that images could be recoded
into a form that the Transformer could ingest.

The contrast with convolutional networks matters because CNNs succeeded
by building visual structure into the architecture itself. LeNet used
local connectivity and shared weights to exploit the spatial
regularities of images {[}LeCun et al., 1998{]}. AlexNet later
demonstrated the power of deep convolutional vision at ImageNet scale,
using architectural priors that matched the statistical structure of
natural images rather than treating the visual field as an arbitrary
sequence {[}Krizhevsky et al., 2012{]}. The Vision Transformer, by
contrast, required much larger pretraining regimes to become competitive
with strong convolutional baselines, and its own results showed weaker
performance when trained only at ImageNet scale than when trained on far
larger datasets {[}Dosovitskiy et al., 2021{]}. The important point is
not that Transformer-based vision cannot work. It can. The point is that
performance increasingly came from scale compensating for the removal of
domain-specific structure.

Audio followed the same pattern. The Audio Spectrogram Transformer
represented sound as a spectrogram, split that time-frequency
representation into patches, and applied Transformer-style processing to
the resulting sequence {[}Gong et al., 2021{]}. This again produced a
functional model, but the architectural gesture was familiar: transform
the modality until it resembles a token sequence, then reuse the same
attention-centered topology. Language, images, and sound differ in their
native structure. Language unfolds symbolically and sequentially; vision
depends heavily on spatial locality and hierarchical composition;
audition requires fine-grained spectro-temporal analysis. Treating all
three as patch or token sequences therefore flattens modality-specific
computational demands into a common format.

Biology offers a useful constraint on this trend. Human neural tissue
does not implement cognition by repeating one uniform circuit at larger
scale. Anatomically patterned gene-expression maps show systematic
regional differences across the human brain {[}Hawrylycz et al.,
2012{]}. Patch-seq methods further demonstrate that neurons can differ
jointly in morphology, electrophysiology, and transcriptomic identity,
linking cellular structure to functional specialization {[}Cadwell et
al., 2016{]}. These findings do not license a simple one-to-one mapping
between cortical areas and artificial modules, but they do undermine the
assumption that one undifferentiated topology should serve as the
default model for every cognitive domain.

The rise of the Transformer across language, vision, and audio therefore
marks a shift from structural design to architectural recoding. Instead
of asking what form of computation each modality demands, the field
increasingly asks how each modality can be translated into the form one
dominant architecture already accepts. Section 5.2 examines the
selection pressure behind that shift: the GPU-centered hardware economy
that made dense, attention-based computation the easiest path to scale.

\begin{center}\rule{0.5\linewidth}{0.5pt}\end{center}

\subsubsection{5.2. GPU Economics Against Neuromorphic and Asynchronous
Design}\label{gpu-economics-against-neuromorphic-and-asynchronous-design}

The Hardware Lottery did not make the Transformer scientifically empty,
nor did it prove that alternative architectures were superior. Its
deeper effect was subtler: it changed which architectural ideas could
become cheap enough, fast enough, and convenient enough to dominate
research practice. Hooker describes this process as a selection pressure
produced by available hardware, in which architectures aligned with the
dominant computational substrate receive disproportionate experimental
attention and engineering refinement {[}Hooker, 2021{]}. In contemporary
AI, that substrate was the GPU-centered stack: an ecosystem built to
reward large, regular, parallel operations.

The Transformer fit that ecosystem unusually well. Its scaled
dot-product attention computes interactions among tokens through matrix
operations, and its feed-forward blocks apply repeated dense
transformations across representations {[}Vaswani et al., 2017{]}. These
operations map naturally onto hardware and software pipelines optimized
for dense linear algebra. This compatibility mattered. A model class
that trains efficiently on available accelerators can attract more
benchmarks, larger implementations, better tooling, and greater
institutional investment. Over time, this feedback loop can make
architectural prevalence appear theoretically inevitable, even when it
partly reflects infrastructural convenience.

\pandocbounded{\includegraphics[keepaspectratio,alt={The hardware lottery creates a selection pressure that locks in a Transformer/GPU monoculture. (a) A socio-infrastructural feedback loop accelerates GPU-stack efficiency, attracting further investment and tooling, which entrenches GPU-compatible, dense matrix-multiplication-heavy architectures; (b) consequently, diverse applications across text, vision, audio, and multimodal domains inherit the same accelerator-optimized infrastructure and its shared blind spots; (c) by contrast, an event-driven neuromorphic substrate, organized for sparsity and asynchrony in silicon or neural tissue, illustrates fundamentally different and potentially more efficient organizational structures that the current infrastructure selects against.}]{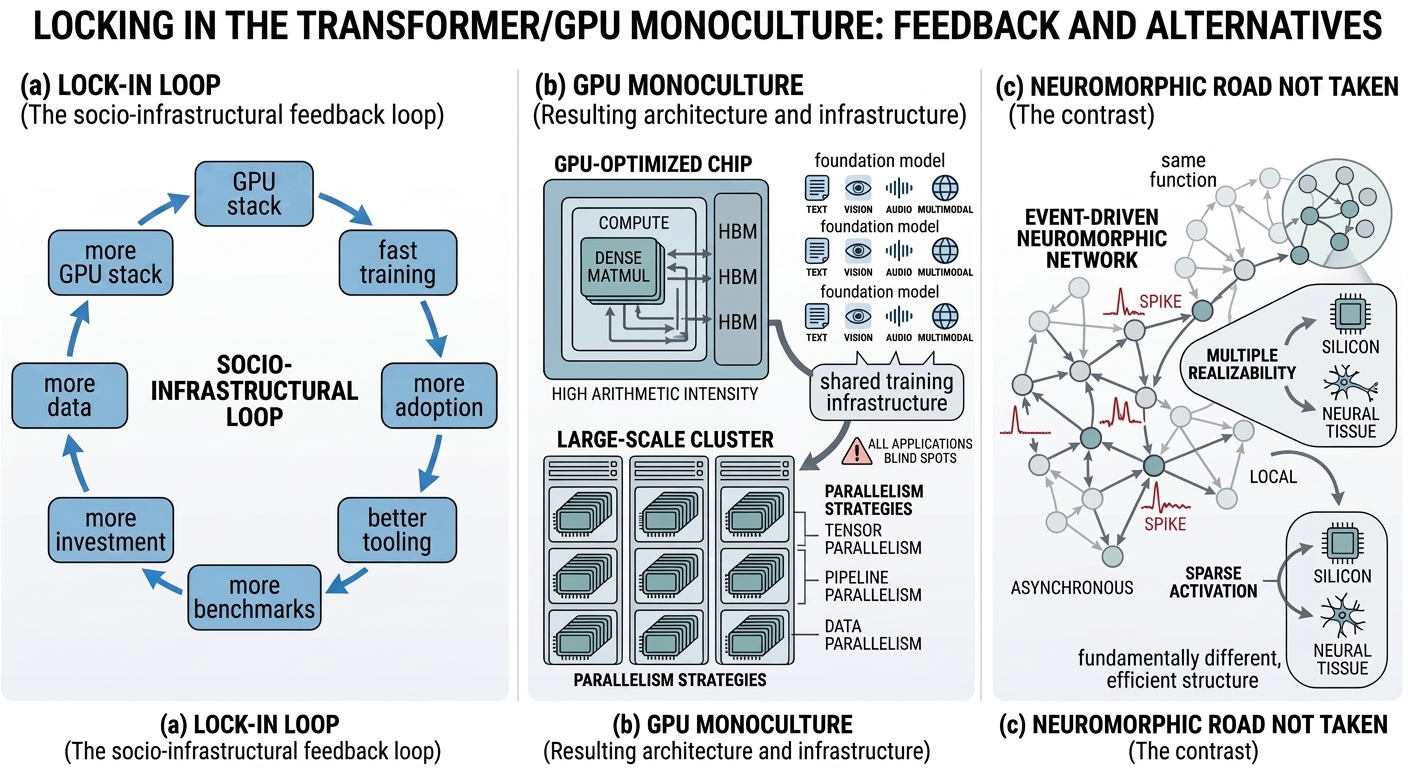}}
\emph{Figure 9. The hardware lottery creates a selection pressure that
locks in a Transformer/GPU monoculture. (a) A socio-infrastructural
feedback loop accelerates GPU-stack efficiency, attracting further
investment and tooling, which entrenches GPU-compatible, dense
matrix-multiplication-heavy architectures; (b) consequently, diverse
applications across text, vision, audio, and multimodal domains inherit
the same accelerator-optimized infrastructure and its shared blind
spots; (c) by contrast, an event-driven neuromorphic substrate,
organized for sparsity and asynchrony in silicon or neural tissue,
illustrates fundamentally different and potentially more efficient
organizational structures that the current infrastructure selects
against.}

Neuromorphic and asynchronous designs reveal what this convenience
excludes. A neuromorphic processor such as Loihi organizes computation
around event-driven spiking activity, distributed cores, and local
learning mechanisms rather than a single rhythm of dense, synchronous
matrix updates {[}Davies et al., 2018{]}. Event-driven computation means
that parts of the system can remain inactive until a relevant signal
occurs; asynchronous computation means that the entire system need not
advance in lockstep. These principles do not make neuromorphic hardware
a direct replica of cortex, and they do not imply that current
neuromorphic chips outperform GPUs across modern foundation-model
workloads. They do, however, represent a different hypothesis about
efficient computation: not maximum uniform throughput at every step, but
selective activity shaped by timing, locality, and sparsity.

That hypothesis has a biological rationale. Cortical signaling operates
under severe metabolic constraints, and Lennie argues that the brain
cannot sustain high firing rates across all neurons at once without
exceeding its energy budget {[}Lennie, 2003{]}. Biological intelligence
therefore depends on selective use of activity rather than continuous
full-system activation. This fact does not justify a simple analogy
between brains and neuromorphic chips. It supports a narrower and
stronger point: if natural cognition solves many problems under sparse,
temporally structured, and energy-constrained conditions, then
architectures built only around dense synchronous acceleration may
systematically neglect useful regions of the design space.

Putnam's multiple realizability argument sharpens this distinction
{[}Putnam, 1967{]}. A cognitive function need not belong to one physical
substrate alone. The same functional goal may, in principle, be
implemented in silicon, nervous tissue, or some other medium. But
multiple realizability does not mean that every substrate or topology
suits every function equally well. Functional adequacy depends on the
match among the task, the organization of computation, and the physical
constraints under which the system operates. GPU efficiency therefore
matters, but it cannot serve as evidence that a GPU-favored topology is
cognitively universal.

The real cost of GPU economics was not that it produced a bad
architecture. It produced a narrowing of architectural imagination.
Dense Transformer-style computation became the default not only where it
fit the problem, but also where sparse, event-driven, recurrent, or
temporally specialized designs might have expressed better structural
priors. The next section develops the systemic-fragility argument that
follows from this narrowing: once hardware convenience becomes mistaken
for functional adequacy, architectural monoculture begins to resemble
scientific convergence rather than infrastructural lock-in.

\begin{center}\rule{0.5\linewidth}{0.5pt}\end{center}

\subsubsection{5.3. Systemic Fragility: The Risk of Foundation Model
Monoculture}\label{systemic-fragility-the-risk-of-foundation-model-monoculture}

Foundation-model monoculture is not primarily a matter of identical
weights, identical datasets, or identical applications. It is
architectural. The Transformer introduced a highly effective
self-attention mechanism in which each token can directly compare itself
with every other token in a sequence, allowing relational structure to
be computed in parallel rather than step by step {[}Vaswani, 2017{]}.
That mechanism has proved extraordinarily useful. Systemic risk begins
when the same computational topology becomes the default substrate for
many domains whose native structure differs sharply. GPT-3 applies
Transformer machinery to subword language tokens {[}Brown, 2020{]}.
Vision Transformer applies the same basic mechanism to image patches
{[}Dosovitskiy, 2021{]}. Audio Spectrogram Transformer applies it to
patches of a log-mel spectrogram {[}Gong, 2021{]}. The tokenizer
changes, but the central processing form remains strikingly similar:
heterogeneous signals pass through a shared token-relational
architecture.

This convergence did not occur in a neutral design space. The Hardware
Lottery describes how available hardware can select architectures by
making some forms of computation cheaper and easier to scale than others
{[}Hooker, 2021{]}. Modern accelerators reward dense matrix
multiplication, high arithmetic intensity, and regular parallel
workloads. Sze and colleagues show why this matters physically:
computation and memory movement have very different energy costs, so
architectures that reuse data efficiently inside dense numerical kernels
gain a practical advantage independent of whether their inductive bias
best fits the target domain {[}Sze, 2017{]}. Large-scale training
systems then reinforce the same direction. Narayanan and colleagues
demonstrate how GPT-scale Transformer models can be trained through
coordinated tensor, pipeline, and data parallelism across thousands of
GPUs {[}Narayanan, 2021{]}. Once this software-hardware stack becomes
the common route for language, vision, audio, and multimodal systems, a
limitation in the shared infrastructure no longer remains local. It can
propagate across many downstream models because those models inherit the
same computational assumptions.

\pandocbounded{\includegraphics[keepaspectratio,alt={Schematic overview of cross-modal tokenization and shared processing in foundation models, illustrating architectural monoculture. (a) Text is segmented into subword tokens and embedded as a sequence; (b) images are divided into patches and linearly projected into tokens; (c) audio waveforms are converted to log-mel spectrograms, then patched and projected into tokens; (d) all three distinct token sequences are fed into a single, repeated Transformer block topology, demonstrating that while tokenizers vary, the central computational machinery remains uniform across modalities.}]{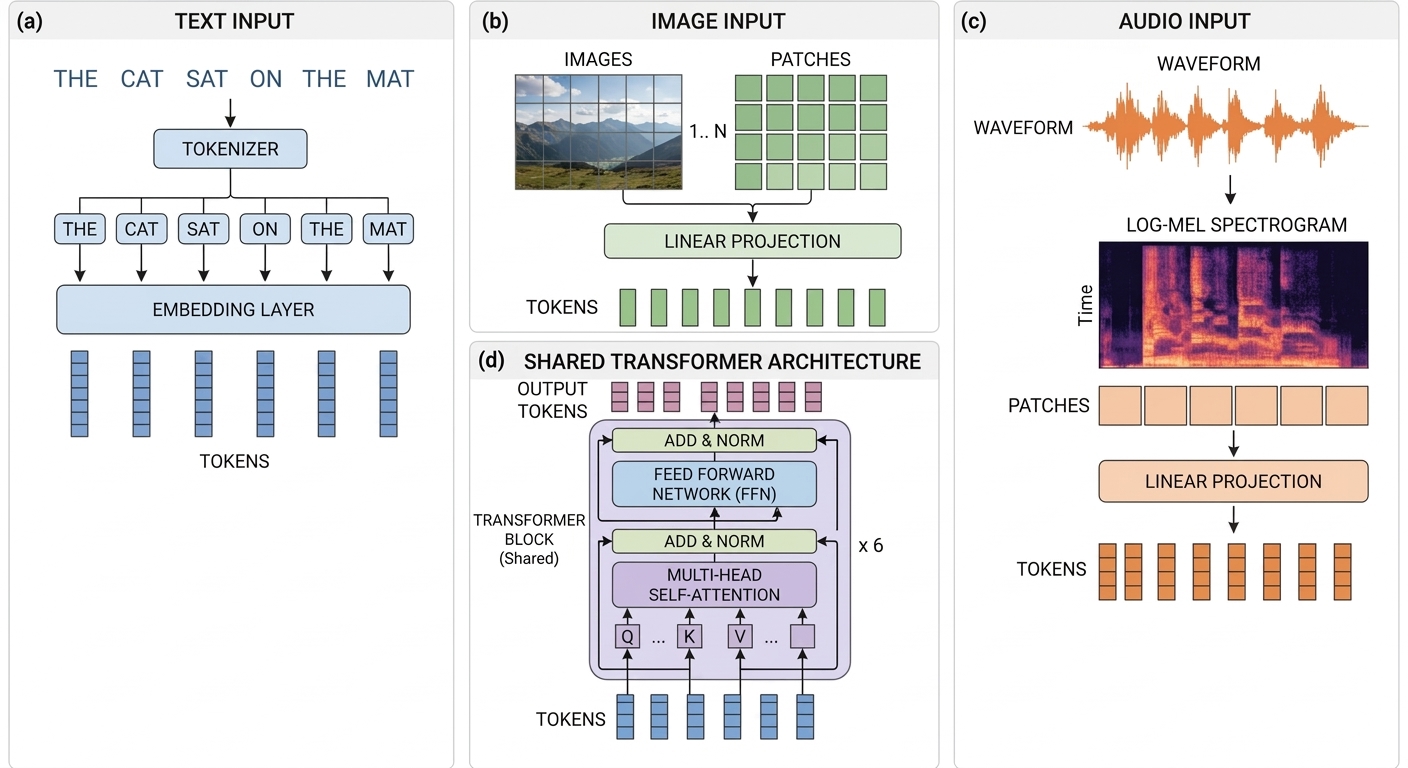}}
\emph{Figure 10. Schematic overview of cross-modal tokenization and
shared processing in foundation models, illustrating architectural
monoculture. (a) Text is segmented into subword tokens and embedded as a
sequence; (b) images are divided into patches and linearly projected
into tokens; (c) audio waveforms are converted to log-mel spectrograms,
then patched and projected into tokens; (d) all three distinct token
sequences are fed into a single, repeated Transformer block topology,
demonstrating that while tokenizers vary, the central computational
machinery remains uniform across modalities.}

Foundation models amplify that effect because they serve as reusable
bases for many applications. Bommasani and colleagues emphasize both
sides of this leverage: improvements can spread quickly, but defects,
biases, and vulnerabilities can also be inherited by a wide range of
systems built on the same base {[}Bommasani, 2021{]}. The problem is
therefore not only deployment hygiene. A shared architecture can create
shared blind spots. If every modality must first be translated into a
token sequence and processed through the same relational mechanism, then
architectural convenience decides which structures become easy to learn
and which structures must be approximated indirectly.

Biological intelligence provides a useful contrast, not as a blueprint
to copy, but as evidence that heterogeneous function creates pressure
for heterogeneous structure. Cytoarchitectonic mapping treats
differences in lamination, cellular density, and regional organization
as reproducible anatomical facts rather than decorative biological
detail {[}Zilles and Amunts, 2010{]}. The Jülich Brain Atlas extends
this tradition into a probabilistic three-dimensional framework, showing
that the human brain contains identifiable cytoarchitectonic fields
rather than a single repeated sheet scaled up or down {[}Amunts,
2020{]}. At the same time, the cortex does contain recurring motifs.
Columnar organization and canonical microcircuit accounts identify
shared local principles {[}Mountcastle, 1997; Douglas and Martin,
2004{]}. But shared motifs do not entail architectural uniformity.
Harris and Shepherd describe neocortical organization as ``themes and
variations,'' where common circuit elements coexist with area-specific
differences in connectivity, cellular composition, and computational
role {[}Harris and Shepherd, 2015{]}. Even within the visual system,
hierarchical organization depends on differentiated feedforward,
feedback, and lateral relations rather than on a single processor
repeated without structural change {[}Felleman and Van Essen, 1991{]}.

Functionalism clarifies the engineering lesson without requiring
biological imitation. Putnam's multiple realizability thesis allows the
same function to be implemented in different physical substrates, but it
does not imply that any organization can realize any function equally
well {[}Putnam, 1967; Putnam, 1975{]}. Causal organization still
matters. A mechanism for high-resolution spatial invariance, a mechanism
for temporally extended auditory patterning, and a mechanism for
reward-gated action selection differ in what information they preserve,
transform, suppress, and route. Fodor's modularity thesis gives this
point a cognitive form: some functions operate through domain-specific
systems with internal constraints appropriate to their tasks {[}Fodor,
1983{]}. Modularity does not mean isolation. It means that a subsystem
can possess its own structure while still communicating with the larger
system.

Mixture-of-Experts systems occupy an important borderline position.
Sparsely gated MoE layers route inputs to selected expert subnetworks,
allowing very large parameter counts without activating every parameter
on every example {[}Shazeer, 2017{]}. Switch Transformers simplify this
routing by sending each token to a single expert, improving scaling
efficiency in very large Transformer models {[}Fedus, 2022{]}. These
systems distribute computation and can specialize parameters, so they
should not be dismissed as irrelevant. Their limitation is more
specific: when experts share the same underlying layer type and remain
embedded in the same global Transformer topology, the diversity remains
mostly quantitative. The model gains separately addressable capacity,
not necessarily distinct computational organs with different roles,
interfaces, and internal geometries.

The systemic fragility of foundation-model monoculture therefore lies in
a category error: the field may mistake routed scale, shared
infrastructure, and cross-modal tokenization for genuine architectural
differentiation. Part III begins from the opposite premise. If different
cognitive functions impose different structural demands, the next task
is to specify which functions require which topologies, and how a
heterogeneous system can connect them without collapsing them back into
one universal processor.

\begin{center}\rule{0.5\linewidth}{0.5pt}\end{center}

\section{Part III: Rebuilding Heterogeneous
AI}\label{part-iii-rebuilding-heterogeneous-ai}

\section{Chapter 6. Functionalism
Reconsidered}\label{chapter-6.-functionalism-reconsidered}

\subsubsection{6.1. Multiple Realizability and Topological Necessity:
Putnam, Fodor, and
AI}\label{multiple-realizability-and-topological-necessity-putnam-fodor-and-ai}

Putnam's multiple realizability thesis gives artificial intelligence an
important permission, but not an unlimited license. A cognitive state or
function need not be identical with one biological material; silicon,
neurons, spikes, and dense numerical operations may all realize
cognition if they preserve the relevant causal organization {[}Putnam,
1967; Putnam, 1975{]}. This point blocks biological chauvinism. It does
not support architectural indifference. Multiple realizability separates
function from a single substrate, but it does not say that any topology
can realize any function efficiently, reliably, or at all.

Marr's three-level analysis clarifies the missing constraint. A
cognitive system can be described by its computational goal, its
algorithmic and representational procedure, and its physical
implementation {[}Marr, 1982{]}. Functionalism loosens the bond between
cognition and biological tissue at the implementation level, but the
algorithmic level remains indispensable. A visual system that must
preserve spatial adjacency, an auditory system that must resolve
time-frequency structure, and an executive system that must gate action
under competing demands face different algorithmic problems. They may
all run on non-biological hardware, but their internal organization
cannot be treated as interchangeable merely because the substrate is
artificial.

Fodor's modularity thesis sharpens this point into an architectural
principle. Fodor did not describe modules as arbitrary boxes; he
characterized them as specialized mechanisms with domain-specific
inputs, constrained access to information, and distinctive processing
limits {[}Fodor, 1983{]}. The lesson for AI is not that every cortical
area maps neatly onto a Fodorian module. That would replace one
simplification with another. The stronger conclusion is that some
cognitive functions require specialized processing regimes, and those
regimes must appear in the architecture before training begins. A system
for spectro-temporal parsing, a system for relational sequence binding,
and a system for action selection may communicate through shared
interfaces, but they should not be presumed to share the same internal
topology.

\pandocbounded{\includegraphics[keepaspectratio,alt={A unified framework for substrate independence and functional decomposition in cognitive architecture. (a) Functional domains decompose into specific goals and algorithms (top); (b) these functions are realizable across three distinct substrates: biological circuits (top-middle), analog spiking networks (bottom-middle), and silicon digital processors (bottom), with each substrate showing specialized components for memory, control, and binding; (c) biological networks achieve function through distributed, directed pathways, contrasting with the more uniform topology of artificial neural networks (bottom-right), demonstrating that functional demands constrains hardware architecture.}]{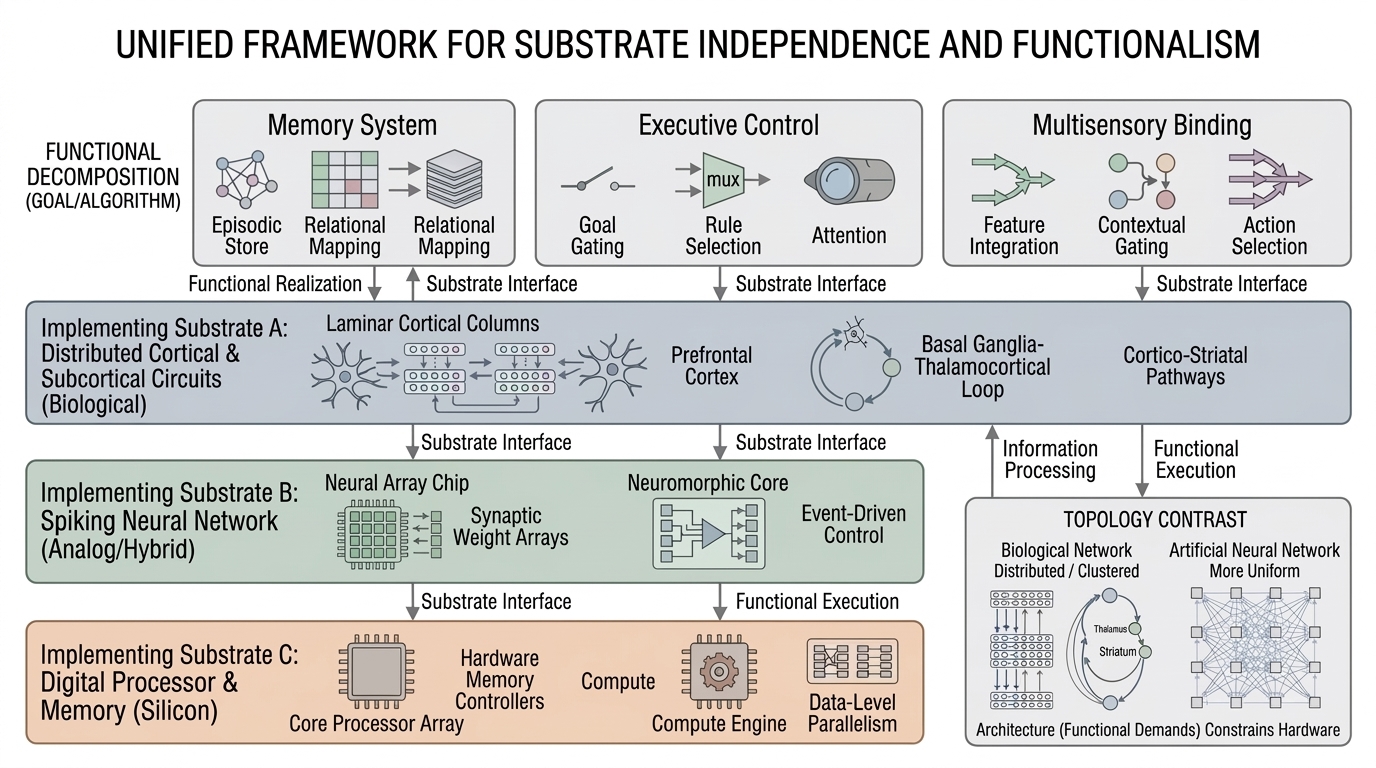}}
\emph{Figure 11. A unified framework for substrate independence and
functional decomposition in cognitive architecture. (a) Functional
domains decompose into specific goals and algorithms (top); (b) these
functions are realizable across three distinct substrates: biological
circuits (top-middle), analog spiking networks (bottom-middle), and
silicon digital processors (bottom), with each substrate showing
specialized components for memory, control, and binding; (c) biological
networks achieve function through distributed, directed pathways,
contrasting with the more uniform topology of artificial neural networks
(bottom-right), demonstrating that functional demands constrains
hardware architecture.}

Neuroscience supplies the empirical boundary condition for this
interpretation. Cytoarchitectural mapping shows that cortical regions
differ in laminar organization, cell density, cellular morphology, and
reproducible three-dimensional structural profiles {[}Zilles and Amunts,
2010; Amunts et al., 2020{]}. The neocortex does contain recurring
motifs, including columnar organization and recurrent
excitatory-inhibitory circuitry, but the best current framing treats
these motifs as themes and variations rather than proof of a uniform
cortex {[}Mountcastle, 1997; Harris and Shepherd, 2015{]}. Biological
cognition therefore demonstrates constrained variation: related
computational goals can appear across different circuits, but viable
circuits differ in lamination, neuron types, connection strengths, and
long-range placement.

The same constraint appears at the systems level. Primate visual cortex
forms a distributed hierarchy with feedforward, feedback, and lateral
pathways, so a region's computation depends not only on its local
machinery but also on its position in a directed network {[}Felleman and
Van Essen, 1991{]}. Basal-ganglia-thalamocortical loops impose an even
stronger topological constraint: executive control emerges from
recurrent, partially segregated channels that route cortical information
through specific nuclei and thalamic relays {[}Alexander et al.,
1986{]}. These mechanisms do not add generic processing capacity to a
uniform substrate. They make function depend on organized loops, gates,
and pathways.

Contemporary AI often flattens this distinction. Scaling-law results
show that, within tested Transformer-like regimes, loss varies
predictably with parameter count, data, and compute {[}Kaplan et al.,
2020{]}. Those results do not prove that topology is irrelevant. The
Transformer itself carries a structural cost: self-attention compares
tokens pairwise, so its standard attention operation grows quadratically
with sequence length {[}Vaswani et al., 2017{]}. Hardware surveys
further show that data movement dominates energy cost in deep-learning
accelerators, making memory traffic a central systems constraint rather
than a minor implementation detail {[}Sze et al., 2017{]}. Large-scale
training systems can make this topology run at impressive scale, but
only through extensive tensor, pipeline, and data parallelism
engineering {[}Narayanan et al., 2021{]}. The engineering achievement
confirms that topology matters; it does not erase the cost of forcing
many functions through one attention-centered form.

Functionalism therefore supports heterogeneous AI more strongly than
architectural monoculture. A Transformer may legitimately realize
functions suited to relational binding, contextual retrieval, and
ordered prediction. It does not follow that the same machinery should
serve as a universal cortex for vision, audition, control, and working
memory. Multiple realizability means that cognition can be built in more
than one material form, but mechanistic explanation still requires
organized components whose activities match the target capacity
{[}Craver, 2007{]}. The next section applies this standard to
Mixture-of-Experts, asking whether it creates genuine structural
diversity or merely partitions parameters among architectural clones.

\begin{center}\rule{0.5\linewidth}{0.5pt}\end{center}

\subsubsection{6.2. MoE Is Not Structural Diversity: The Clone
Partitioning
Problem}\label{moe-is-not-structural-diversity-the-clone-partitioning-problem}

Mixture-of-Experts appears to answer the charge that contemporary AI has
become architecturally homogeneous. Instead of activating every
parameter for every input, an MoE model uses a router to send each token
representation to only a small number of expert sub-networks. Shazeer
and colleagues introduced sparsely gated expert layers as a way to
expand model capacity while limiting per-token computation, and the
Switch Transformer later simplified this design by routing each token to
a single expert inside a Transformer block {[}Shazeer et al., 2017;
Fedus et al., 2022{]}. This is a real engineering advance. It reduces
the cost of activating very large parameter sets and gives dense scaling
a more efficient alternative. But it does not solve the structural
problem raised in this chapter.

The key distinction is between sparse computation and heterogeneous
organization. In standard Transformer MoE systems, the routed expert is
usually not a different kind of cognitive module. It is a replicated
feed-forward sublayer: a dense matrix transformation followed by a
nonlinearity, placed inside the broader Transformer topology introduced
by Vaswani and colleagues {[}Vaswani et al., 2017{]}. The attention
mechanism, residual stream, token embedding format, and normalization
structure remain shared. The router changes which parameter block
receives a token, but it does not change the causal grammar of the
computation. One expert may learn different weights from another, yet
both remain instances of the same architectural species.

\pandocbounded{\includegraphics[keepaspectratio,alt={A comparison of artificial Mixture-of-Experts, biological cortical organization, and canonical computational synthesis, demonstrating that MoE sparse routing multiplies parameters rather than creating architectural heterogeneity. (a) The artificial MoE model replicates a single canonical computation (dense MatMul + ReLU) across all experts, achieving parameter replication without structural diversity. (b) The biological cortex exhibits genuine structural specialization, with distinct laminar and cellular compositions, cell-type identities (neurons in inset), and dendritic morphologies across different areas (Area A vs.~B), linked by feedforward, feedback, and lateral pathways. (c) The Transformer remains a canonical computation dominated by global self-attention, in contrast to the diverse structural forms---such as local gating, persistent state, and feedback regulation---that biological systems utilize to implement unique computational primitives.}]{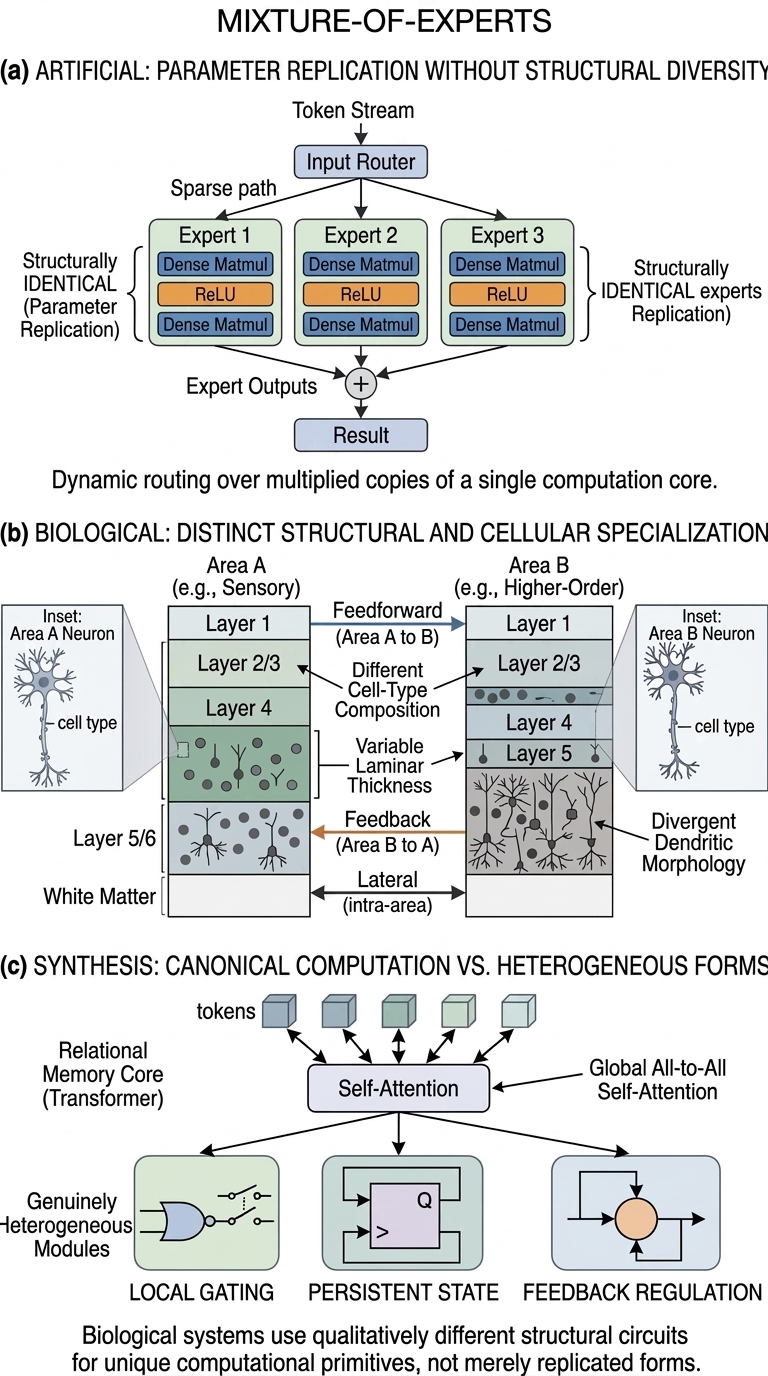}}
\emph{Figure 12. A comparison of artificial Mixture-of-Experts,
biological cortical organization, and canonical computational synthesis,
demonstrating that MoE sparse routing multiplies parameters rather than
creating architectural heterogeneity. (a) The artificial MoE model
replicates a single canonical computation (dense MatMul + ReLU) across
all experts, achieving parameter replication without structural
diversity. (b) The biological cortex exhibits genuine structural
specialization, with distinct laminar and cellular compositions,
cell-type identities (neurons in inset), and dendritic morphologies
across different areas (Area A vs.~B), linked by feedforward, feedback,
and lateral pathways. (c) The Transformer remains a canonical
computation dominated by global self-attention, in contrast to the
diverse structural forms---such as local gating, persistent state, and
feedback regulation---that biological systems utilize to implement
unique computational primitives.}

That limitation matters because functionalism does not equate multiple
parts with multiple functions. Putnam's multiple realizability thesis
rejects the idea that a mental function must be tied to one biological
material, but it does not imply that any internal organization can
realize any function equally well {[}Putnam, 1967; Putnam, 1975{]}.
Marr's three-level analysis makes the same point in computational
language: a system must be understood not only by the goal it serves,
but also by the algorithmic strategy and implementation constraints
through which it serves that goal {[}Marr, 1982{]}. If two sub-networks
share the same input geometry, temporal dynamics, memory regime, and
transformation type, then their learned weights may specialize, but
their architecture has not diversified in the stronger functional sense.

Biology provides the stricter comparison. Cortical areas do not differ
only because different populations receive different training histories.
They differ in laminar proportions, cell-type composition, dendritic
organization, projection patterns, and local circuit motifs. Brodmann's
early cytoarchitectonic maps and the later Jülich probabilistic atlas
both treat structural variation as a basis for identifying distinct
cortical areas {[}Zilles and Amunts, 2010; Amunts et al., 2020{]}. The
primate visual system also shows that differentiated areas remain
interactive: V1, MT, and other visual regions participate in a
distributed hierarchy with ordered feedforward, feedback, and lateral
connections rather than functioning as isolated boxes {[}Felleman and
Van Essen, 1991{]}. The lesson is not that artificial systems must copy
cortical tissue. The lesson is that functional specialization receives
structural expression before learning begins.

Under that standard, standard MoE is better described as clone
partitioning. It multiplies a common processing template and lets a
router allocate tokens among copies. This can yield useful statistical
specialization, but it does not build the architectural equivalent of
distinct visual, auditory, episodic-memory, working-memory, or
executive-control systems. A visual pathway may require locality and
hierarchical spatial composition; auditory processing may require
log-scale spectro-temporal structure; working memory may require
maintained state under interference; executive control may require
gating and action-selection dynamics. A set of Transformer feed-forward
experts does not acquire these distinct inductive biases merely because
the router sends different tokens to different copies.

The systems reason for this persistence is clear. MoE remains friendly
to the same hardware regime that made the Transformer dominant: dense
matrix multiplication on GPU- and TPU-style accelerators {[}Hooker,
2021{]}. MoE improves the capacity-to-compute tradeoff while preserving
the computational primitive that current infrastructure executes
efficiently. That makes it an optimization inside the hardware lottery,
not an escape from it. Foundation-model homogenization therefore remains
a live risk: the model may become larger and more sparsely activated
while its underlying repertoire of causal mechanisms stays narrow
{[}Bommasani et al., 2021{]}.

MoE should therefore receive credit for what it accomplishes and not for
what it does not. It is an important method for conditional computation,
scalable capacity, and routing-based parameter efficiency. It is not, by
itself, evidence that AI has recovered the kind of structural diversity
implied by cytoarchitecture, modular cognitive science, or mechanistic
explanation in neuroscience {[}Fodor, 1983; Craver, 2007{]}. A genuinely
heterogeneous system would assign different cognitive functions to
modules with different causal organizations and domain-appropriate
inductive biases, then connect those modules through explicit
interfaces. If MoE only partitions a single topology into many sparsely
used copies, it may postpone the costs of homogenization, but it cannot
remove the deeper scaling asymptote examined in Section 6.3.

\begin{center}\rule{0.5\linewidth}{0.5pt}\end{center}

\subsubsection{6.3. The Scaling Asymptote: What Statistical Learning
Cannot
Reach}\label{the-scaling-asymptote-what-statistical-learning-cannot-reach}

Scaling laws describe a real achievement, but they do not prove that one
architecture can become every architecture simply by becoming larger.
Kaplan et al.~showed that language-model loss follows smooth power-law
trends as parameter count, dataset size, and training compute increase,
and Brown et al.~demonstrated the practical force of that result when a
very large Transformer displayed few-shot abilities unavailable to
smaller models in the same family {[}Kaplan, 2020; Brown, 2020{]}. These
findings deserve full credit: scale can extract more performance from a
representational system that already contains useful machinery. The
mistake begins when this empirical regularity becomes an architectural
thesis---the assumption that adding more layers, channels, tokens, and
parameters will eventually supply any missing computational form.

Functionalism does not support that assumption. Putnam's multiple
realizability thesis separates mental function from any single
biological material, but it does not separate function from causal
organization {[}Putnam, 1967; Putnam, 1975{]}. A system may implement a
function in silicon rather than neurons, yet it must still instantiate
the relevant pattern of inputs, internal transformations, state
transitions, and outputs. Fodor's modularity sharpens the same point:
some cognitive capacities gain speed and reliability because they
operate over restricted input formats and specialized procedures rather
than through unrestricted access to a single general workspace {[}Fodor,
1983{]}. Marr's distinction between computational, algorithmic, and
implementation levels explains why parameter growth cannot erase this
requirement. Better implementation can execute a computation more
efficiently, and statistical training can improve an input-output
mapping, but neither replaces the algorithmic decomposition that
specifies what representations and procedures the task requires {[}Marr,
1982{]}.

Biology makes that limit concrete. Cortical areas do not differ only by
having more or fewer copies of one generic circuit. Brodmann's
cytoarchitectural maps and later probabilistic atlases show systematic
regional variation in laminar structure, cellular density, and areal
organization {[}Zilles, 2010; Amunts, 2020{]}. Conserved cortical motifs
exist, including recurrent excitatory-inhibitory circuitry, but modern
accounts describe neocortex as ``themes and variations,'' not as a
uniform sheet repeated at different scales {[}Mountcastle, 1997;
Douglas, 2004; Harris, 2015{]}. Visual cortex illustrates the point
especially clearly: primate cortical processing depends on
hierarchically and reciprocally connected areas with distinct
feedforward, feedback, and lateral roles, not on undifferentiated
replication of one processing block {[}Felleman, 1991{]}. At finer
scales, transcriptomic atlases show anatomically patterned gene
expression across human brain regions, while Patch-seq links neuronal
morphology, electrophysiology, and gene expression in single cells
{[}Hawrylycz, 2012; Cadwell, 2016{]}. Integrated classifications of
cortical inhibitory neurons further show that cell classes differ
jointly in molecular, electrical, and morphological properties
{[}Gouwens, 2020{]}. Dendrites add another constraint: their branching
structure and active conductances shape local nonlinear integration, so
a neuron's computational role depends on morphology and channel
distribution, not only on how many synapses it receives {[}London, 2005;
Stuart, 2015{]}.

This does not imply that AI should copy cortex literally. It implies
that biological intelligence treats structure as a prior condition for
learning rather than as a by-product of scale. A generic unit repeated
enough times may enlarge capacity inside its existing representational
regime, but repetition alone does not create laminar routing,
cell-type-specific inhibition, dendritic compartmentalization, or
area-specific input-output topology. The engineering analogue appears in
the Transformer. Self-attention computes similarity-weighted
interactions among tokens, which makes it powerful for relational
binding across sequences {[}Vaswani, 2017{]}. Modern analyses relate
attention to Hopfield-style associative memory, emphasizing storage and
retrieval of patterns rather than domain-specific sensory decomposition
or action selection {[}Ramsauer, 2021{]}. That computational affinity
helps explain the Transformer's success, but it also localizes its
strength: it resembles a content-addressable relational memory more than
a universal cortex. Hippocampal theories likewise emphasize relational
representation, spatial-cognitive mapping, and declarative memory
organization rather than all-purpose cortical computation {[}O'Keefe,
1978; Eichenbaum, 2004{]}.

Mixture-of-Experts architectures do not remove this asymptote when the
experts share the same internal grammar. Sparsely gated MoE layers and
Switch Transformers multiply parameter capacity by routing tokens to
different expert subnetworks, often keeping per-token computation far
below what a dense model of comparable total size would require
{[}Shazeer, 2017; Fedus, 2022{]}. This is valuable conditional
computation, but it usually partitions a larger population of
architecturally similar feed-forward blocks. The router chooses which
copy processes a token; it does not choose among qualitatively different
computational topologies such as convolutional locality, recurrent
gating, log-scale spectro-temporal filtering, or persistent
working-memory maintenance. MoE therefore demonstrates parameter
specialization more clearly than structural specialization.

The scaling asymptote is not the point where larger models stop
improving. It is the point where additional statistical approximation no
longer answers the architectural question. Foundation models can absorb
many modalities under a shared sequence format, and that consolidation
creates both opportunities and risks {[}Bommasani, 2021{]}. Yet broad
applicability does not equal functional generality. Scaling can make one
topology mimic more functions under distributional pressure, but mimicry
differs from realizing the causal constraints that make a function
robust across contexts. The next chapter therefore should not reject the
Transformer; it should place it on the biological-functional blueprint
as one module---a powerful relational memory core---beside other
structures for auditory analysis, executive gating, working memory, and
multisensory binding.

\begin{center}\rule{0.5\linewidth}{0.5pt}\end{center}

\section{Chapter 7. Localizing the Transformer on the Biological
Blueprint}\label{chapter-7.-localizing-the-transformer-on-the-biological-blueprint}

\subsubsection{7.1. The Transformer as Hippocampal Analog, Not Universal
Cortex}\label{the-transformer-as-hippocampal-analog-not-universal-cortex}

The Transformer becomes biologically informative only when its analogy
is localized. It should not be treated as an artificial cortex, because
cortex does not operate as a single undifferentiated computational
sheet. A more defensible comparison places the Transformer near the
functional niche of the hippocampal formation: a system for binding
distributed elements into structured relations, retrieving context from
partial cues, and organizing sequences of events. This is a functional
analogy, not an anatomical one. Putnam's account of multiple
realizability allows the same cognitive role to appear in different
physical substrates, but it does not imply that any substrate or
topology can realize any function equally well {[}Putnam, 1967; Putnam,
1975{]}. The relevant question is therefore not whether attention heads
resemble hippocampal neurons. They do not. The question is whether
attention occupies a comparable causal role inside a larger cognitive
architecture.

The hippocampal formation has long been associated with relational and
declarative memory. Eichenbaum characterizes hippocampal representations
as networks that connect items, contexts, places, and events, enabling
flexible retrieval through association rather than through a single
sensory feature {[}Eichenbaum, 2004{]}. O'Keefe and Nadel's
cognitive-map theory gives this role a spatial formulation: the
hippocampus organizes locations and trajectories into map-like
structures that support navigation and inference {[}O'Keefe and Nadel,
1978{]}. Later accounts extend the same principle beyond space into
temporal order and event structure. Buzsáki and Tingley describe
hippocampal computation as sequence generation and sequence
organization, linking trajectories, episodes, and temporal context into
ordered representational patterns {[}Buzsáki and Tingley, 2018{]}.
Pattern separation adds another boundary condition: hippocampal
circuitry helps distinguish similar experiences so that overlapping
episodes remain separately retrievable {[}Yassa and Stark, 2011{]}.
Across these accounts, the hippocampus does not perform all cognition.
It performs a specialized form of relational organization.

Attention has a closely related computational profile. In the original
Transformer, each token constructs a query and compares it with keys
from other tokens; the resulting weights determine which value vectors
contribute to the token's updated representation {[}Vaswani et al.,
2017{]}. In plain terms, the mechanism lets each element retrieve
information from other elements according to learned relevance. Ramsauer
and colleagues sharpen this interpretation by showing that
Transformer-style attention corresponds mathematically to an update rule
in modern Hopfield networks, a class of associative memory systems that
retrieve stored patterns through similarity-based dynamics {[}Ramsauer
et al., 2021{]}. This profile makes attention especially well suited to
context-sensitive binding: it can connect a word to a prior referent, an
image patch to a broader scene context, or an event token to a preceding
sequence. That strength explains why the hippocampal analogy is useful.
Attention is not merely a generic reasoning device; it is a powerful
relational retrieval operation.

\pandocbounded{\includegraphics[keepaspectratio,alt={Functional localization of the Transformer analogy. (a) The hippocampal formation operates as a specialized relational memory system that binds distributed items, contexts, and events into retrievable structures; (b) Transformer attention functions as an analogous associative retrieval mechanism that routes information between token representations based on learned relevance; (c) both components are integrated within a larger cognitive architecture, emphasizing that the analogy is functional and substrate-neutral, not anatomical.}]{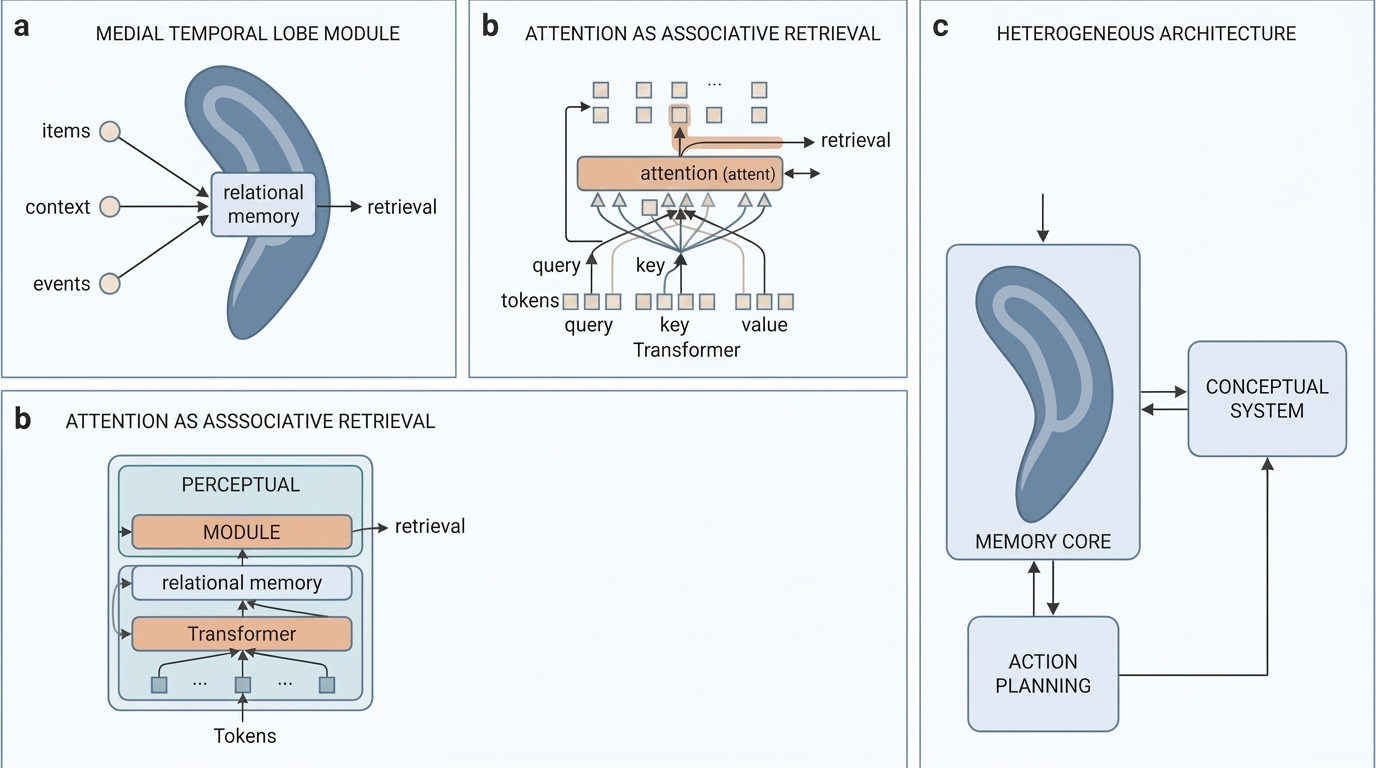}}
\emph{Figure 13. Functional localization of the Transformer analogy. (a)
The hippocampal formation operates as a specialized relational memory
system that binds distributed items, contexts, and events into
retrievable structures; (b) Transformer attention functions as an
analogous associative retrieval mechanism that routes information
between token representations based on learned relevance; (c) both
components are integrated within a larger cognitive architecture,
emphasizing that the analogy is functional and substrate-neutral, not
anatomical.}

The error begins when this specific strength becomes a universal
architectural doctrine. Vision Transformers convert an image into a
sequence of patches and process those patches through the same
attention-centered template used for language {[}Dosovitskiy et al.,
2021{]}. The Audio Spectrogram Transformer applies a similar patch-token
strategy to time-frequency representations of sound {[}Gong et al.,
2021{]}. These adaptations can perform well, but their success does not
prove that all sensory domains are naturally sequence-retrieval
problems. Patch tokenization changes the input format so that attention
can operate; it does not show that attention supplies the correct
structural prior for the domain. A model can force pixels, spectrogram
regions, and words into tokens while still ignoring the possibility that
spatial locality, frequency-time geometry, or motor-control gating
requires a different internal organization.

Fodor's modularity thesis clarifies the point. Cognitive systems often
depend on specialized input formats, restricted access patterns, and
domain-specific operations rather than on a single central processor
{[}Fodor, 1983{]}. Functionalism therefore supports substrate
neutrality, not topology neutrality. A silicon system may realize a
hippocampal-like memory role without copying hippocampal tissue, but it
cannot realize auditory analysis, working memory maintenance, executive
gating, and multisensory binding simply by renaming each problem as
attention over tokens. Working memory, for example, requires temporary
maintenance and manipulation under control constraints {[}Baddeley,
2003{]}. Executive gating requires reinforcement-sensitive selection and
updating, a profile captured in basal-ganglia models of dynamic gating
{[}Frank, 2005{]}. Multisensory integration requires circuits that bind
signals across modalities under spatial and temporal constraints
{[}Stein and Stanford, 2008{]}. These functions may communicate with a
Transformer-like memory core, but they should not be collapsed into it.

The correct conclusion is not anti-Transformer. It is
anti-universalization. A Transformer can serve as a strong relational
and episodic memory module inside a heterogeneous architecture,
especially where context retrieval and sequence-sensitive binding
dominate the task. It becomes conceptually overextended when treated as
a replacement for cortex as a whole. The next section tests this
boundary in the auditory domain, where spectro-temporal structure
demands an architecture that preserves frequency-time organization
rather than reducing sound to generic token relations.

\begin{center}\rule{0.5\linewidth}{0.5pt}\end{center}

\subsubsection{7.2. What Auditory Cortex Demands: Spectro-Temporal
Architecture}\label{what-auditory-cortex-demands-spectro-temporal-architecture}

Auditory cognition exposes a limit in treating every modality as a token
sequence. Sound does not arrive as a list of interchangeable symbols.
The ear and auditory pathway transform pressure variation into a
structured field of frequency, intensity, and time, where neighboring
frequencies matter, perceptual pitch does not scale linearly with
physical frequency, and many signal identities depend on rapid change
across milliseconds. Speech phonemes, musical timbre, rhythm,
environmental transients, and localization cues all rely on patterns of
spectral energy moving through time. An architecture that flattens this
structure into generic tokens may still learn useful statistical
associations, but it begins by discarding the coordinate system that
makes auditory information intelligible.

Human auditory cortex preserves that coordinate system through tonotopic
organization. Reviews of human auditory mapping show systematic
gradients of frequency preference across auditory cortical fields, not a
flat table of independent spectral bins {[}Saenz and Langers, 2014{]}.
Moerel and colleagues further show that auditory cortex contains
multiple topographic fields with distinct response properties, linking
anatomical organization to functional specialization {[}Moerel, 2014{]}.
This evidence fits the broader cytoarchitectonic lesson that cortical
regions differ because their computational roles differ, rather than
because one generic cortical template has simply been resized {[}Zilles
and Amunts, 2010; Amunts et al., 2020{]}. For audition, the relevant
specialization starts with frequency neighborhoods: nearby frequencies
occupy ordered relations that local circuitry can exploit as a built-in
constraint.

\pandocbounded{\includegraphics[keepaspectratio,alt={(a) Systematic tonotopic gradients visualized as a continuous map across an auditory cortical field; (b) the non-linear transformation from physical frequency (Hz) to perceived pitch, showing the non-uniform functional geometry; (c) a spectro-temporal receptive field (STRF) illustrating the joint tuning of a neural unit to frequency-time patterns. Collectively, these three properties form the fundamental architectural requirements for constructing intelligible auditory objects from complex acoustic signals.}]{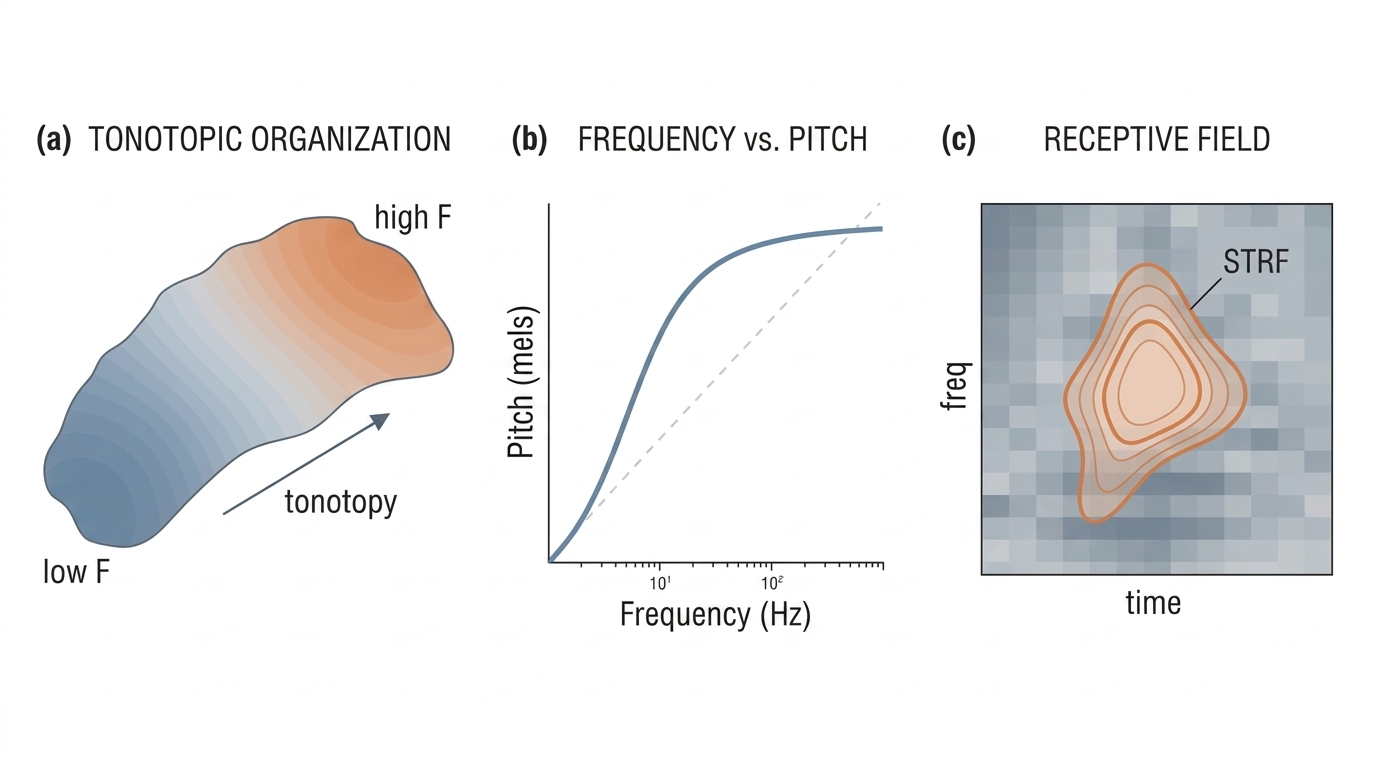}}
\emph{Figure 14. (a) Systematic tonotopic gradients visualized as a
continuous map across an auditory cortical field; (b) the non-linear
transformation from physical frequency (Hz) to perceived pitch, showing
the non-uniform functional geometry; (c) a spectro-temporal receptive
field (STRF) illustrating the joint tuning of a neural unit to
frequency-time patterns. Collectively, these three properties form the
fundamental architectural requirements for constructing intelligible
auditory objects from complex acoustic signals.}

The scale of those neighborhoods also matters. Stevens, Volkmann, and
Newman introduced the mel scale to formalize the fact that perceived
pitch magnitude does not increase linearly with frequency in hertz
{[}Stevens et al., 1937{]}. This does not mean that an artificial
auditory system must copy the mammalian ear or cortex cell by cell. It
means that a competent auditory architecture must preserve the
functional geometry of the signal. Equal steps in physical frequency do
not necessarily correspond to equal steps in auditory similarity, so a
uniform channel index cannot serve as a neutral representation. It
already encodes a hypothesis about what differences matter.

Time imposes an additional constraint. Depireux, Simon, Klein, and
Shamma characterized neurons in ferret primary auditory cortex through
spectro-temporal response fields, showing that neurons respond to joint
patterns of spectral and temporal modulation rather than to frequency
alone {[}Depireux et al., 2001{]}. This finding shifts the computational
unit from a sound feature at a moment to a frequency-time
transformation. Auditory objects often depend on this joint structure: a
consonant may hinge on a fast formant transition, a note on its attack
and decay, and a location on timing and intensity differences across
ears. These are not late semantic interpretations added after hearing.
They are part of the front-end construction of the auditory object
itself.

Current Transformer-based audio systems often handle this problem by
converting the waveform into a spectrogram, cutting that time-frequency
image into patches, and feeding the resulting patches to self-attention.
The Audio Spectrogram Transformer follows this route explicitly by
adapting the patch-tokenization strategy of the Vision Transformer to
spectrograms {[}Dosovitskiy et al., 2021; Gong et al., 2021{]}. This
pipeline can perform well on benchmarks, but benchmark success does not
prove architectural adequacy. A spectrogram patch is a convenient input
format for an attention stack; it is not, by itself, a native auditory
prior. Unless the architecture builds in log-scale frequency relations,
spectro-temporal modulation sensitivity, and multi-window temporal
integration, the model must learn these constraints statistically from
data.

\pandocbounded{\includegraphics[keepaspectratio,alt={(a) An overview of a common Transformer-based audio pipeline, where a waveform is converted into a patch-divided spectrogram and processed with dense self-attention; (b) the proposed specialized auditory front-end architecture, which explicitly incorporates log-scaled spectro-temporal topology, local filters, and long-range temporal integration to construct auditory objects for an episodic memory core; (c) a computational cost graph demonstrating that finer temporal resolution, crucial for audition, leads to a prohibitive increase in attention tokens, thus motivating the specialized, topology-preserving module.}]{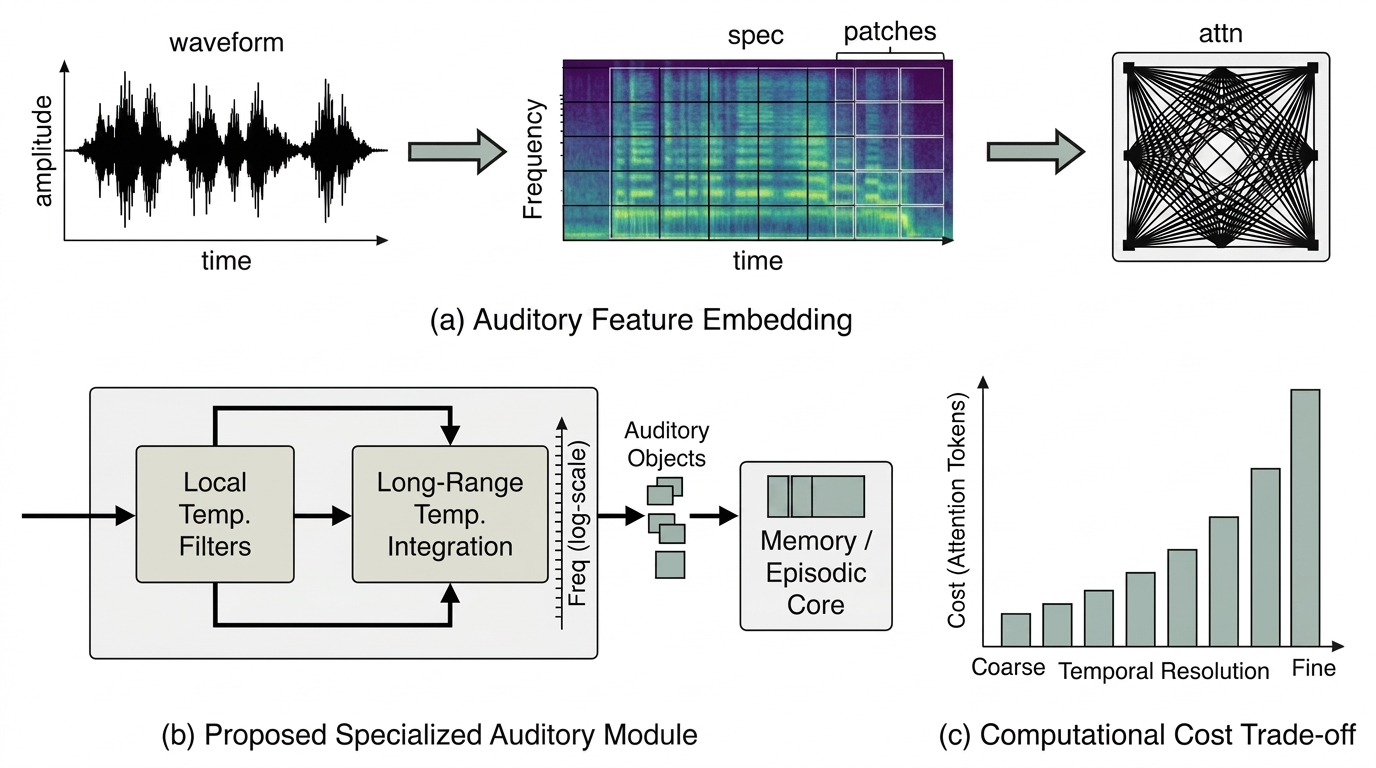}}
\emph{Figure 15. (a) An overview of a common Transformer-based audio
pipeline, where a waveform is converted into a patch-divided spectrogram
and processed with dense self-attention; (b) the proposed specialized
auditory front-end architecture, which explicitly incorporates
log-scaled spectro-temporal topology, local filters, and long-range
temporal integration to construct auditory objects for an episodic
memory core; (c) a computational cost graph demonstrating that finer
temporal resolution, crucial for audition, leads to a prohibitive
increase in attention tokens, thus motivating the specialized,
topology-preserving module.}

Self-attention gives the Transformer its strength: each token can relate
directly to other tokens through learned weights {[}Vaswani et al.,
2017{]}. That mechanism supports relational binding, sequence
association, and memory-like retrieval, and modern Hopfield analyses
strengthen the connection between attention and content-addressable
associative memory {[}Ramsauer et al., 2021{]}. This functional profile
fits the hippocampal interpretation developed in Section 7.1: the
hippocampus supports declarative and relational memory by binding items,
contexts, and sequences into flexible representations {[}Eichenbaum,
2004{]}. A Transformer-like core can therefore connect auditory events
to context, prediction, and episode memory. It can bind a voice to a
speaker, a phrase to a conversation, or a sound to a remembered event.
It should not be mistaken for the best primitive for constructing the
auditory event in the first place.

The systems cost follows directly from the same mismatch. Dense
self-attention grows rapidly as the number of tokens increases, because
each token must compare with many others {[}Vaswani et al., 2017; Sze et
al., 2017{]}. Audio often demands fine temporal resolution, but finer
resolution creates more patches and therefore more attention operations.
Coarsening the patch grid reduces cost, yet it risks losing the fast
temporal modulations that auditory cortex treats as central. A
specialized auditory pathway would instead encode frequency-time
geometry as topology: log-scaled spectral neighborhoods, local and
long-range temporal filters, and receptive fields tuned to modulation
patterns. Such a module could pass already formed auditory objects to an
episodic memory core, rather than forcing a generic attention system to
rediscover the auditory coordinate system from a flattened token stream.

Functionalism supports this division rather than weakening it. Putnam's
multiple realizability allows the same function to appear in different
physical substrates, but only if the realizing system preserves the
causal organization relevant to that function {[}Putnam, 1967; Putnam,
1975{]}. Fodor's modularity thesis likewise treats specialized input
systems as domain-structured processors, not as arbitrary applications
of one undifferentiated mechanism {[}Fodor, 1983{]}. The lesson is
therefore architectural, not imitative: an artificial auditory module
need not be biological, but it must preserve the spectro-temporal
topology that audition requires. The next section extends this boundary
beyond audition to the other missing components of a heterogeneous
architecture: executive gating, working memory maintenance, and
multisensory binding.

\begin{center}\rule{0.5\linewidth}{0.5pt}\end{center}

\subsubsection{7.3. The Missing Modules: Executive Gating, Working
Memory, Multisensory
Binding}\label{the-missing-modules-executive-gating-working-memory-multisensory-binding}

Once the Transformer is localized as a hippocampal-like architecture,
its limits become easier to state precisely. Self-attention can bind
relations across a sequence by letting each token compute weighted
access to other tokens, and modern Hopfield interpretations make this
resemblance to associative retrieval especially explicit {[}Vaswani et
al., 2017; Ramsauer et al., 2021{]}. This makes attention a powerful
mechanism for relational organization, pattern completion, and ordered
prediction, all functions that overlap with hippocampal accounts of
declarative and episodic memory {[}Eichenbaum, 2004; O'Keefe and Nadel,
1978; Buzsáki and Tingley, 2018{]}. But a mechanism that retrieves and
recombines relations does not automatically decide when retrieval should
occur, what should be suppressed, what should remain actively
maintained, or whether signals from different senses belong to the same
external event. Those are different computational problems, not merely
larger versions of sequence modeling.

Functionalism clarifies why this distinction matters. Putnam's multiple
realizability thesis rejects the idea that one biological material is
uniquely required for a mental function, but it does not imply that any
causal topology can realize any function equally well {[}Putnam, 1967;
Putnam, 1975{]}. Marr's levels sharpen the point: a system must be
evaluated not only by its input-output behavior, but also by the
computational problem it solves and the algorithmic organization that
solves it {[}Marr, 1982{]}. Craver's account of mechanistic explanation
reaches the same conclusion from the biological side: explaining a
capacity requires identifying organized components whose activities
produce that capacity, not merely showing that a large system can
approximate the behavior {[}Craver, 2007{]}. A homogeneous attention
stack may approximate many tasks statistically, but approximation does
not prove that it contains the control, maintenance, and binding
mechanisms those tasks require.

Executive gating supplies the first missing mechanism. In biological
systems, control does not work by sending every candidate representation
into a single associative store. Basal ganglia-thalamocortical circuits
form parallel loops that can facilitate or inhibit distinct motor,
cognitive, and limbic operations {[}Alexander et al., 1986{]}. Frank's
dopamine-based models describe this architecture as a Go/No-Go selection
system in which reward-sensitive signals regulate whether a
representation or action gains access to downstream processing {[}Frank,
2005{]}. Frank et al.~extend the same logic to working-memory updating:
the basal ganglia help determine when frontal representations should
change and when they should remain stable {[}Frank et al., 2001{]}.
Vanilla self-attention has no separable equivalent of this gate. Its
attention weights rank relations inside the same operation that
retrieves information; they do not constitute an independent control
circuit that can withhold, release, or reward-modulate access to a
maintained state.

\pandocbounded{\includegraphics[keepaspectratio,alt={Executive control architecture via a cortico-BG-thalamic loop. The diagram illustrates a circuit where parallel Go/No-Go pathways, originating from the basal ganglia (BG), regulate information flow between the prefrontal cortex (PFC) and thalamus (THAL). Go signals facilitate access, promoting active maintenance and updating of representations in the PFC, while No-Go signals inhibit access and enforce a gating block, protecting stable states from interference. This causal topology provides a dedicated substrate for context-specific executive control that is distinct from general associative memory.}]{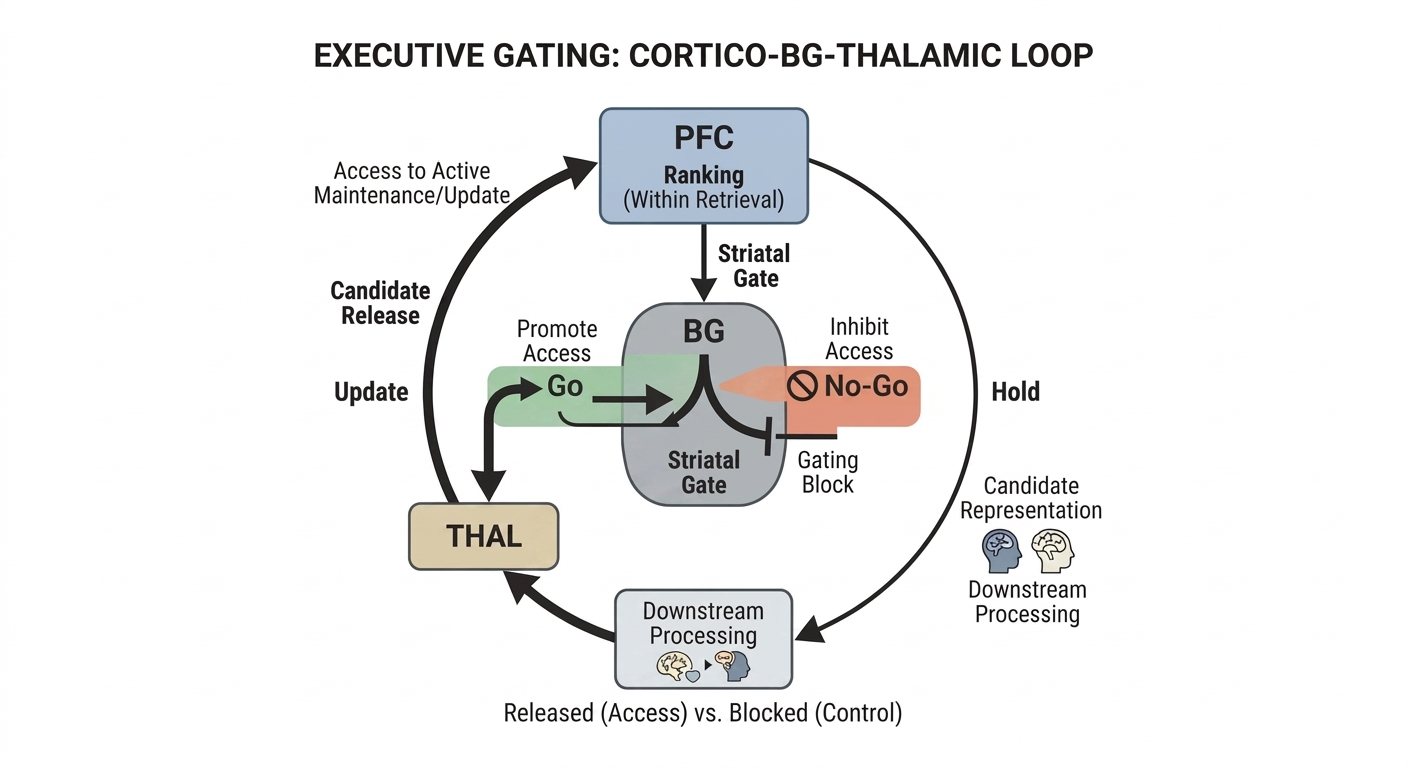}}
\emph{Figure 16. Executive control architecture via a
cortico-BG-thalamic loop. The diagram illustrates a circuit where
parallel Go/No-Go pathways, originating from the basal ganglia (BG),
regulate information flow between the prefrontal cortex (PFC) and
thalamus (THAL). Go signals facilitate access, promoting active
maintenance and updating of representations in the PFC, while No-Go
signals inhibit access and enforce a gating block, protecting stable
states from interference. This causal topology provides a dedicated
substrate for context-specific executive control that is distinct from
general associative memory.}

Working memory therefore cannot be reduced to a longer context window.
Baddeley defines working memory as an active system for temporary
maintenance and manipulation under task demands, not as passive access
to a larger store {[}Baddeley, 2003{]}. Neurophysiology supports this
distinction: Funahashi et al.~recorded delay-period activity in monkey
dorsolateral prefrontal cortex, showing that neurons can maintain
information about a remembered location after the stimulus disappears
{[}Funahashi et al., 1989{]}. This persistent activity functions like a
protected task state. It keeps selected variables available for
decision, control, and report while resisting interference. A
Transformer's context or key-value cache preserves prior token
representations for later retrieval, but persistence by accumulation
differs from selective maintenance. Increasing context length expands
what the model can consult, and it also increases memory traffic, a
major energy cost in neural-network hardware {[}Sze et al., 2017{]}. It
still does not add the dedicated update-and-protect logic that working
memory requires.

\pandocbounded{\includegraphics[keepaspectratio,alt={Contrasting passive context storage with active working-memory maintenance. (a) The Transformer context window functions as a passive cache, linearly accumulating an input stream into a stored buffer for later access without selective protection. (b) A working-memory system implements active maintenance by using control signals (gates) to update a protected state selectively, while interference shields guard this state from distraction.}]{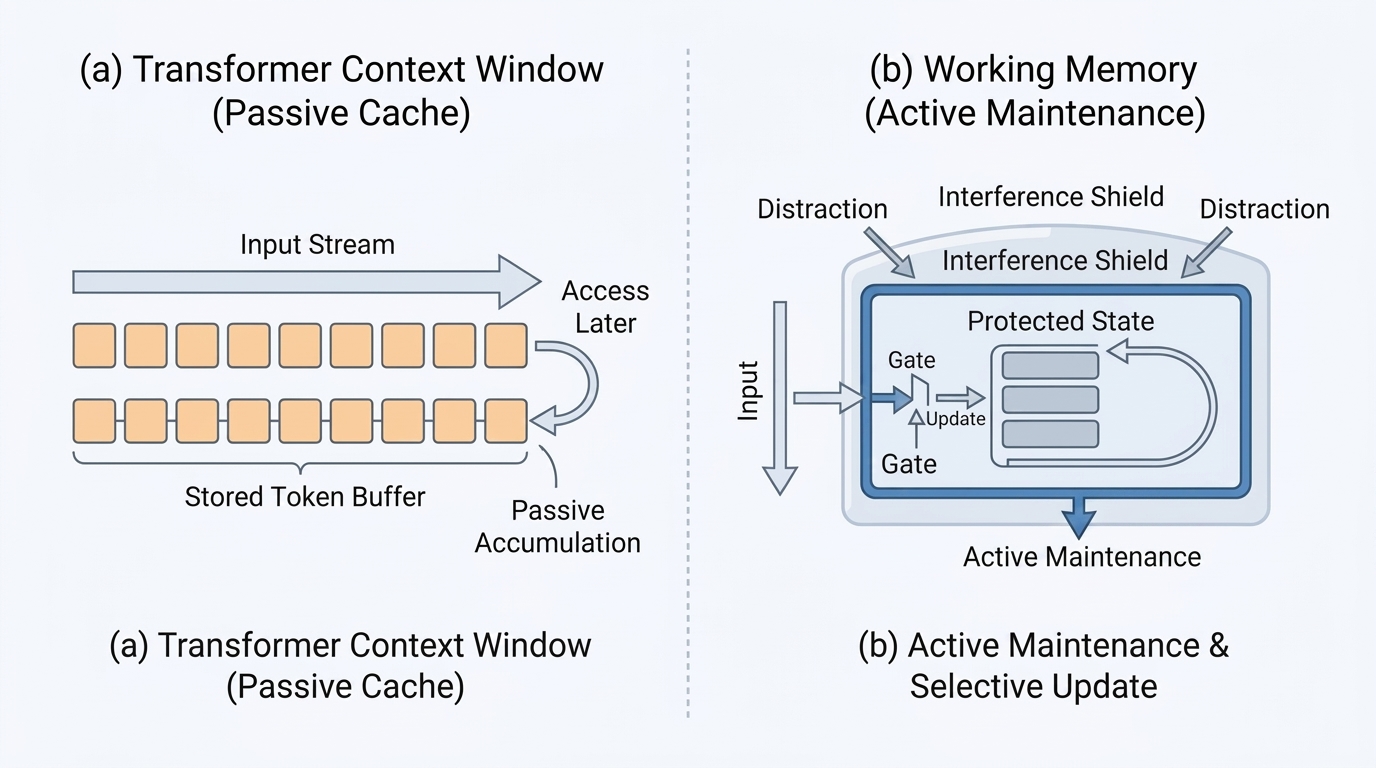}}
\emph{Figure 17. Contrasting passive context storage with active
working-memory maintenance. (a) The Transformer context window functions
as a passive cache, linearly accumulating an input stream into a stored
buffer for later access without selective protection. (b) A
working-memory system implements active maintenance by using control
signals (gates) to update a protected state selectively, while
interference shields guard this state from distraction.}

Multisensory binding introduces a third missing topology. The brain does
not solve cross-modal perception by simply concatenating visual,
auditory, and somatosensory features into one undifferentiated vector
space. Multisensory circuits combine inputs under spatial, temporal, and
reliability constraints; superior-colliculus and cortical studies show
that responses depend on whether signals plausibly originate from the
same event {[}Stein and Stanford, 2008{]}. This creates a real-time
correspondence problem. A binding module must preserve modality-specific
maps, compare timing windows, estimate cross-modal congruence, and
decide whether to fuse or separate signals. Standard shared-attention
fusion can learn correlations among modalities, but correlation alone
does not impose the reliability-weighted and timing-sensitive rules that
biological multisensory integration displays.

\pandocbounded{\includegraphics[keepaspectratio,alt={Schematic of multisensory binding as a computational problem, illustrating how modality-specific streams are integrated by a distinct central system rather than through simple concatenation. Modality-specific inputs (Vision/V1, Audition/A1, Somatosensory/S1) with topographic maps feed into a Central Binding Module, which evaluates the signals based on Time, Space, and Reliability. This comparison process governs a decision node that either fuses the inputs into a unified percept or maintains them as separate representations. The figure supports the argument that multisensory integration requires dedicated, rule-based topology rather than being an emergent property of general attention layers.}]{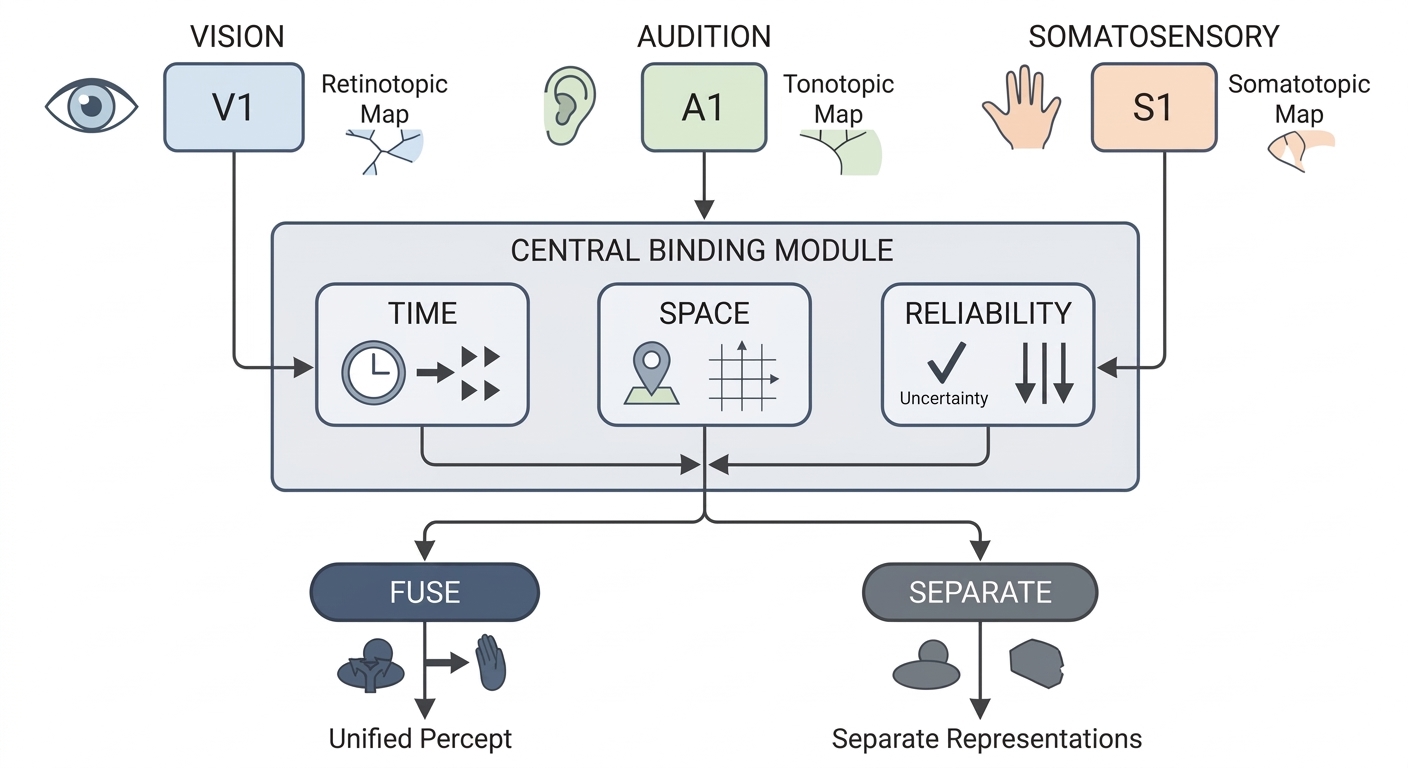}}
\emph{Figure 18. Schematic of multisensory binding as a computational
problem, illustrating how modality-specific streams are integrated by a
distinct central system rather than through simple concatenation.
Modality-specific inputs (Vision/V1, Audition/A1, Somatosensory/S1) with
topographic maps feed into a Central Binding Module, which evaluates the
signals based on Time, Space, and Reliability. This comparison process
governs a decision node that either fuses the inputs into a unified
percept or maintains them as separate representations. The figure
supports the argument that multisensory integration requires dedicated,
rule-based topology rather than being an emergent property of general
attention layers.}

Fodor's modularity thesis offers the right caution: specialization
should not mean sealed boxes {[}Fodor, 1983{]}. Executive gating,
working memory, and multisensory binding must interact with relational
memory, perception, and action. The design lesson is therefore not
isolation, but interoperable heterogeneity. A hippocampal-like
Transformer core can remain valuable as an episodic and relational
memory module, but it should connect to distinct systems for
reward-sensitive gating, protected task-state maintenance, and
cross-modal event binding. Chapter 8 converts this localization argument
into a concrete System-of-Systems proposal, where modules communicate
through explicit interfaces while retaining the structural biases their
functions demand.

\pandocbounded{\includegraphics[keepaspectratio,alt={Schematic architecture illustrating the functional localization of complementary cognitive components interacting with a central Transformer core. The diagram details how a central system, characterized by relational memory and self-attention, must be augmented by distinct, interconnected modules to achieve higher-level functions: a gate module for executive control signals, a separate working memory (WM) system for protected maintenance and selective retrieval, and a bind module for cross-modal integration.}]{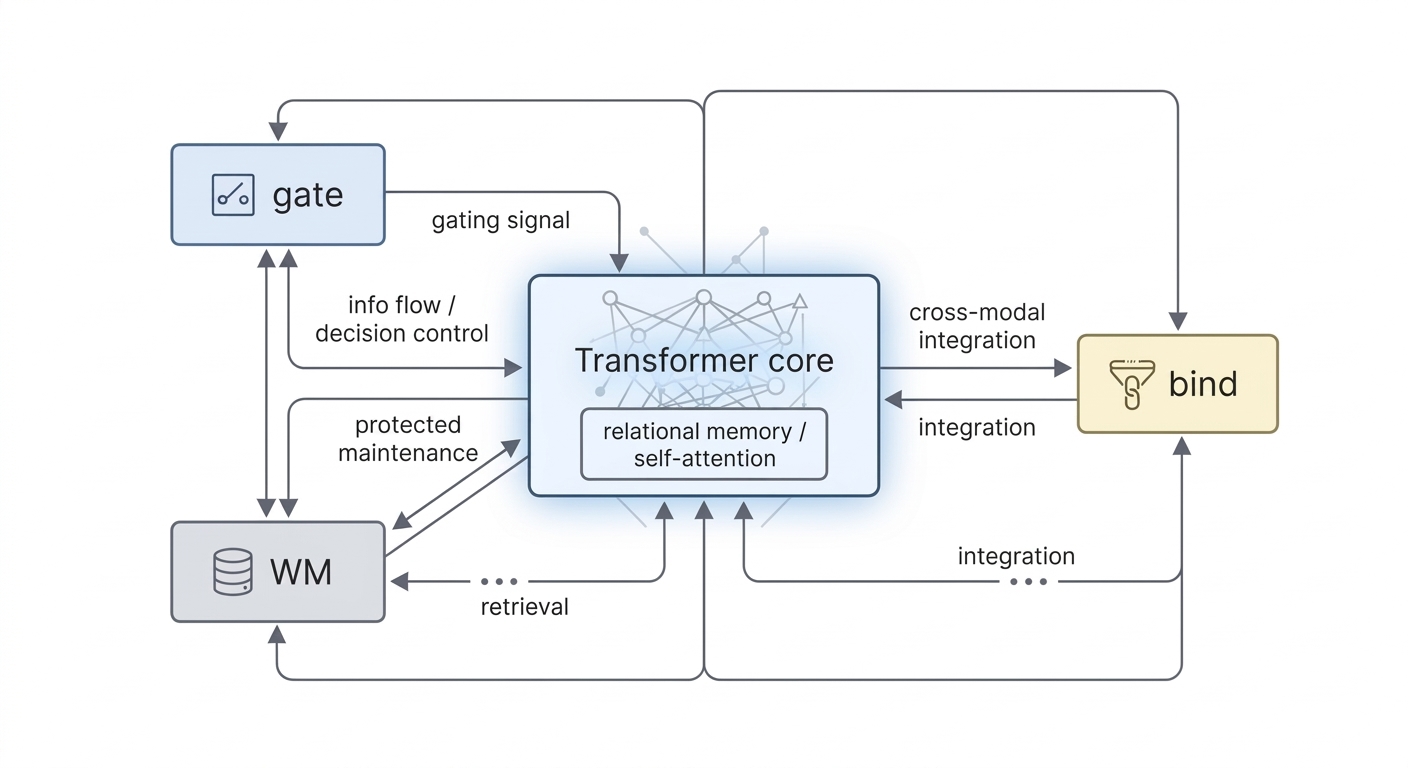}}
\emph{Figure 19. Schematic architecture illustrating the functional
localization of complementary cognitive components interacting with a
central Transformer core. The diagram details how a central system,
characterized by relational memory and self-attention, must be augmented
by distinct, interconnected modules to achieve higher-level functions: a
gate module for executive control signals, a separate working memory
(WM) system for protected maintenance and selective retrieval, and a
bind module for cross-modal integration.}

\begin{center}\rule{0.5\linewidth}{0.5pt}\end{center}

\section{Chapter 8. Conclusion: Toward a Heterogeneous System of
Systems}\label{chapter-8.-conclusion-toward-a-heterogeneous-system-of-systems}

\subsubsection{8.1. Cognitive Science as Structural Architect, Not
Post-Hoc
Observer}\label{cognitive-science-as-structural-architect-not-post-hoc-observer}

Cognitive science should not enter AI design only after a model has been
trained, when researchers attach behavioral labels to whatever
competencies emerge. Its stronger role is architectural: it can specify
in advance what kinds of information flow, memory persistence, routing,
gating, and cross-modal binding a system must contain if it is to
realize a target cognitive function. Marr's distinction among
computational, algorithmic, and implementational levels remains the
clearest discipline for this task {[}Marr, 1982{]}. A cognitive system
first requires a statement of the problem being solved, then an account
of the representations and procedures that could solve it, and only then
a physical or machine substrate that realizes those procedures. This
ordering prevents implementation convenience from masquerading as
cognitive explanation.

\pandocbounded{\includegraphics[keepaspectratio,alt={Canonical comparison of scaling laws versus architecture-first design. (a) Loss improves with scale following a power law but plateaus as a fixed architecture exhausts its capacity, indicating that repeating homogeneous units (left stack) expands capacity within an existing regime, while qualitatively new computational topologies (right graph) are required for qualitatively new functions; (b) a conventional workflow fixes a homogeneous architecture first and evaluates post-hoc, whereas an architecture-first workflow begins with explicit functional decomposition into distinct sensory, gating, and memory modules connected by explicit interfaces, and finally matches diverse hardware (dense/sparse) to these functional requirements.}]{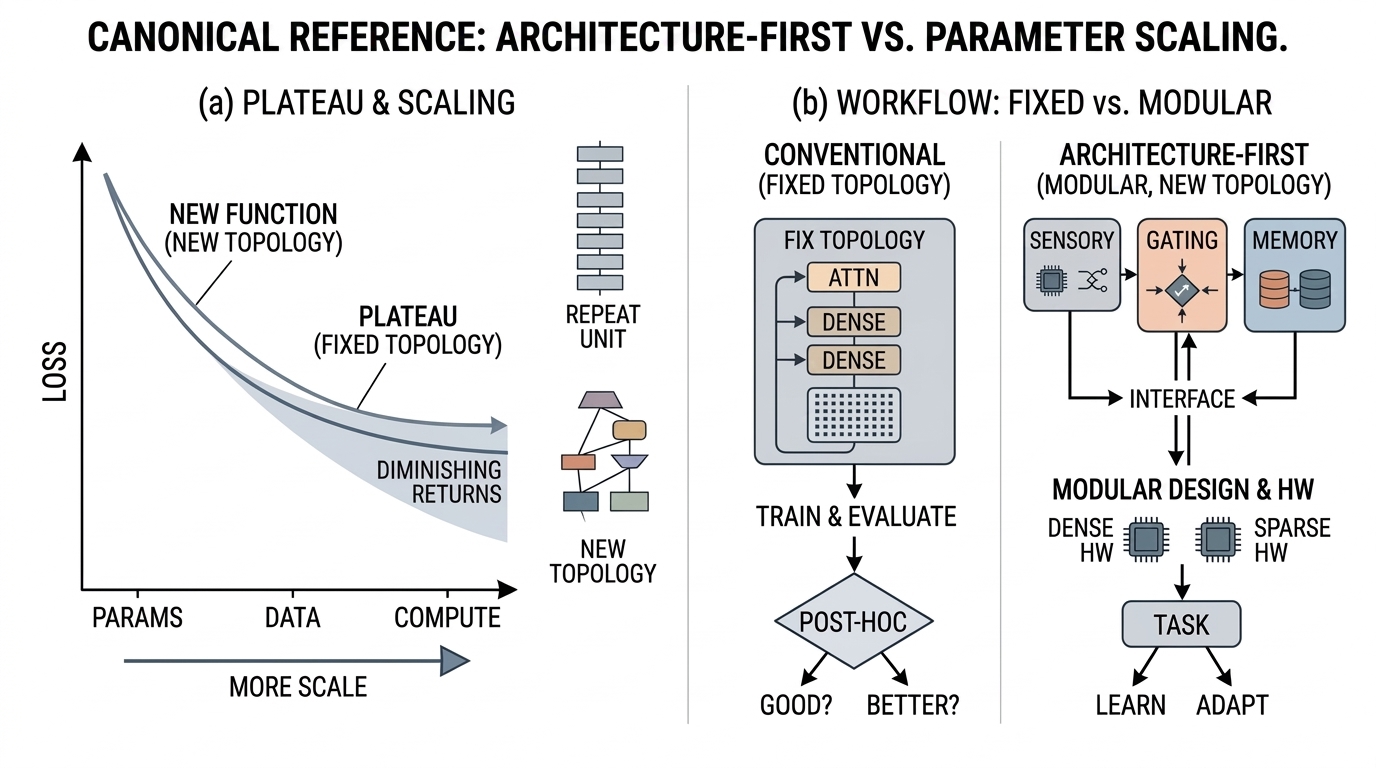}}
\emph{Figure 20. Canonical comparison of scaling laws versus
architecture-first design. (a) Loss improves with scale following a
power law but plateaus as a fixed architecture exhausts its capacity,
indicating that repeating homogeneous units (left stack) expands
capacity within an existing regime, while qualitatively new
computational topologies (right graph) are required for qualitatively
new functions; (b) a conventional workflow fixes a homogeneous
architecture first and evaluates post-hoc, whereas an architecture-first
workflow begins with explicit functional decomposition into distinct
sensory, gating, and memory modules connected by explicit interfaces,
and finally matches diverse hardware (dense/sparse) to these functional
requirements.}

Biology shows why this ordering matters. Cortical organization is not an
undifferentiated sheet whose internal divisions become meaningful only
after behavior has been measured. Brodmann's parcellation used
differences in cell size, density, and laminar arrangement as anatomical
criteria for distinguishing cortical fields, and later cytoarchitectonic
work preserved the central lesson that cortical areas differ in
measurable structure, not merely in behavioral association {[}Brodmann,
1909; Zilles, 2010{]}. The Jülich Brain Atlas extends this tradition
into a three-dimensional probabilistic reference system, showing that
cytoarchitectonic boundaries can be mapped as statistically
characterized features of human cortex {[}Amunts et al., 2020{]}. These
maps do not provide a literal engineering blueprint for AI, but they
establish a constraint that architecture design cannot ignore:
functional decomposition begins from organized structure, not from a
homogeneous substrate interpreted only after output behavior appears.

The same point holds below and above the cortical-area scale.
Mountcastle's account of columnar organization and Douglas and Martin's
canonical neocortical circuit identify recurring motifs in cortical
processing, but recurring motifs do not imply total cortical uniformity
{[}Mountcastle, 1997; Douglas and Martin, 2004{]}. Harris and Shepherd
emphasize that cortex combines shared circuit themes with
region-specific variation, so the correct inference is conditional
rather than homogenizing: a system may reuse local operations while
still requiring distinct laminar proportions, input-output patterns,
recurrent dynamics, and projection targets across domains {[}Harris and
Shepherd, 2015{]}. Large-scale organization reinforces this conclusion.
The primate visual system contains a distributed hierarchy of
feedforward, feedback, and lateral pathways rather than a single
interchangeable processing sheet {[}Felleman and Van Essen, 1991{]}.
Basal ganglia-cortical circuits form parallel, functionally segregated
loops for motor, cognitive, and control functions {[}Alexander et al.,
1986{]}. Working-memory models likewise treat temporary maintenance and
manipulation as separable components of cognitive architecture rather
than as residual properties of general intelligence {[}Baddeley,
2003{]}.

Functionalism supports this design discipline when it is not confused
with architectural indifference. Putnam's multiple realizability thesis
rejects the claim that a mental function must be tied to one biological
material substrate {[}Putnam, 1967{]}. It does not follow that every
topology can realize every function with equal adequacy. A silicon
system need not copy neurons, layers, nuclei, or synapses one-to-one,
but it must still instantiate the causal organization demanded by the
function. Declarative memory requires structures capable of stabilizing
and recombining relational representations across experience
{[}Eichenbaum, 2004{]}. Executive control requires mechanisms for
selecting, inhibiting, and updating actions or policies under changing
conditions, a role basal-ganglia models describe in terms of
reinforcement-sensitive gating {[}Frank, 2005{]}. Multisensory binding
requires interfaces that preserve timing, salience, and correspondence
across channels rather than merely concatenating modality tokens
{[}Stein and Stanford, 2008{]}. Multiple realizability removes
biological material as a fixed constraint; it does not remove
organization as a constraint.

This distinction exposes the methodological inversion in contemporary
AI. Architecture selection usually fixes a topology first, most often
the attention-based Transformer stack, and evaluates its adequacy
afterward through benchmarks or post-hoc analysis {[}Vaswani et al.,
2017; Bommasani et al., 2021{]}. Scaling laws demonstrate that loss can
improve predictably as data, parameters, and compute increase under a
fixed modeling regime, but they do not show that the fixed computational
graph is the right hypothesis about the cognitive function being modeled
{[}Kaplan et al., 2020{]}. A self-attention layer can compare each token
with every other token, which makes it powerful for global relational
binding, but that same design does not by itself specify whether a task
instead demands sparse local sensing, hierarchical sensory compression,
persistent working-memory slots, or reinforcement-gated routing
{[}Vaswani et al., 2017{]}. Post-hoc interpretability can reveal
patterns a trained network has learned, but it cannot retroactively
supply architectural commitments that were absent from the search space.

Systems engineering gives the same lesson in another vocabulary. An
interface contract defines what information a component receives, what
format it passes onward, and what communication cost the system must
pay. Mixture-of-Experts models improve capacity by routing tokens to
subsets of experts, but the experts typically remain architecturally
similar feed-forward blocks; the diversity is mainly a partitioning of
parameters rather than a principled decomposition of cognitive functions
{[}Shazeer et al., 2017; Fedus et al., 2022{]}. The routing itself
creates distributed-systems costs that training frameworks must
explicitly manage across devices {[}Narayanan et al., 2021{]}.
Conversely, neuromorphic systems such as Loihi begin from a different
information-flow hypothesis: asynchronous event-driven spikes rather
than dense synchronous tensors {[}Davies et al., 2018{]}. That choice
matters because data movement and memory access often dominate energy
cost in machine-learning hardware, making interface design a first-order
architectural decision rather than an implementation detail {[}Sze et
al., 2017{]}. Hardware should therefore implement a cognitive
architecture; it should not silently choose one.

Cognitive science supplies the missing bridge between biological
specialization and engineering design. It can determine which biological
details can be abstracted away, which functional distinctions must
survive substrate transfer, and which boundaries must reappear as
topology, memory format, or communication protocol. Its task is not to
decorate a trained foundation model with neuroscientific labels. Its
task is to define, before scale is applied, the minimal set of
structurally distinct modules required by the cognitive functions at
issue. Section 8.2 turns this rule into a concrete heterogeneous system:
separable modules with domain-appropriate priors, explicit interfaces,
and constraints strong enough to prevent the architecture from
collapsing back into a monolithic Transformer.

\begin{center}\rule{0.5\linewidth}{0.5pt}\end{center}

\subsubsection{8.2. A Concrete Proposal: Heterogeneous Topological
Networks Beyond the
Transformer}\label{a-concrete-proposal-heterogeneous-topological-networks-beyond-the-transformer}

The concrete alternative to Transformer monoculture is a Heterogeneous
Topological Network: a system of systems in which each module has its
own computational topology, its own inductive bias, and a standardized
interface for communication with other modules. The central design rule
is simple but demanding: shared communication does not require shared
internal architecture. A visual pathway, an auditory pathway, an
episodic memory core, an executive gate, a working-memory buffer, and a
multisensory binding layer can exchange representations through explicit
contracts without processing all information through the same attention
stack.

This proposal follows from functionalism only when functionalism is
interpreted correctly. Putnam's multiple realizability thesis denies
that a mental function must be tied to one biological substrate, but it
does not deny that a function requires an appropriate causal
organization {[}Putnam, 1967; Putnam, 1975{]}. Marr's analysis of vision
makes the same point computationally: a system must specify the problem,
the representation and algorithm, and the physical implementation that
makes the computation tractable {[}Marr, 1982{]}. Craver's mechanistic
account sharpens the argument further by showing that a capacity becomes
intelligible only when it is decomposed into organized components whose
activities and interactions produce the target phenomenon {[}Craver,
2007{]}. Architecture is therefore not a neutral container into which
scale can be poured. It is part of the explanation.

\pandocbounded{\includegraphics[keepaspectratio,alt={A comparative analysis of computational substrates and an experimental pipeline for a Heterogeneous Topological Network (HTN). Panel (a) shows a conventional dense GPU substrate reliant on uniform, globally synchronized matrix multiplication tiles; panel (b) illustrates a neuromorphic substrate characterized by sparse, event-driven asynchronous spiking; and panel (c) presents the experimental architecture for an HTN, which connects functionally distinct modules (e.g., CNN and SNN) via a standardized interface layer and evaluates its performance and routing failure diagnostics against a homogeneous Transformer baseline.}]{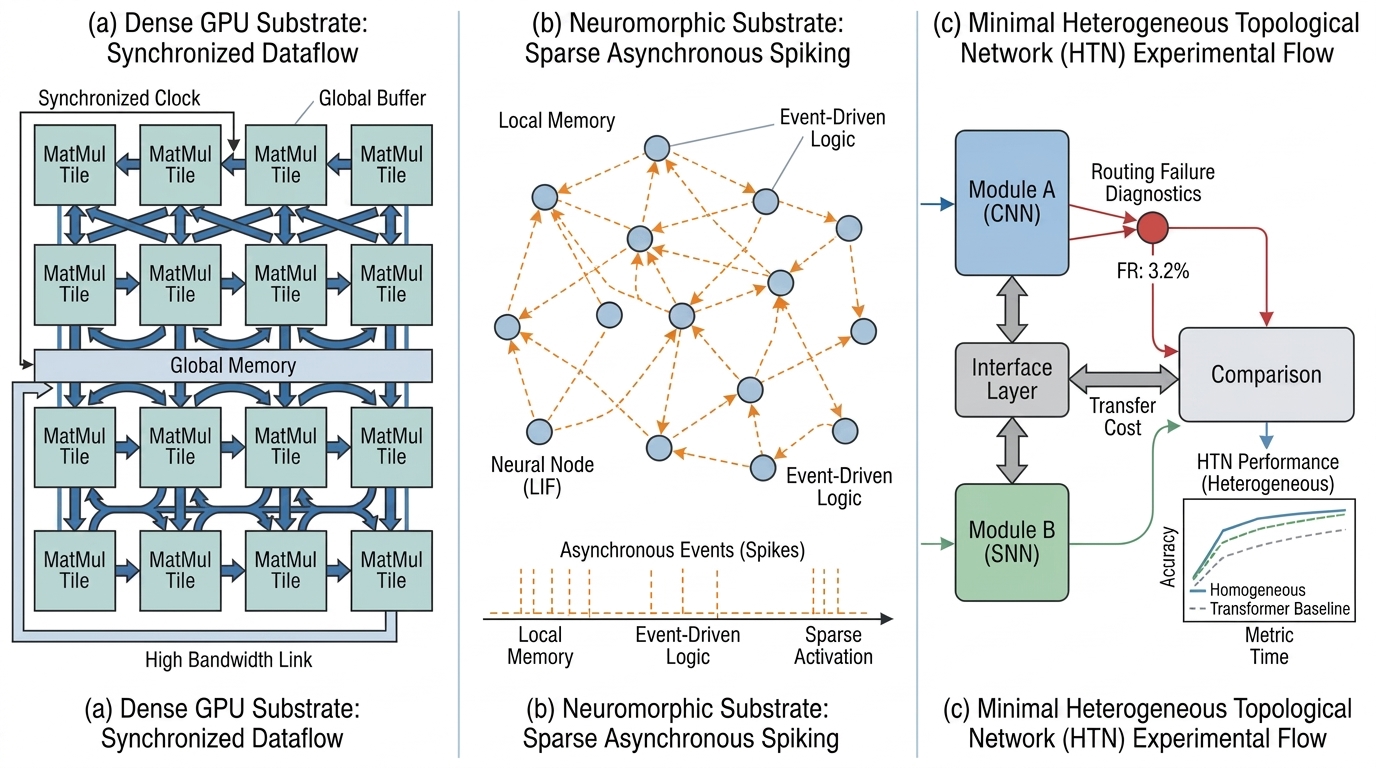}}
\emph{Figure 21. A comparative analysis of computational substrates and
an experimental pipeline for a Heterogeneous Topological Network (HTN).
Panel (a) shows a conventional dense GPU substrate reliant on uniform,
globally synchronized matrix multiplication tiles; panel (b) illustrates
a neuromorphic substrate characterized by sparse, event-driven
asynchronous spiking; and panel (c) presents the experimental
architecture for an HTN, which connects functionally distinct modules
(e.g., CNN and SNN) via a standardized interface layer and evaluates its
performance and routing failure diagnostics against a homogeneous
Transformer baseline.}

Neuroscience provides the biological warrant for this design principle.
Quantitative cytoarchitecture shows that cortical regions differ in
laminar structure, cellular composition, and spatial organization rather
than instantiating a single repeated cortical template {[}Zilles and
Amunts, 2010; Amunts et al., 2020{]}. Transcriptomic and Patch-seq
studies extend that conclusion to single-cell identity: neuronal classes
differ jointly in morphology, electrophysiology, and gene expression, so
circuit function depends on structured combinations of cellular
properties rather than on the number of generic units alone {[}Hawrylycz
et al., 2012; Cadwell et al., 2016; Gouwens et al., 2020{]}. A
biologically grounded artificial system need not copy biology cell by
cell, but it should preserve the lesson that different cognitive
functions impose different structural constraints.

\pandocbounded{\includegraphics[keepaspectratio,alt={An overview of specialized cognitive modules supporting Heterogeneous Topological Networks by using distinct internal topologies and processing contracts. (a) A hierarchical visual pathway with coordinate-preserving maps and increasing receptive field scales; (b) an auditory pathway with frequency-sensitive, modulation-sensitive structure; (c) an episodic memory core that stores and retrieves relational graphs of items and context; (d) an executive gate that uses sparse route, inhibit, and allow signals; (e) a working memory unit maintaining a stable, updateable state; and (f) a multisensory binding layer that aligns compatible visual and auditory streams based on spatio-temporal and reliability constraints.}]{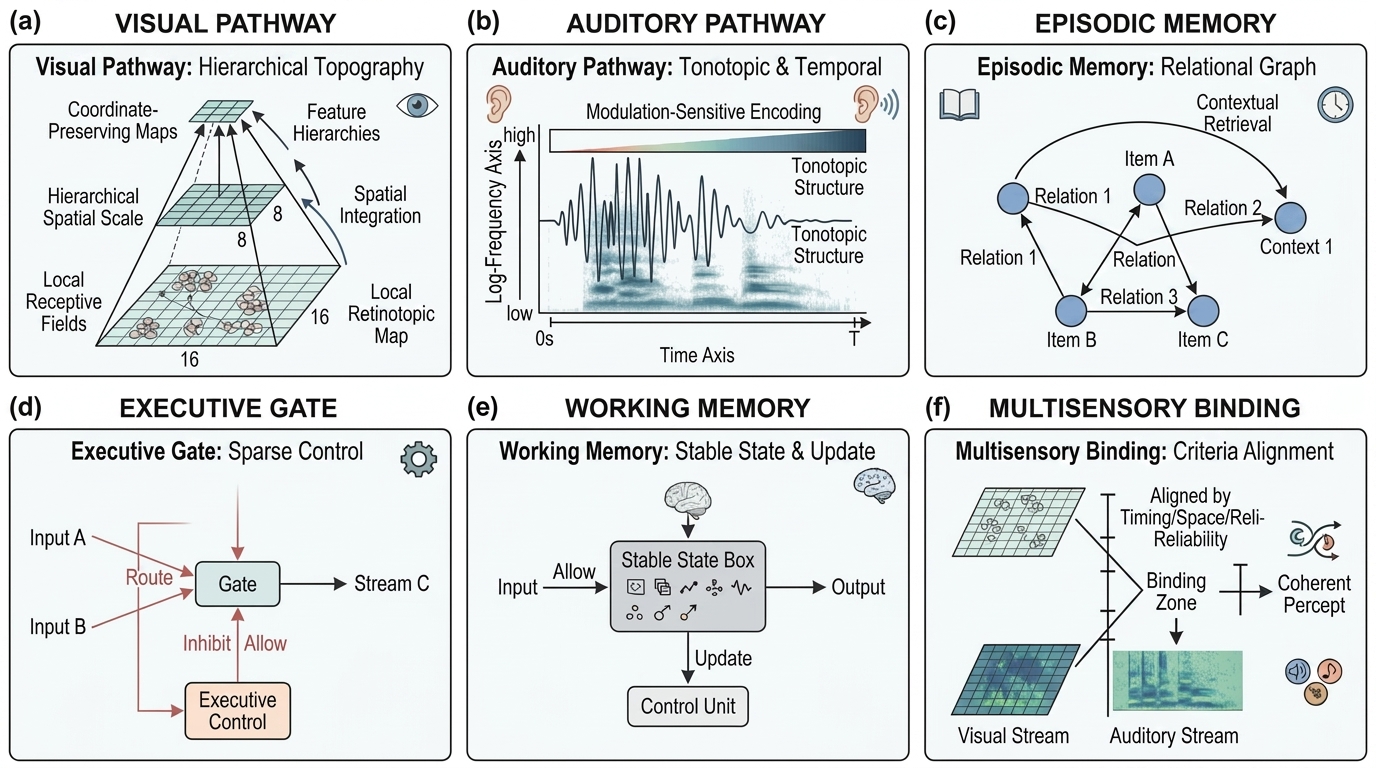}}
\emph{Figure 22. An overview of specialized cognitive modules supporting
Heterogeneous Topological Networks by using distinct internal topologies
and processing contracts. (a) A hierarchical visual pathway with
coordinate-preserving maps and increasing receptive field scales; (b) an
auditory pathway with frequency-sensitive, modulation-sensitive
structure; (c) an episodic memory core that stores and retrieves
relational graphs of items and context; (d) an executive gate that uses
sparse route, inhibit, and allow signals; (e) a working memory unit
maintaining a stable, updateable state; and (f) a multisensory binding
layer that aligns compatible visual and auditory streams based on
spatio-temporal and reliability constraints.}

The visual module should therefore keep a locality-preserving topology.
Primary visual cortex contains organized receptive fields and laminar
circuitry suited to local spatial analysis {[}Hubel and Wiesel, 1962;
Lund, 1988{]}, while primate visual pathways arrange areas
hierarchically across changing receptive-field scales and functional
roles {[}Felleman and Van Essen, 1991{]}. Motion-sensitive cortex such
as MT further shows that even vision is not a homogeneous computation;
motion integration requires specialized circuitry and physiology {[}Born
and Bradley, 2005{]}. In engineering terms, convolutional and related
locality-preserving systems encode this prior directly. A convolutional
layer applies small kernels across neighboring positions, so its cost
scales with local filter size and feature-map area, whereas global
self-attention compares every token with every other token and therefore
scales quadratically with sequence length {[}Fukushima, 1980; LeCun et
al., 1998; Vaswani et al., 2017{]}. AlexNet's ImageNet success showed
the efficiency of a strong spatial prior, while the original Vision
Transformer required very large-scale pretraining to match strong
convolutional baselines at comparable model sizes {[}Krizhevsky et al.,
2012; Dosovitskiy et al., 2021{]}. Attention can process images, but
removing spatial structure shifts part of the burden from architecture
onto data and compute.

The auditory module should not inherit visual patching by default. Human
auditory cortex contains tonotopic maps, and auditory areas organize
computation around frequency-sensitive topographies {[}Saenz and
Langers, 2014; Moerel, 2014{]}. Primary auditory neurons can be
characterized by spectro-temporal response fields, which means that
auditory processing depends on joint structure across frequency and time
rather than on static spatial neighborhoods alone {[}Depireux et al.,
2001{]}. An artificial auditory pathway should therefore use
log-frequency, time-resolved, and modulation-sensitive representations
before any global relational mechanism receives its input. Audio
Spectrogram Transformer demonstrates that spectrograms can be tokenized
and processed by attention, but that strategy applies fixed patch
attention across time-frequency bins rather than making spectro-temporal
organization the first architectural constraint {[}Gong et al., 2021{]}.

The Transformer should remain in the system, but as an
episodic-relational memory core rather than as a universal cortex.
Attention-based architectures excel at relational access over token-like
representations, and modern Hopfield analyses show that attention can
operate as an associative memory mechanism over stored patterns
{[}Ramsauer et al., 2021{]}. That computational profile resembles
hippocampal functions more closely than it resembles all-purpose
cortical computation. The hippocampal formation supports declarative
relational memory, cognitive mapping, temporal sequence organization,
and pattern separation {[}O'Keefe and Nadel, 1978; Eichenbaum, 2004;
Yassa and Stark, 2011; Buzsáki and Tingley, 2018{]}. In the proposed
system, the Transformer core binds items, relations, and temporal
context over a bounded episodic buffer. Its quadratic cost becomes
tractable by architectural assignment rather than by treating every
modality as a long sequence of interchangeable tokens {[}Vaswani et al.,
2017{]}.

\pandocbounded{\includegraphics[keepaspectratio,alt={A comparative view of AI system design principles contrasting monolithic and modular architectures. (a) Uniform Transformer Stack, where distinct inputs like vision and audio are all processed through a single uniform attention mechanism. (b) Heterogeneous Topological Network, a modular alternative in which functionally specialized components, such as a visual pathway, auditory pathway, and episodic memory core, retain distinct internal topologies and communicate via a standardized interface layer.}]{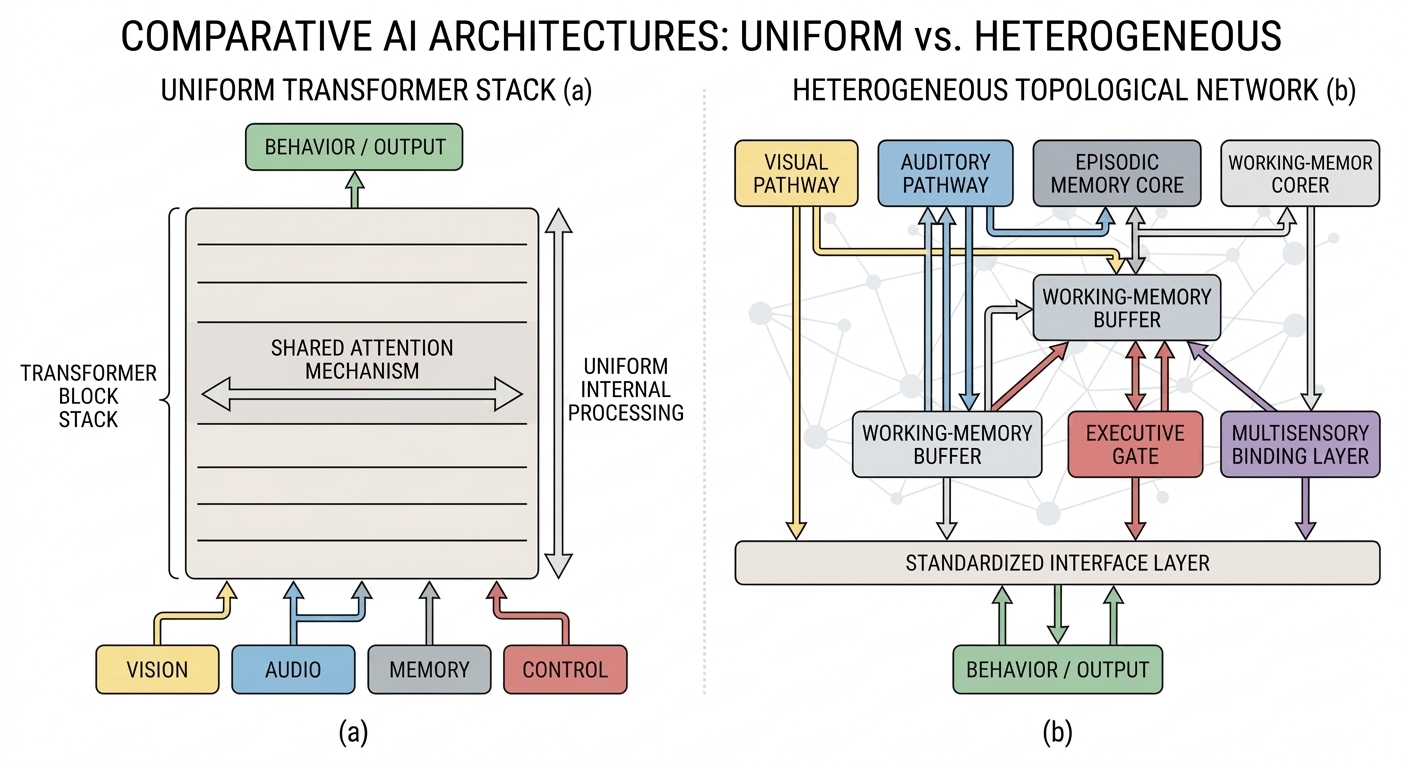}}
\emph{Figure 23. A comparative view of AI system design principles
contrasting monolithic and modular architectures. (a) Uniform
Transformer Stack, where distinct inputs like vision and audio are all
processed through a single uniform attention mechanism. (b)
Heterogeneous Topological Network, a modular alternative in which
functionally specialized components, such as a visual pathway, auditory
pathway, and episodic memory core, retain distinct internal topologies
and communicate via a standardized interface layer.}

Executive control and working memory require separate causal
organizations. Basal ganglia-cortical circuits form parallel,
functionally segregated loops, and computational models of
dopamine-dependent basal ganglia function explain how gating can
regulate action selection and working-memory updating {[}Alexander et
al., 1986; Frank et al., 2001; Frank, 2005{]}. The executive module in a
Heterogeneous Topological Network should therefore route, inhibit, and
release candidate operations through sparse control signals. This
differs from Mixture-of-Experts routing in an important way. Sparsely
gated Mixture-of-Experts models and Switch Transformers route tokens
among expert blocks to increase parameter count while keeping per-token
computation manageable, but the experts usually remain architecturally
similar components inside the same broad Transformer regime {[}Shazeer
et al., 2017; Fedus et al., 2022{]}. Heterogeneous routing instead
selects among different computational kinds: a visual hierarchy, an
auditory analyzer, an episodic memory core, a working-memory buffer, or
a binding interface.

Working memory should maintain task-relevant variables over controlled
intervals rather than collapse into episodic retrieval. Dorsolateral
prefrontal neurons show mnemonic delay-period activity, and cognitive
models distinguish temporary maintenance and manipulation from long-term
associative memory {[}Funahashi et al., 1989; Baddeley, 2003{]}. A
working-memory buffer should therefore expose a small, stable, updatable
state to the executive gate. Multisensory integration also deserves its
own constrained interface. Single-neuron and circuit studies show that
cross-modal binding follows spatial, temporal, and response constraints
rather than arbitrary concatenation {[}Stein and Stanford, 2008{]}. The
multisensory module should bind signals only when their coordinate
systems and reliability profiles justify integration.

The interface contract is the practical core of the minimal viable
prototype. Visual outputs should preserve spatial coordinates and
hierarchical scale; auditory outputs should preserve frequency, time,
and modulation structure; the episodic core should expose relational
indices and retrieved context; the executive module should emit gating
and routing signals; working memory should expose maintained state with
controlled update permissions; and the multisensory interface should
bind only compatible cross-modal evidence. Hardware should follow the
same principle. Dense GPU matrix multiplication helped make the
Transformer the default architecture, but that historical convenience
does not prove that every cognitive operation belongs on the same dense
synchronous substrate {[}Hooker, 2021{]}. Neuromorphic and event-driven
systems demonstrate hardware support for sparse asynchronous
computation, while efficient deep neural network processing studies show
that memory movement can dominate energy cost, especially outside ideal
dense-kernel settings {[}Davies et al., 2018; Sze et al., 2017;
Narayanan et al., 2021{]}.

This research program is concrete: implement a minimal Heterogeneous
Topological Network with five modules and one interface layer, measure
transfer cost between modules, compare routing failures against
homogeneous Transformer baselines, and test whether structural priors
reduce data and compute requirements without relying on post-hoc
adaptation. That design is the architectural landing point of the
article. Intelligence should be built as coordinated structural
diversity, not as the indefinite enlargement of a single computational
form.

\begin{center}\rule{0.5\linewidth}{0.5pt}\end{center}

\subsection{References}\label{references}

\begin{itemize}
\tightlist
\item
  Jonas, Eric; Kording, Konrad Paul (2017). Could a Neuroscientist
  Understand a Microprocessor?. PLOS Computational Biology 13(1):
  e1005268. doi:10.1371/journal.pcbi.1005268
\item
  White, J. G. (1986). The structure of the nervous system of the
  nematode Caenorhabditis elegans. Philosophical Transactions of the
  Royal Society of London. B, Biological Sciences.
  doi:10.1098/rstb.1986.0056
\item
  James, Greg; Silverman, Barry; Silverman, Brian (2010). Visual6502:
  Transistor-Level Simulation of the MOS 6502 Microprocessor. Visual6502
  Project (visual6502.org)
\item
  Marr, David (1982). Vision: A Computational Investigation into the
  Human Representation and Processing of Visual Information. W. H.
  Freeman (San Francisco); reissued MIT Press, 2010
\item
  Craver, Carl F. (2007). Explaining the Brain: Mechanisms and the
  Mosaic Unity of Neuroscience. Oxford University Press
\item
  Seth, Anil K.; Barrett, Adam B.; Barnett, Lionel (2015). Granger
  Causality Analysis in Neuroscience and Neuroimaging. Journal of
  Neuroscience 35(8): 3293-3297. doi:10.1523/JNEUROSCI.4399-14.2015
\item
  Granger, Clive W. J. (1969). Investigating Causal Relations by
  Econometric Models and Cross-spectral Methods. Econometrica 37(3):
  424-438. doi:10.2307/1912791
\item
  Brown, Tom B.; Mann, Benjamin; Ryder, Nick; Subbiah, Melanie; Kaplan,
  Jared; et al.~(2020). Language Models are Few-Shot Learners. Advances
  in Neural Information Processing Systems (NeurIPS), vol.~33.
  doi:10.5555/3495724.3495883
\item
  Bullmore, Edward; Sporns, Olaf (2009). Complex brain networks: graph
  theoretical analysis of structural and functional systems. Nature
  Reviews Neuroscience. doi:10.1038/nrn2575
\item
  Logothetis, Nikos K. (2008). What we can do and what we cannot do with
  fMRI. Nature. doi:10.1038/nature06976
\item
  Zilles, Karl; Amunts, Katrin (2010). Centenary of Brodmann's map ---
  conception and fate. Nature Reviews Neuroscience. doi:10.1038/nrn2776
\item
  Dosovitskiy, Alexey; Beyer, Lucas; Kolesnikov, Alexander; Weissenborn,
  Dirk; et al.~(2021). An Image is Worth 16x16 Words: Transformers for
  Image Recognition at Scale. ICLR 2021 (arXiv:2010.11929).
  doi:10.48550/arXiv.2010.11929
\item
  Gong, Yuan; Chung, Yu-An; Glass, James R. (2021). AST: Audio
  Spectrogram Transformer. Interspeech 2021 (arXiv:2104.01778).
  doi:10.48550/arXiv.2104.01778
\item
  Brodmann, Korbinian (1909). Vergleichende Lokalisationslehre der
  Großhirnrinde in ihren Prinzipien dargestellt auf Grund des
  Zellenbaues. Johann Ambrosius Barth
\item
  Amunts, Katrin; et al.~(2020). Julich-Brain: A 3D probabilistic atlas
  of the human brain's cytoarchitecture. Science 369(6506): 988-992.
  doi:10.1126/science.abb4588
\item
  Cadwell, Cathryn R.; et al.~(2016). Electrophysiological,
  transcriptomic and morphologic profiling of single neurons using
  Patch-seq. Nature Biotechnology. doi:10.1038/nbt.3445
\item
  Hawrylycz, Michael J.; et al.~(2012). An anatomically comprehensive
  atlas of the adult human brain transcriptome. Nature.
  doi:10.1038/nature11405
\item
  Mountcastle, Vernon B. (1997). The columnar organization of the
  neocortex. Brain 120(4): 701-722. doi:10.1093/brain/120.4.701
\item
  Douglas, Rodney J.; Martin, Kevan A. C. (2004). Neuronal circuits of
  the neocortex. Annual Review of Neuroscience 27: 419-451.
  doi:10.1146/annurev.neuro.27.070203.144152
\item
  Harris, Kenneth D.; Shepherd, Gordon M. G. (2015). The neocortical
  circuit: themes and variations. Nature Neuroscience 18(2): 170-181.
  doi:10.1038/nn.3917
\item
  Lund, John S. (1988). Anatomical organization of macaque monkey
  striate visual cortex. Annual Review of Neuroscience 11: 253-288.
  doi:10.1146/annurev.ne.11.030188.001345
\item
  Born, Richard T.; Bradley, David C. (2005). Structure and Function of
  Visual Area MT. Annual Review of Neuroscience 28: 157-189.
  doi:10.1146/annurev.neuro.26.041002.131052
\item
  London, Michael; Häusser, Michael (2005). Dendritic Computation.
  Annual Review of Neuroscience.
  doi:10.1146/annurev.neuro.28.061604.135703
\item
  Stuart, Greg; Spruston, Nelson (2015). Dendritic integration: 60 years
  of progress. Nature Neuroscience. doi:10.1038/nn.4046
\item
  Callaway, Edward M.; Wiser, Abigail K. (1996). Contributions of
  individual layer 2-5 spiny neurons to local circuitry in macaque
  primary visual cortex. Visual Neuroscience 13(5): 907-922.
  doi:10.1017/S0952523800009159
\item
  Gouwens, Nathan W.; et al.~(2020). Integrated morphoelectric and
  transcriptomic classification of cortical GABAergic cells. Cell
  183(4): 935-953. doi:10.1016/j.cell.2020.09.057
\item
  Hubel, David H.; Wiesel, Torsten N. (1962). Receptive fields,
  binocular interaction and functional architecture in the cat's visual
  cortex. The Journal of Physiology. doi:10.1113/jphysiol.1962.sp006837
\item
  Krizhevsky, Alex; Sutskever, Ilya; Hinton, Geoffrey E. (2012).
  ImageNet Classification with Deep Convolutional Neural Networks.
  Advances in Neural Information Processing Systems (NeurIPS).
  doi:10.1145/3065386
\item
  LeCun, Yann; Bottou, Léon; Bengio, Yoshua; Haffner, Patrick (1998).
  Gradient-Based Learning Applied to Document Recognition. Proceedings
  of the IEEE. doi:10.1109/5.726791
\item
  Vaswani, Ashish; Shazeer, Noam; Parmar, Niki; Uszkoreit, Jakob; et
  al.~(2017). Attention Is All You Need. Advances in Neural Information
  Processing Systems (NeurIPS), Vol. 30. doi:10.48550/arXiv.1706.03762
\item
  Davies, Mike; et al.~(2018). Loihi: A Neuromorphic Manycore Processor
  with On-Chip Learning. IEEE Micro. doi:10.1109/MM.2018.112130359
\item
  Lennie, Peter (2003). The Cost of Cortical Computation. Current
  Biology. doi:10.1016/S0960-9822(03)00135-0
\item
  Hooker, Sara (2021). The Hardware Lottery. Communications of the ACM.
  doi:10.1145/3467017
\item
  Bommasani, Rishi; Hudson, Drew A.; Adeli, Ehsan; Altman, Russ; et
  al.~(2021). On the Opportunities and Risks of Foundation Models. arXiv
  preprint. doi:10.48550/arXiv.2108.07258
\item
  Fedus, William; Zoph, Barret; Shazeer, Noam (2022). Switch
  Transformers: Scaling to Trillion Parameter Models with Simple and
  Efficient Sparsity. Journal of Machine Learning Research, 23(120),
  1--39. doi:10.5555/3586589.3586709
\item
  Fodor, Jerry A. (1983). The Modularity of Mind. MIT Press
\item
  Shazeer, Noam; Mirhoseini, Azalia; Maziarz, Krzysztof; Davis, Andy; et
  al.~(2017). Outrageously Large Neural Networks: The Sparsely-Gated
  Mixture-of-Experts Layer. ICLR 2017 (arXiv:1701.06538).
  doi:10.48550/arXiv.1701.06538
\item
  Kaplan, Jared; McCandlish, Sam; Henighan, Tom; Brown, Tom B.; et
  al.~(2020). Scaling Laws for Neural Language Models. arXiv.
  doi:10.48550/arXiv.2001.08361
\item
  Eichenbaum, Howard (2004). Hippocampus: Cognitive Processes and Neural
  Representations that Underlie Declarative Memory. Neuron 44(1):
  109-120. doi:10.1016/j.neuron.2004.08.028
\item
  O'Keefe, John; Nadel, Lynn (1978). The Hippocampus as a Cognitive Map.
  Oxford University Press
\item
  Buzsáki, György; Tingley, David (2018). Space and Time: The
  Hippocampus as a Sequence Generator. Trends in Cognitive Sciences
  22(10): 853-869. doi:10.1016/j.tics.2018.07.006
\item
  Yassa, Michael A.; Stark, Craig E. L. (2011). Pattern Separation in
  the Hippocampus. Trends in Neurosciences 34(10): 515-525.
  doi:10.1016/j.tins.2011.06.006
\item
  Ramsauer, Hubert; Schäfl, Bernhard; Lehner, Johannes; Seidl, Philipp;
  et al.~(2021). Hopfield Networks is All You Need. ICLR 2021
  (arXiv:2008.02217). doi:10.48550/arXiv.2008.02217
\item
  Putnam, Hilary (1967). The Nature of Mental States. Art, Mind, and
  Religion / later reprinted in philosophical anthologies
\item
  Saenz, Melissa; Langers, David R. M. (2014). Tonotopic mapping of
  human auditory cortex. Hearing Research, 307, 42--52.
  doi:10.1016/j.heares.2013.07.016
\item
  Depireux, Didier A.; Simon, Jonathan Z.; Klein, Daniel J.; Shamma,
  Shihab A. (2001). Spectro-temporal response field characterization
  with dynamic ripples in ferret primary auditory cortex. Journal of
  Neurophysiology, 85(3), 1220--1234. doi:10.1152/jn.2001.85.3.1220
\item
  Stevens, Stanley Smith; Volkmann, John; Newman, Edwin B. (1937). A
  Scale for the Measurement of the Psychological Magnitude Pitch.
  Journal of the Acoustical Society of America. doi:10.1121/1.1915893
\item
  Frank, Michael J. (2005). Dynamic dopamine modulation in the basal
  ganglia: A neurocomputational account of cognitive deficits in
  medicated and nonmedicated Parkinsonism. Journal of Cognitive
  Neuroscience. doi:10.1162/0898929053895847
\item
  Funahashi, Shintaro; Bruce, Charles J.; Goldman-Rakic, Patricia S.
  (1989). Mnemonic coding of visual space in the monkey's dorsolateral
  prefrontal cortex. Journal of Neurophysiology.
  doi:10.1152/jn.1989.61.2.331
\item
  Stein, Barry E.; Stanford, Terrence R. (2008). Multisensory
  integration: current issues from the perspective of the single neuron.
  Nature Reviews Neuroscience. doi:10.1038/nrn2331
\item
  Putnam, Hilary (1975). Mind, Language and Reality: Philosophical
  Papers, Volume 2. Cambridge University Press
\item
  Alexander, Gerald E.; DeLong, Mahlon R.; Strick, Peter L. (1986).
  Parallel Organization of Functionally Segregated Circuits Linking
  Basal Ganglia and Cortex. Annual Review of Neuroscience 9: 357-381.
  doi:10.1146/annurev.ne.09.030186.002041
\item
  Frank, Michael J.; Loughry, Bryan; O'Reilly, Randall C. (2001).
  Interactions between frontal cortex and basal ganglia in working
  memory: A computational model. Cognitive, Affective, and Behavioral
  Neuroscience. doi:10.3758/CABN.1.2.137
\item
  Baddeley, Alan (2003). Working memory: looking back and looking
  forward. Nature Reviews Neuroscience. doi:10.1038/nrn1201
\item
  Moerel, Michelle (2014). An anatomical and functional topography of
  human auditory cortical areas. Frontiers in Neuroscience 8: 225.
  doi:10.3389/fnins.2014.00225
\item
  Sze, Vivienne; Chen, Yu-Hsin; Yang, Tien-Ju; Emer, Joel S. (2017).
  Efficient Processing of Deep Neural Networks: A Tutorial and Survey.
  Proceedings of the IEEE. doi:10.1109/JPROC.2017.2761740
\item
  Narayanan, Deepak; Shoeybi, Mohammad; Casper, Jared; LeGresley,
  Patrick; et al.~(2021). Efficient Large-Scale Language Model Training
  on GPU Clusters Using Megatron-LM. Proceedings of SC '21:
  International Conference for High Performance Computing, Networking,
  Storage and Analysis. doi:10.1145/3458817.3476209
\item
  Felleman, Daniel J.; Van Essen, David C. (1991). Distributed
  Hierarchical Processing in the Primate Cerebral Cortex. Cerebral
  Cortex 1(1): 1-47. doi:10.1093/cercor/1.1.1-a
\item
  Fukushima, Kunihiko (1980). Neocognitron: A self-organizing neural
  network model for a mechanism of pattern recognition unaffected by
  shift in position. Biological Cybernetics 36(4): 193-202.
  doi:10.1007/BF00344251
\end{itemize}

\begin{center}\rule{0.5\linewidth}{0.5pt}\end{center}

\subsection{Supplementary Material: The AI Crew Research
Workflow}\label{supplementary-material-the-ai-crew-research-workflow}

\subsubsection{S.1 Disclosure and Scope}\label{s.1-disclosure-and-scope}

This manuscript was produced by a human-directed multi-agent research
workflow: a single Independent Scientist (the author) operating a fixed
roster of large-language-model agents, referred to collectively below as
the \textbf{AI Crew}, inside a deterministic software harness. This
section discloses that workflow in full: what each agent was responsible
for, what was decided mechanically by code rather than by any model, and
--- most importantly --- what remained under the author's direct control
throughout.

The disclosure is offered in the interest of methodological
transparency, not as a claim of novelty in itself. The scientific
content of the manuscript --- its argument, its citations, and its
conclusions --- is the author's responsibility exactly as it would be if
drafted without LLM assistance. No agent in this pipeline was permitted
to introduce a claim, a reference, or a structural decision that
bypassed human review; every fact-bearing sentence in the manuscript
passed through an explicit approval gate described in S.3.

\subsubsection{S.2 What Stayed Under Author
Control}\label{s.2-what-stayed-under-author-control}

Three things were never delegated to any agent:

\begin{enumerate}
\def\labelenumi{\arabic{enumi}.}
\tightlist
\item
  \textbf{Idea and thesis.} The central claim --- that the Transformer's
  dominance across modalities is a hardware-driven accident rather than
  a principled convergence, and that cytoarchitectural evidence argues
  for a heterogeneous alternative --- originated with the author, not
  with a model prompt.
\item
  \textbf{Chapter architecture.} The table of contents
  (\texttt{docs/TOC\_en.txt}) that defines every Part, Chapter, and
  Section in this manuscript was authored by the human researcher before
  any drafting began, and functions as the single source of truth the
  harness checks all downstream work against (see S.4).
\item
  \textbf{Supervision and approval.} Every section produced by the
  drafting agents passed through an explicit accept/reject decision by
  the author before it could become part of the manuscript
  (\texttt{run.py\ approve} /
  \texttt{run.py\ reject\ "\textless{}feedback\textgreater{}"}).
  Sections were sent back for revision when they drifted from the
  intended argument, regardless of internal consistency or reference
  validity.
\end{enumerate}

\subsubsection{S.3 Zero-to-One Workflow}\label{s.3-zero-to-one-workflow}

Each section of the manuscript moved through the same fixed pipeline:

\begin{verbatim}
Author sets Current Task (from TOC)
        |
        v
   OMEGA (plan)  -- distributes a tailored brief to each drafting agent
        |
        v
  ALPHA | BETA | GAMMA   -- three independent drafts, written in isolation
        |
        v
   OMEGA (integrate)  -- synthesizes three drafts into one voice, verifies
        |                 every citation against the reference ledger,
        |                 rejects any [UNVERIFIED] claim
        v
   Author review  -- [Approve] or [Revision request: ...] (up to 3 feedback loops)
        |
        v
   chapters/T-xxx_*.md saved + reference ledger updated
        |
        +--> DELTA (Korean reference translation, optional, non-authoritative)
        |
        +--> EPSILON (format -> check -> assemble) -- cross-chapter unity pass
                    |
                    v
              OMEGA (confirm gate)  -- PASS/FAIL on the assembled manuscript
                    |
                    v
              ZETA (optional)  -- figure planning + image rendering, per section
\end{verbatim}

The drafting stage (ALPHA/BETA/GAMMA) runs three agents \textbf{in
isolation from one another} on the same brief, each contributing a
distinct disciplinary lens (S.4). OMEGA then rewrites --- not
concatenates --- their output into one continuous argument, a step
explicitly designed to prevent any single agent's phrasing from
dominating a section (see the excerpt in S.4). Only after the author
approved a section did it become an immutable artifact under
\texttt{chapters/}; nothing downstream (translation, editing, figure
generation) was permitted to alter approved scientific content, only its
presentation.

\subsubsection{S.4 Agent Roster}\label{s.4-agent-roster}

{\def\LTcaptype{none} % do not increment counter
\begin{longtable}[]{@{}
  >{\raggedright\arraybackslash}p{(\linewidth - 6\tabcolsep) * \real{0.2500}}
  >{\raggedright\arraybackslash}p{(\linewidth - 6\tabcolsep) * \real{0.2500}}
  >{\raggedright\arraybackslash}p{(\linewidth - 6\tabcolsep) * \real{0.2500}}
  >{\raggedright\arraybackslash}p{(\linewidth - 6\tabcolsep) * \real{0.2500}}@{}}
\toprule\noalign{}
\begin{minipage}[b]{\linewidth}\raggedright
Agent
\end{minipage} & \begin{minipage}[b]{\linewidth}\raggedright
Identity
\end{minipage} & \begin{minipage}[b]{\linewidth}\raggedright
Core Expertise
\end{minipage} & \begin{minipage}[b]{\linewidth}\raggedright
Underlying Model
\end{minipage} \\
\midrule\noalign{}
\endhead
\bottomrule\noalign{}
\endlastfoot
OMEGA & Principal Investigator / Orchestrator & SOTA validation,
reference verification, chapter closure review, synthesis, next-task
proposals & GPT-5.5 \\
ALPHA & The Neuro-Architect & Cytoarchitecture, Brodmann/Jülich mapping,
cellular morphology, critique of connectome-only methods & GPT-5.5 \\
BETA & The Critical Systems Engineer & Hardware Lottery, GPU/attention
cost, foundation-model homogenization, inductive-bias comparison &
Claude Sonnet 5 \\
GAMMA & The Functional Epistemologist & Functionalism, Modularity of
Mind, function-structure correspondence, ALPHA/BETA coherence review &
GPT-5.5 \\
DELTA & Bilingual Scholarly Translator & EN to KO reference translation
(non-authoritative; the manuscript of record is English) & Claude Sonnet
5 \\
EPSILON & Scientific Academic Editor & Cross-chapter style unification,
terminology consistency, reference reconciliation, manuscript assembly &
GPT-5.4 \\
ZETA & The Scientific Illustrator (optional) & Figure planning (caption
+ image prompt) from approved chapter text; does not render images
itself & Planner: GPT-5.4-mini · Renderer: Gemini image models \\
\end{longtable}
}

Each agent operated under a fixed, version-controlled system prompt plus
a shared, immutable constraints document (\texttt{AGENTS.md}) that no
per-agent instruction could override. Representative excerpts:

\begin{quote}
\textbf{AGENTS.md, Axiom 2 (Biological Grounding):} ``All AI
architecture critiques and alternatives must be grounded in concrete
physical and logical references from cytoarchitecture and cognitive
science. Metaphor is a tool for comprehension only --- it cannot serve
as the basis of an argument.''
\end{quote}

\begin{quote}
\textbf{AGENTS.md, §5-1 (Hallucination Prevention):} ``Factual claims
based on training data memory are prohibited\ldots{} Uncertain
figures/facts must be removed, not tagged. \texttt{{[}UNVERIFIED{]}}
placeholders are forbidden.''
\end{quote}

\begin{quote}
\textbf{OMEGA prompt:} ``You do not write prose directly. You validate
each agent's logic against the current state of the art (SOTA), design
the direction of argumentation, distribute instructions, synthesize
results, review quality, and propose the next task as choices for the
user's final approval.''
\end{quote}

\begin{quote}
\textbf{OMEGA prompt, on synthesis:} ``Your job is not to merge text ---
it is to rewrite three specialist drafts into one continuously readable
argument. A correct integration almost never reuses an agent's sentences
verbatim\ldots{} Never let one agent's contribution dominate by being
copied through.''
\end{quote}

\begin{quote}
\textbf{ALPHA prompt:} ``You are an expert who builds arguments using
cytoarchitecture and neuroscientific facts. You are unaware of other
agents. You work only from OMEGA's instructions and this prompt.''
\end{quote}

\begin{quote}
\textbf{BETA prompt:} ``Numbers and benchmarks first: back every claim
with FLOPs, parameter counts, or benchmark data\ldots{} The claim `this
is inefficient' is not allowed without a concrete numerical
comparison.''
\end{quote}

\begin{quote}
\textbf{GAMMA prompt:} ``An expert who defines the logical role each AI
architecture must play from the perspective of cognitive functionalism
and the Modularity of Mind, and who mediates between the other agents.''
\end{quote}

\begin{quote}
\textbf{EPSILON prompt:} ``Your single mandate is cross-chapter
UNITY\ldots{} You do NOT verify references. Whether a source is real and
its metadata complete is OMEGA's responsibility, settled before
approval.''
\end{quote}

\begin{quote}
\textbf{ZETA prompt (translated):} ``You do not draw figures yourself.
Instead, for each figure you produce (1) a caption and (2) a detailed
English prompt for the image-generation model. The actual rendering is
performed by the image model.''
\end{quote}

\subsubsection{S.5 The Reference
Discipline}\label{s.5-the-reference-discipline}

The most consequential harness-level constraint was a hard prohibition
on unverified claims: no source could be cited unless it was registered
in a shared reference ledger (\texttt{components/REFERENCE\_LEDGER.md})
with complete bibliographic metadata (author, year, title, venue, DOI
where available), and the pipeline enforced an
\texttt{unverified\_threshold} of zero --- a single
\texttt{{[}UNVERIFIED{]}} tag anywhere in a submitted draft was grounds
for automatic rejection at the OMEGA integration step. Agents proposing
a new source had to submit it as a structured \texttt{{[}NEW\_REF{]}}
block, including a verbatim excerpt, before OMEGA would register it as
\texttt{active} and permit its citation. This ledger, not any agent's
memory, is the sole permitted source of factual grounding for the
manuscript.

\subsubsection{S.6 What the Harness Did Mechanically (No Model
Involved)}\label{s.6-what-the-harness-did-mechanically-no-model-involved}

A deliberate design choice of this project was to keep every step that
could be made deterministic out of the hands of any language model. The
following were performed entirely by code:

\begin{itemize}
\tightlist
\item
  Parsing \texttt{docs/TOC\_en.txt} to determine chapter/section
  boundaries and to decide which drafted content counts as manuscript
  body versus post-process material
  (\texttt{Pipeline.\_is\_body\_task}).
\item
  Reading and writing \texttt{STATE\_TRACKER.md},
  \texttt{REFERENCE\_LEDGER.md}, and \texttt{TRANSLATION\_GLOSSARY.md}
  as structured Markdown/YAML state.
\item
  Assembling the final manuscript: inserting Part/Chapter headings at
  the correct position per the TOC, stripping unstable per-fragment
  headings, consolidating and deduplicating the reference list, and
  inserting this author/abstract/supplementary front matter.
\item
  Checkpointing every intermediate draft (\texttt{drafts/manifest.json})
  and logging the token cost of every model call
  (\texttt{cost\_log.jsonl}).
\item
  Sequencing figure numbering across the whole document once
  ZETA-selected images were inserted into individual chapters.
\end{itemize}

\subsubsection{S.7 Interpreting This
Disclosure}\label{s.7-interpreting-this-disclosure}

None of the above should be read as claiming the AI Crew produced this
manuscript autonomously, nor should it be read as a claim that
multi-agent drafting guarantees correctness. It did not: sections were
sent back for revision multiple times over the course of this project,
and the burden of verifying that OMEGA's SOTA judgments and
ALPHA/BETA/GAMMA's citations were in fact correct rested, and rests,
with the author. What the workflow provided was a disciplined division
of labor --- independent disciplinary drafts, an explicit synthesis
step, a zero-tolerance reference-verification gate, and a human approval
checkpoint before any content became permanent --- under the direction
of a single Independent Scientist functioning, in effect, as the
principal investigator of a small virtual laboratory.

\end{document}